\documentclass[10pt,twocolumn,letterpaper]{article}
\usepackage{colortbl}
\usepackage{array}
\usepackage{arydshln}
\usepackage[pagenumbers]{iccv}
\definecolor{iccvblue}{rgb}{0.21,0.49,0.74}
\usepackage[pagebackref,breaklinks,colorlinks,allcolors=iccvblue]{hyperref}
\usepackage[norelsize, linesnumbered, ruled, lined, boxed, commentsnumbered]{algorithm2e}
\usepackage{dsfont}
\usepackage{makecell}
\usepackage{multirow}

\definecolor{mypurple}{RGB}{105,52,117}
\definecolor{myyellow}{RGB}{253,236,81}
\definecolor{myblue}{RGB}{0,82,147}

\title{FreeFlux: Understanding and Exploiting Layer-Specific Roles in RoPE-Based MMDiT for Versatile Image Editing}
\author{ Tianyi Wei\textsuperscript{\rm 1}, Yifan Zhou\textsuperscript{\rm 1}, Dongdong Chen\textsuperscript{\rm 2}, Xingang Pan\textsuperscript{\rm 1}\\
	\normalsize\textsuperscript{\rm 1}S-Lab, Nanyang Technological University  \ \normalsize\textsuperscript{\rm 2}Microsoft GenAI  \  \\
	{\tt\small\{tianyi.wei, yifan006, xingang.pan\}@ntu.edu.sg }, {\tt\small cddlyf@gmail.com} \\
        {\tt\small \url{https://wtybest.github.io/projects/FreeFlux/}}
}

\begin{document}

\twocolumn[{
	\renewcommand\twocolumn[1][]{#1}
	\maketitle
	\setlength\tabcolsep{0.5pt}
	\centering
	\small
	\begin{tabular}{c}
		\includegraphics[width=0.99\textwidth]{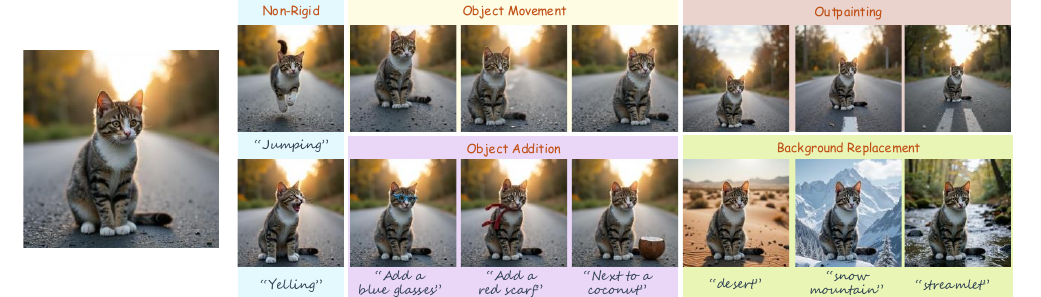}
	\end{tabular}
	\captionof{figure}{Leveraging the layer-specific roles in RoPE-based MMDiT we discovered, versatile training-free image editing is tailored to different task characteristics, including non-rigid editing, object addition, background replacement, object movement, and outpainting.}
	\label{fig:teaser}
        \vspace{2.0em}
}]

\maketitle
\begin{abstract}

The integration of Rotary Position Embedding (RoPE) in Multimodal Diffusion Transformer (MMDiT) has significantly enhanced text-to-image generation quality. However, the fundamental reliance of self-attention layers on positional embedding versus query-key similarity during generation remains an intriguing question. We present the first mechanistic analysis of RoPE-based MMDiT models (e.g., FLUX), introducing an automated probing strategy that disentangles positional information versus content dependencies by strategically manipulating RoPE during generation. Our analysis reveals distinct dependency patterns that do not straightforwardly correlate with depth, offering new insights into the layer-specific roles in RoPE-based MMDiT. Based on these findings, we propose a training-free, task-specific image editing framework that categorizes editing tasks into three types: position-dependent editing (e.g., object addition), content similarity-dependent editing (e.g., non-rigid editing), and region-preserved editing (e.g., background replacement). For each type, we design tailored key-value injection strategies based on the characteristics of the editing task. Extensive qualitative and quantitative evaluations demonstrate that our method outperforms state-of-the-art approaches, particularly in preserving original semantic content and achieving seamless modifications.

\end{abstract}    
\section{Introduction}
\label{sec:intro}

After years of rapid development, diffusion~\cite{ho2020denoising,songdenoising}-based text-to-image generation models have become the default modeling paradigm. Recently, new models, exemplified by FLUX~\cite{flux} and Stable Diffusion 3~\cite{esser2024scaling}, have pushed generation quality to an unprecedented level. Compared to previous state-of-the-art text-to-image models such as SD 1~\cite{rombach2022high}, SD 2, and SDXL~\cite{podell2023sdxl}, these models share several key improvements: a rectified flow formulation~\cite{liuflow}, a novel Multimodal Diffusion Transformer (MMDiT)~\cite{peebles2023scalable} replacing the conventional U-Net architecture~\cite{ronneberger2015u}, and a unified information interaction scheme where text and image tokens are concatenated and processed through self-attention with the classic query-key-value design.

A key distinction between FLUX and SD3 lies in how positional information is incorporated. SD3 injects Positional Embedding only at the input layer of the network, whereas FLUX applies Rotary Position Embedding (RoPE)~\cite{su2024roformer} to both queries and keys at every self-attention layer. This more explicit encoding of absolute and relative positional information enables FLUX to achieve superior generation quality and better high-resolution extrapolation, positioning it as a standout model in the text-to-image generation field. This naturally leads us to ask an important question: \textit{During generation, when each layer performs self-attention, \textbf{does FLUX rely more on the similarity between queries and keys, or on the positional embedding?}}

To answer this question, we propose a novel automated strategy to probe the dependence of different self-attention layers on positional relationships during generation. Specifically, during sampling, we manipulate each self-attention layer in FLUX by preserving the RoPE for queries while either removing or shifting the RoPE for keys, thereby obtaining the corresponding generated results. By measuring the similarity between the original sampled results and the modified outputs, we can infer the functional role of each layer: lower similarity indicates a stronger reliance on positional relationships, whereas higher similarity suggests a greater dependence on the content similarity between queries and keys.

Surprisingly, we find that layers relying on positional information and those relying on content similarity do not exhibit a simple correlation with their indices within the network. With these dependency patterns as a guideline, we suggest that \textit{\textbf{training-free image editing based on FLUX should be tailored to the specific characteristics of the editing task.}} Versatile image editing shares a classic editing mechanism~\cite{cao2023masactrl}: it generates both the source image and the edited image in parallel. During generation, editing is achieved by injecting keys and/or values from the source image into the edited image. Based on the nature of different editing tasks, we categorize versatile image editing into three types and design corresponding injection strategies for them: (1) \textit{Position-Dependent Editing}, (2) \textit{Content Similarity-Dependent Editing}, and (3) \textit{Region-Preserved Editing}. 

For \textit{Position-Dependent Editing}, such as object addition, we leverage the layers that are more reliant on positional information to propose a reasoning-before-generation strategy, effectively mitigating conflicts between object addition and consistency with the original content. For \textit{Content Similarity-Dependent Editing}, such as non-rigid editing, we perform modifications in layers that rely more on content similarity. For \textit{Region-Preserved Editing}, using background replacement as an example, we achieve perfect preservation of regions that should remain unchanged by injecting values across all layers. Furthermore, we demonstrate that this full-layer injection approach enables a highly harmonious blending effect, significantly outperforming latent blending~\cite{avrahami2023blended}. With minimal modifications, it can also be extended to object movement and outpainting tasks. Diverse image editing results are presented in Figure~\ref{fig:teaser}.

Qualitative and quantitative comparisons with state-of-the-art editing methods on object addition, non-rigid editing, and background replacement tasks demonstrate the superiority of our approach. Comprehensive ablation studies further validate the effectiveness of our key designs.

To summarize, our contributions are three-fold as below:
\begin{itemize}
	\item We are the first to reveal the existence of different dependency mechanisms in self-attention layers of RoPE-based MMDiT models during generation.

        \item Guided by this mechanism, we tailor different editing methods for versatile image editing tasks based on their characteristics.
	
	\item The strong editing performance demonstrates the great potential of our approach.
\end{itemize}
\section{Related Work}
\label{sec:related}

\noindent\textbf{Text-to-Image Diffusion Models.} Marking the emergence of Latent Diffusion~\cite{rombach2022high}, diffusion-based models~\cite{gu2022vector,podell2023sdxl,saharia2022photorealistic,ramesh2022hierarchical} have come to dominate the text-to-image generation field. Among them, the U-Net-based Stable Diffusion family (SD 1, SD 2, SDXL) gained immense popularity due to its high synthesis quality and open-source nature. Recently, more powerful MMDiT-based models, such as SD3~\cite{esser2024scaling} and FLUX~\cite{flux}, have drawn significant attention from the community. Both SD3 and FLUX adopt a rectified flow formulation~\cite{liuflow,lipmanflow} and replace the U-Net~\cite{ronneberger2015u} with MMDiT~\cite{peebles2023scalable}, which incorporates numerous self-attention layers that enable rich interactions between textual and visual information. Unlike SD3, which injects positional embeddings only at the input layer, FLUX explicitly applies Rotary Position Embedding (RoPE)~\cite{su2024roformer} to both queries and keys in every self-attention layer. This distinction raises our curiosity about the role of RoPE in self-attention computations. In this paper, we investigate the positional dependency mechanisms in self-attention layers of RoPE-based MMDiT models and leverage these insights to develop task-specific editing strategies for different image editing tasks.

\noindent\textbf{Text-Guided Image Editing.} Text-guided image editing is a challenging task that aims to modify images based on textual descriptions. Early text-driven image editing methods~\cite{patashnik2021styleclip,xia2021tedigan,wei2022hairclip,wei2023hairclipv2,li2020manigan} were typically based on Generative Adversarial Networks (GANs)~\cite{Goodfellow2014GenerativeAN,karras2019style,karras2020analyzing,karras2021alias}. While they achieved promising results in specific domains (\eg, faces), they often struggled with editing images across arbitrary domains. With the rapid advancement of diffusion-based text-to-image models, diffusion-based image editing techniques have become the mainstream approach.

Diffusion-based image editing methods can be categorized into training-free and training-based approaches. Training-free methods~\cite{mengsdedit,avrahami2023blended,li2024zone,ravi2023preditor,cao2023masactrl,hertz2023prompt,tumanyan2023plug,tewel2024add} leverage pretrained text-to-image models with high-quality generation capabilities and achieve editing through techniques such as prompt refinement~\cite{ravi2023preditor}, attention sharing~\cite{cao2023masactrl,hertz2023prompt,tumanyan2023plug}, and mask guidance~\cite{avrahami2023blended,li2024zone}. Training-based methods~\cite{kawar2023imagic,brooks2023instructpix2pix,zhang2023magicbrush,sheynin2024emu,xiao2024omnigen} typically train from scratch or fine-tune well-established diffusion models using datasets constructed with editing instructions. Recently, a few training-free editing methods for FLUX emerged. StableFlow~\cite{avrahami2024stable} identifies layers crucial to image formation and leverages them for editing, while TamingRF~\cite{wang2024taming} performs edits within FLUX’s single-stream blocks. However, neither method considers task-specific customization of editing strategies based on the characteristics of different editing tasks. In this paper, we propose customized training-free editing strategies for versatile image editing, leveraging our findings on the generation mechanisms of RoPE-based MMDiT.
\section{Proposed Method}
\label{sec:methods}

\begin{figure}[t]
	\centering
	\includegraphics[width=0.95\columnwidth]{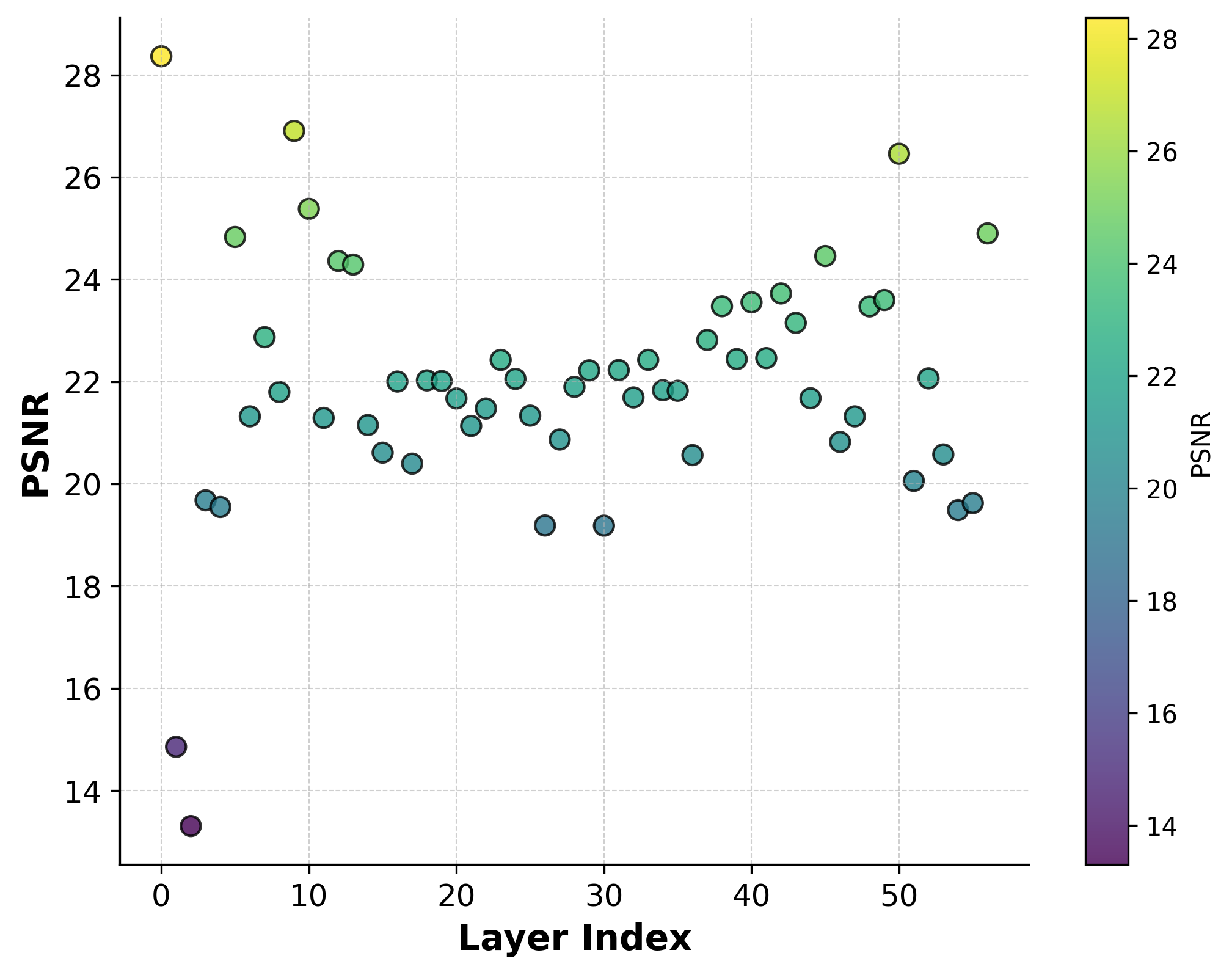} 
        \vspace{-1.0em}
	\caption{Quantitative analysis of the positional dependency of joint self-attention layers in RoPE-based MMDiT. Lower PSNR values indicate a stronger dependence on positional information, while higher PSNR values suggest a greater reliance on the content similarity between query and key.}
	\label{fig:layer_dep_scatter}
        \vspace{-1.0em}
\end{figure}

\subsection{Preliminaries}

\noindent\textbf{FLUX}~\cite{flux} follows the classic design of the Stable Diffusion family, performing diffusion sampling in the autoencoder's latent space to reduce computational cost. After an $8\times$ downsampling compression by the autoencoder and an additional $2\times$ downsampling patchify operation, the input image is transformed to $\frac{1}{16}$ of its original resolution. For example, a $1024\times 1024$ image is converted into $64\times 64=4096$ image tokens for diffusion sampling.

In terms of network architecture, FLUX replaces the traditional U-Net framework used in previous SD models with the more expressive Multimodal Diffusion Transformer (MMDiT)~\cite{peebles2023scalable}. MMDiT employs a joint self-attention mechanism to process concatenated text and image tokens in a unified attention operation, enabling bidirectional interactions that capture both self-information and cross-modal information between text and image modalities. For the widely adopted $12$ Billion version of FLUX.1-dev, its MMDiT consists of $57$ layers of blocks, where the first $19$ blocks are multi-stream blocks, followed by $38$ single-stream blocks.

\noindent\textbf{Joint Self-Attention in MMDIT Blocks.} The only difference between multi-stream blocks and single-stream blocks is that multi-stream blocks use separate projection matrices $(\mathbf{W}_Q, \mathbf{W}_K, \mathbf{W}_V)$ for text and image tokens, whereas single-stream blocks share the same projection matrix for both modalities. Both types of blocks contain a joint self-attention layer, which is formally computed as follows:
\begin{multline}
	Attn=softmax([Q_{txt},RoPE(Q_{img})]\\
    [K_{txt},RoPE(K_{img})]^{\top} / \sqrt{d_k})\cdot[V_{txt}, V_{img}],
\label{eq:attn_formu}
\end{multline}
where $[Q_{txt},RoPE(Q_{img})]$ denotes the concatenation of text and image query tokens, with the same concatenation operation applied to key $K$ and value $V$ tokens. To effectively enhance the perception of both absolute and relative positions during computation, FLUX explicitly injects Rotary Positional Embedding (RoPE)~\cite{su2024roformer} into the query and key at each self-attention layer. Since FLUX injects all-zero positional encoding into text tokens, we apply RoPE only to image tokens in the above formulation for better clarity.

\begin{figure}[t]
	\begin{center}
		\setlength{\tabcolsep}{0.5pt}
		\begin{tabular}{m{0.3cm}<{\centering}m{1.55cm}<{\centering}|m{1.55cm}<{\centering}m{1.55cm}<{\centering}m{1.55cm}<{\centering}m{1.55cm}<{\centering}}
			& \scriptsize{Sampled Image} & \scriptsize{Remove RoPE} & \scriptsize{Shift $(0,20)$} & \scriptsize{Shift $(10,10)$} & \scriptsize{Shift $(64,0)$}
			\\

			\multirow{2}{*}{\raisebox{-0.75cm}{\rotatebox[origin=c]{90}{\footnotesize{{\textcolor{myyellow}{Layer 0}}}}}}
			&\includegraphics[width=1.5cm]{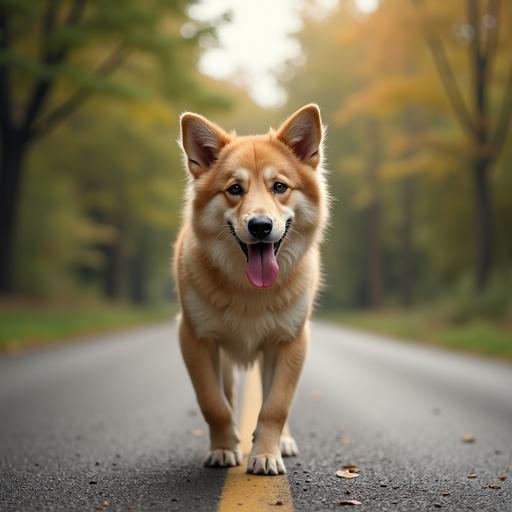}
			&\includegraphics[width=1.5cm]{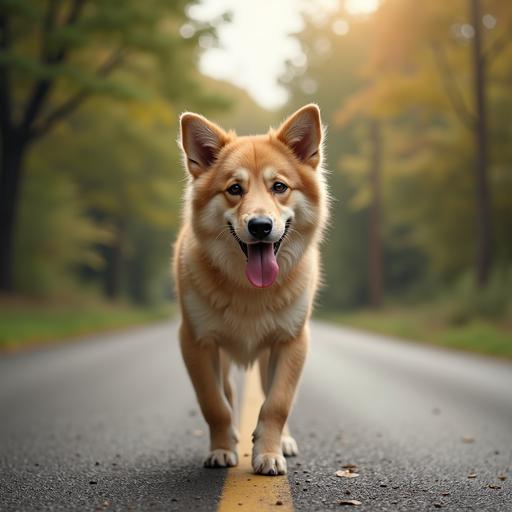}
			&\includegraphics[width=1.5cm]{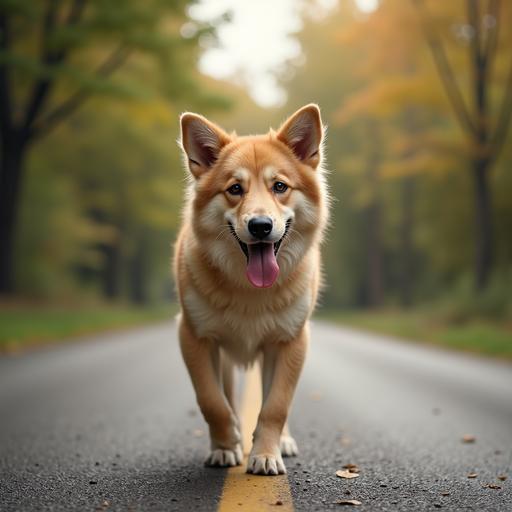}
			&\includegraphics[width=1.5cm]{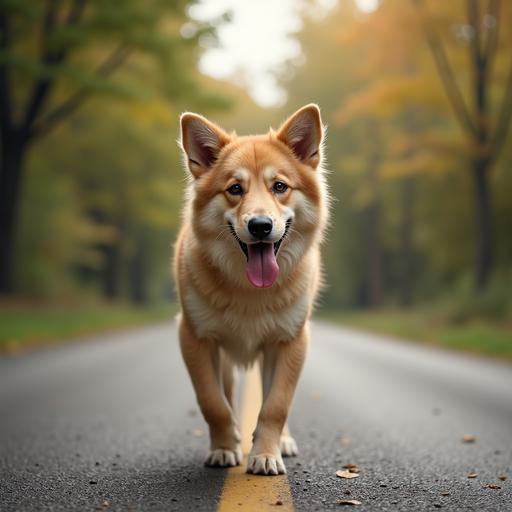}
			&\includegraphics[width=1.5cm]{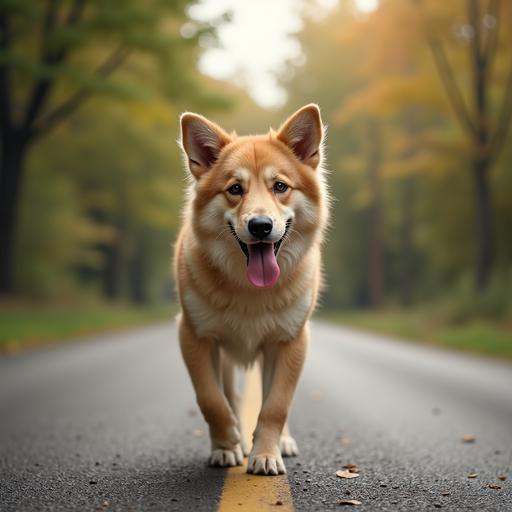}	
			\\
			&\includegraphics[width=1.5cm]{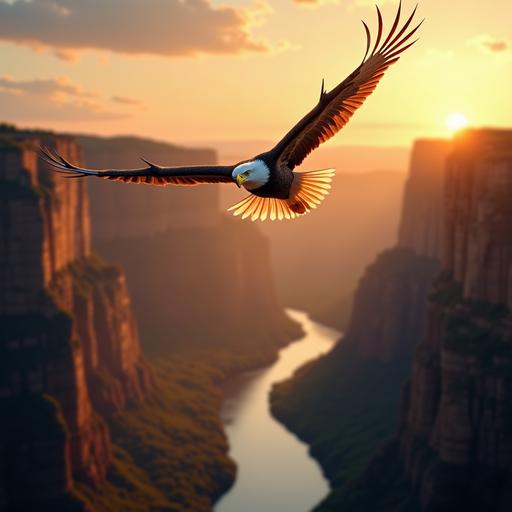}
			&\includegraphics[width=1.5cm]{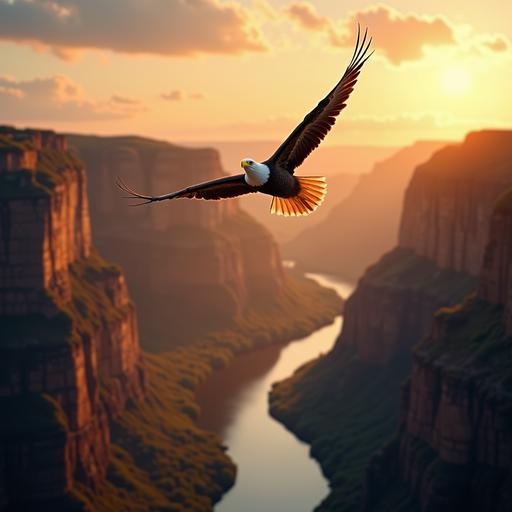}
			&\includegraphics[width=1.5cm]{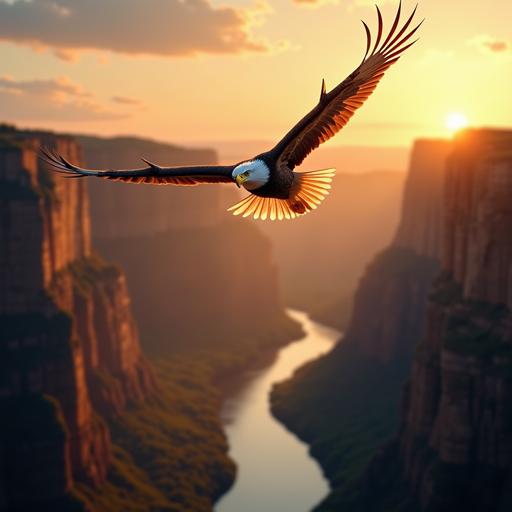}
			&\includegraphics[width=1.5cm]{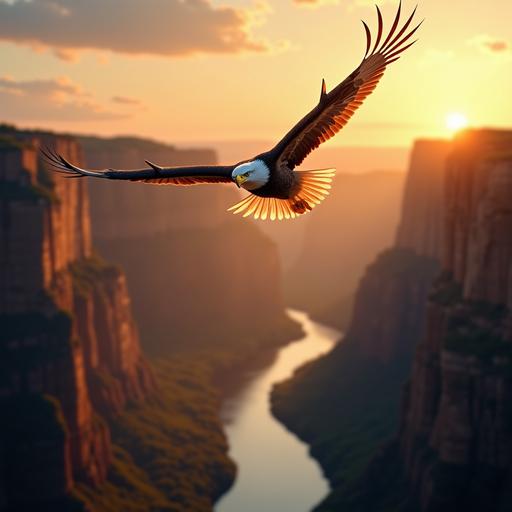}
			&\includegraphics[width=1.5cm]{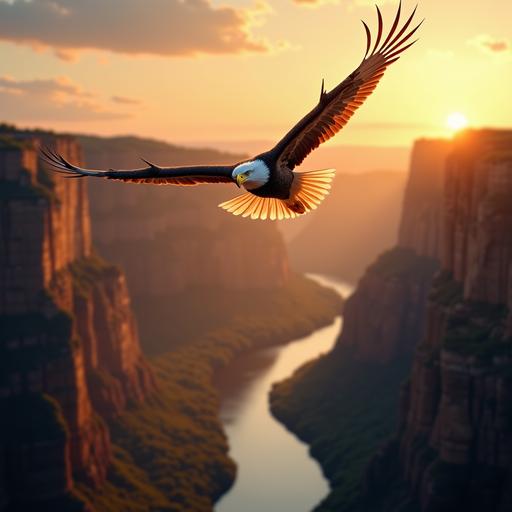}	
			\\

                \cdashline{1-6}
                \noalign{\vskip 0.15cm}
            
			\multirow{2}{*}{\raisebox{-0.75cm}{\rotatebox[origin=c]{90}{\footnotesize{{\textcolor{mypurple}{Layer 2}}}}}}
			&\includegraphics[width=1.5cm]{images/layer_dep_visual/dog_sample.jpg}
			&\includegraphics[width=1.5cm]{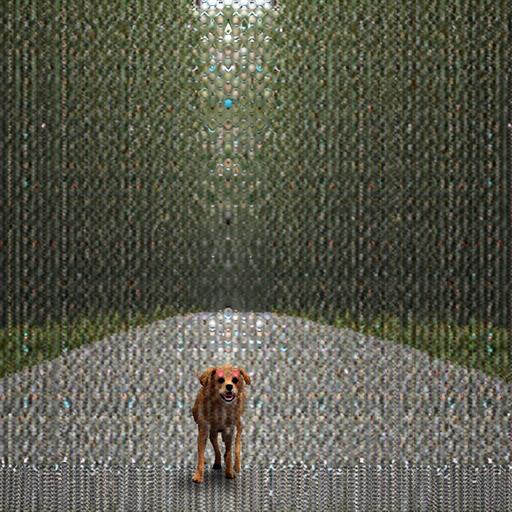}
			&\includegraphics[width=1.5cm]{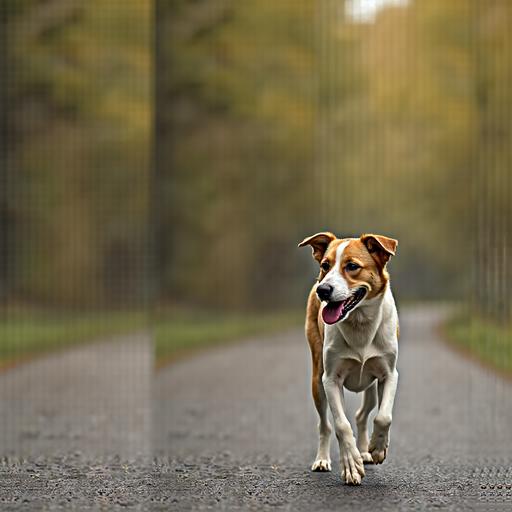}
			&\includegraphics[width=1.5cm]{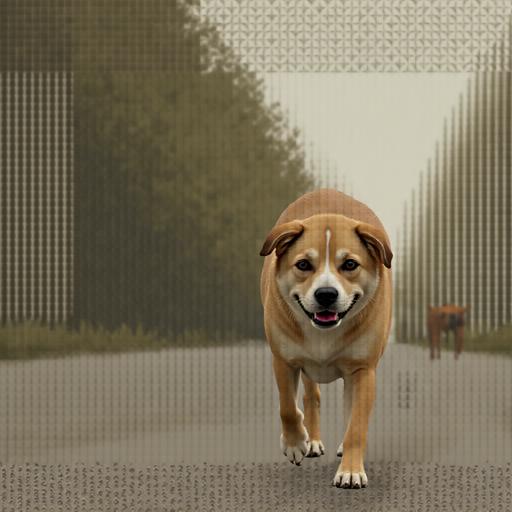}
			&\includegraphics[width=1.5cm]{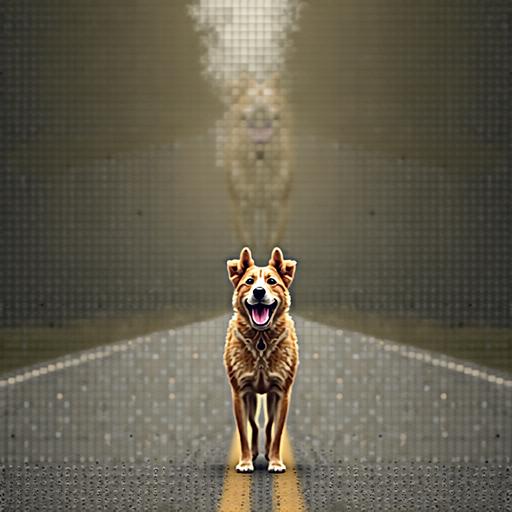}	
			\\
			&\includegraphics[width=1.5cm]{images/layer_dep_visual/eagel_sample.jpg}
			&\includegraphics[width=1.5cm]{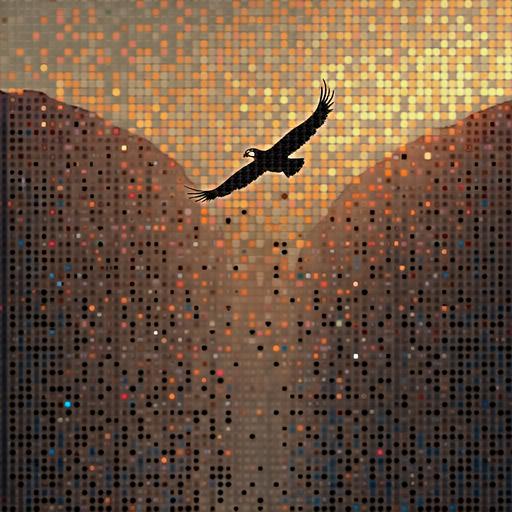}
			&\includegraphics[width=1.5cm]{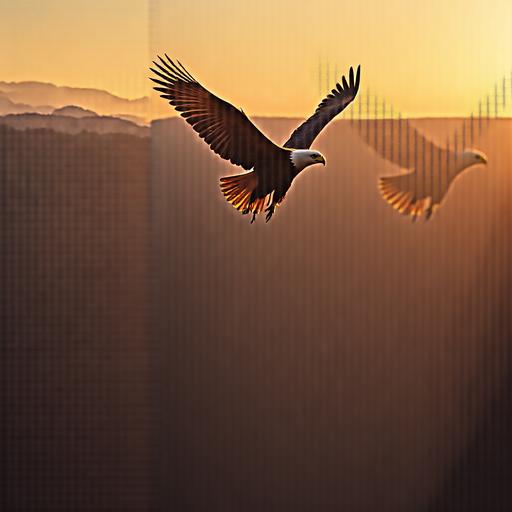}
			&\includegraphics[width=1.5cm]{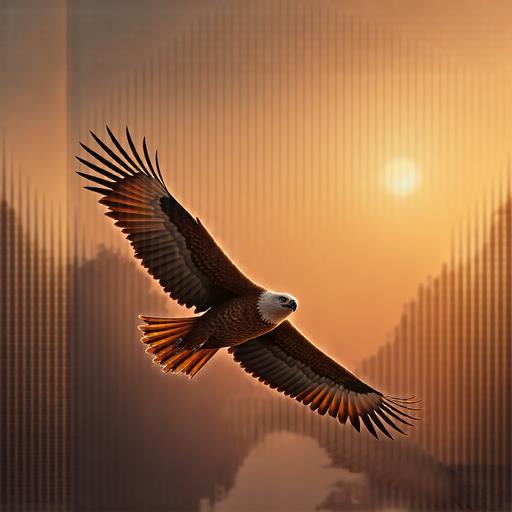}
			&\includegraphics[width=1.5cm]{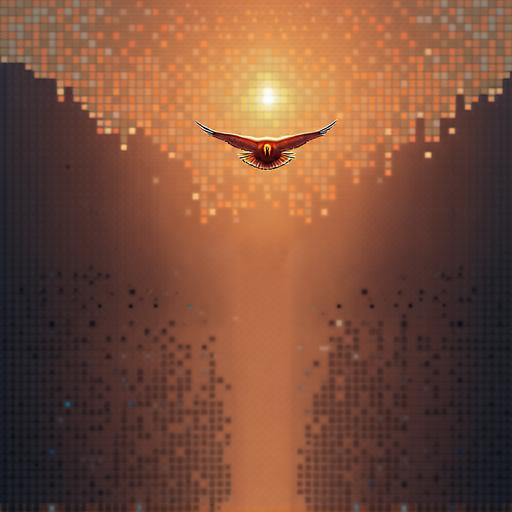}	
			\\
			
		\end{tabular}
	\end{center}
        \vspace{-1.5em}
	\caption{Visual results of modifying the RoPE of $K$ at different layers. Here, we present the sampled and probing images for Layer $2$ (the most position-dependent, with the lowest PSNR) and Layer $0$ (the most content-similarity-dependent, with the highest PSNR). ``Shift $(0,20)$" indicates that the RoPE of $K$ is shifted by $20$ positions in the horizontal direction only at the probed layer.} 
	\label{fig:layer_dep_visual}
        \vspace{-0.5em}
\end{figure}

\subsection{Probing Layer-wise Positional Dependency}

Thanks to the explicit injection of positional information into query and key at each layer via RoPE, FLUX demonstrates significantly superior performance over SD3 in both generation quality and high-resolution synthesis, making it a focal point in the text-to-image domain. This also raises an intriguing question: During generation, \textit{does the RoPE-based MMDiT rely on positional embedding to retrieve information, or does it depend on the content similarity between query and key?}

To address this question, we designed an automated probing strategy to understand the dependency of each self-attention layer on positional information during generation. Specifically, we first used ChatGPT~\cite{openai2024chatgpt} to generate $N$ $(N = 50)$ diverse text descriptions. For each description, we sampled $M$ $(M = 5)$ random seeds to synthesize images, resulting in $N \times M$ sampled images as our ground truth. Next, for each sampled image, we generated a series of probing images. Keeping the text description and initial seed unchanged, we systematically modified RoPE layer by layer across the $57$ blocks of FLUX (altering only one layer at a time while keeping RoPE intact in the others) to obtain the corresponding probing images.

Regarding the RoPE modification strategy, we keep the RoPE of the query $Q$ unchanged while either removing or shifting the RoPE of the key $K$, thereby altering the positional relationship between query and key. For a $1024 \times 1024$ image, the positional encoding originally ranges from $(0,0)$ to $(63,63)$. When shifting the positional encoding by $(10,10)$ in both the vertical and horizontal directions, the key's positional encoding is adjusted to range from $(10,10)$ to $(73,73)$. When RoPE is removed, Equation~\ref{eq:attn_formu} transforms into the following form:
\begin{multline}
	Attn=softmax([Q_{txt},RoPE(Q_{img})]\\
    [K_{txt},K_{img}]^{\top} / \sqrt{d_k})\cdot[V_{txt}, V_{img}].
\end{multline}

After obtaining all probing images, we compute the similarity (PSNR~\cite{Hor2010ImageQM}) between each probing image and its corresponding original sampled image for each layer and take the average. This serves as an indicator of the layer’s dependency on positional embeddings during self-attention operations. The statistical results are shown in Figure~\ref{fig:layer_dep_scatter}, where lower PSNR values indicate a stronger dependence on positional information, while higher PSNR values suggest a greater reliance on the content similarity between query and key. Surprisingly, we found that in RoPE-based MMDiT, \textit{whether a self-attention layer relies more on positional information or content similarity does not exhibit a simple correlation with its layer index.}

In Figure~\ref{fig:layer_dep_visual}, we present the visual results of Layer $2$, which exhibits the strongest dependence on positional information, and Layer $0$, which relies most on content similarity. It is evident that modifying the RoPE of $K$ in Layer $0$ has little to no impact, as the probing images remain nearly identical to the original sampled images. In contrast, altering the RoPE of $K$ in Layer $2$ leads to significant differences. For instance, removing RoPE severely degrades image quality, while shifting positions (by $20$ positions horizontally or $10$ positions in both horizontal and vertical directions) introduces noticeable stripe-like artifacts in the probing images, where the model appears to treat the remaining visible region as the available canvas for generation. We hypothesize that this occurs because Layer $2$ heavily relies on absolute positional information to retrieve content. For example, a query at the top-left $(0,0)$ position fails to match with any key, as the keys have been shifted and no longer contain $(0,0)$, leading to artifacts. Inspired by this intriguing observation, we began \textit{exploring customized editing strategies tailored to different editing tasks based on their specific characteristics.}

\subsection{Customized Strategies for Versatile Editing}
In U-Net-based diffusion models, image editing methods~\cite{cao2023masactrl} leveraging attention sharing mechanism have achieved remarkable success. This approach performs editing by replacing the key and value of the edited image with those from the source image in the self-attention layers, transferring the source image's appearance to the edited image and producing the result. In this work, we extend this mechanism to the joint self-attention layers of MMDiT. We generate the source image and the edited image in parallel using the same initial noise, where the source image is conditioned on a textual description of this image, while the edited image is conditioned on the same description with additional text specifying the desired edits. Since our approach does not modify the query, key, or value of the text tokens, for clarity, we denote Equation~\ref{eq:attn_formu} as:$Attn(Q,K,V)$, where $Q$, $K$, and $V$ specifically refer to the image token representations $Q_{img}$, $K_{img}$, and $V_{img}$, respectively. Accordingly, the self-attention operations of the parallelly generated source image and edited image at the $t$-th timestep in the $i$-th block are denoted as $Attn(Q_{src}^{t-i},K_{src}^{t-i},V_{src}^{t-i})$ and $Attn(Q_{edit}^{t-i},K_{edit}^{t-i},V_{edit}^{t-i})$, respectively. Next, we elaborate on the customized attention-sharing strategies tailored for versatile image editing based on the specific characteristics of each task.

\noindent\textbf{Position-Dependent Editing.} Object addition is a typical position-dependent editing task, aiming to add the specified object to an appropriate region of the source image based on the editing instruction. This task requires not only seamless object integration but also pixel-wise preservation of the irrelevant regions. Therefore, we choose to perform attention sharing in the most position-dependent layers. This allows us to inject source image information into the edited image at these layers while preserving the flexibility of the remaining layers to follow the textual editing instructions and achieve the object addition goal. The formalized attention-sharing mechanism for the most position-dependent layers is as follows: 
\begin{equation}
	Attn(Q_{edit}^{t-i},K_{src}^{t-i},V_{src}^{t-i}), i \in \mathds{P},
\label{eq:add_obj_reasoning}
\end{equation}
where $\mathds{P}$ represents the most position-dependent layers.

\begin{figure}[t]
	\centering
	\includegraphics[width=0.99\columnwidth]{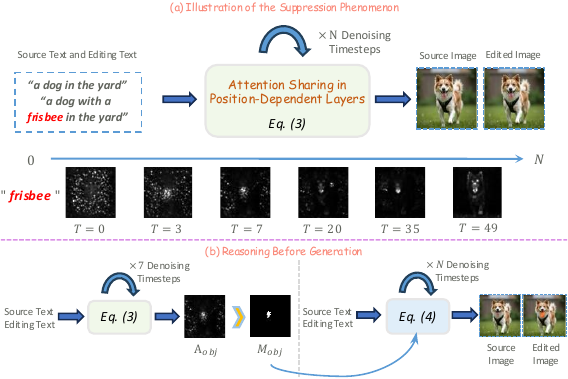} 
	\caption{Illustration of the suppression phenomenon and the reasoning-before-generation process.}
        \vspace{-1.0em}
	\label{fig:add_it_visual}
\end{figure}

However, we observed that in some cases, the generated edited image replicates the source image without adding the intended object. We believe this is reasonable: injecting all key and value features from the source image into the most position-dependent layers suppresses the generation of new objects. By visualizing the cross-attention between text tokens and image tokens in the joint self-attention layers, we observed a suppression effect consistent with our hypothesis. As shown in Figure~\ref{fig:add_it_visual} (a), during the early denoising stages (the first few timesteps), the attention mask corresponding to the added object's word gradually highlights the intended region for object placement. However, as denoising progresses, the activated region gradually shrinks and disperses, eventually losing its significance.

To resolve this conflict, we propose a novel \textit{\textbf{Reasoning-before-Generation}} strategy, illustrated in Figure~\ref{fig:add_it_visual} (b). In the initial steps of denoising, we first apply the attention-sharing mechanism from Equation~\ref{eq:add_obj_reasoning} to inject spatial information from the source image into the edited image. Then, at a certain early timestep, we leverage the strong reasoning capability of the joint self-attention to obtain the attention mask $A_{obj}$ for the added object, which highlights the appropriate region where the object should appear. Next, we binarize $A_{obj}$ using a threshold of $0.3$ and extract the largest connected component to obtain $M_{obj}$, which indicates the object's placement and defines the region where suppression should be alleviated. After obtaining $M_{obj}$, we restart the parallel sampling process with the same seed, and the attention-sharing mechanism is updated as follows:
\begin{equation}
	Attn(Q_{edit}^{t-i},K_{obj}^{t-i},V_{obj}^{t-i}), i \in \mathds{P},
\label{eq:add_obj_final}
\end{equation}
where $K_{obj}^{t-i}=M_{obj}\times K_{edit}^{t-i}+(1-M_{obj})\times K_{src}^{t-i}$ and $V_{obj}^{t-i}=M_{obj}\times V_{edit}^{t-i}+(1-M_{obj})\times V_{src}^{t-i}$. By injecting source image information only into the irrelevant regions, we perfectly resolve the conflict between the synthesis of the added object and content preservation.

\noindent\textbf{Content Similarity-Dependent Editing.} Non-rigid editing refers to modifications involving non-rigid deformations of objects in an image, typically affecting shape, pose, and surface details—for example, ``making a standing dog jump". Injecting source image information based on spatial position alone fails to achieve satisfactory results for such edits. Therefore, we perform attention sharing in layers that rely more on content similarity, as follows:
\begin{equation}
	Attn(Q_{edit}^{t-i},K_{src}^{t-i},V_{src}^{t-i}), i \in \mathds{C},
\label{eq:non-rigid}
\end{equation}
where $\mathds{C}$ represents the layers that rely more on content similarity. In this way, the non-rigid deformation is guided by the editing text (\eg, ``a jumping dog"), while the texture information (\eg, the dog's appearance) is transferred from the source image through layers that rely more on content similarity, enabling non-rigid editing.

\noindent\textbf{Region-Preserved Editing.} We use background replacement as an example to illustrate Region-Preserved Editing. Background replacement requires pixel-level preservation of the foreground object while modifying the background content. We first obtain a coarse mask $M_{fg}$ of the foreground object using the method similar to that in Position-Dependent Editing, except that this mask is derived from the cross-attention part of the source image. To achieve a more precise foreground mask, we propose an automated strategy: several foreground points are randomly sampled from $M_{fg}$ as inputs for SAM-2~\cite{ravi2024sam}, generating an accurate mask $M_{fg}^{sam}$ of the foreground object. The attention-sharing strategy is designed as follows:
\begin{equation}
	Attn(Q_{edit}^{t-i},K_{edit}^{t-i},V_{fg}^{t-i}), i \in \mathds{L},
\label{eq:bg_replace}
\end{equation}
where $\mathds{L}$ represents all layers, and $V_{fg}^{t-i}=M_{fg}^{sam}\times V_{src}^{t-i}+(1-M_{fg}^{sam})\times V_{edit}^{t-i}$. We found that this value-only replacement strategy effectively preserves the target region in both position-dependent and content-similarity-dependent layers, while background modification is guided by the desired background described in the editing text. In the ablation study, we will demonstrate that our proposed \textit{\textbf{value replacement}} outperforms the classical latent blending~\cite{avrahami2023blended}.

\begin{figure*}[t]
	\begin{center}
		\setlength{\tabcolsep}{0.5pt}
		\begin{tabular}{m{0.3cm}<{\centering}m{1.85cm}<{\centering}m{1.85cm}<{\centering}m{1.85cm}<{\centering}m{1.85cm}<{\centering}m{1.85cm}<{\centering}m{1.85cm}<{\centering}m{1.85cm}<{\centering}m{1.85cm}<{\centering}m{1.85cm}<{\centering}}
			 & \multicolumn{3}{c}{\small{Object Addition}} & \multicolumn{3}{c}{\small{Non-Rigid Editing}} & \multicolumn{3}{c}{\small{Background Replacement}}
			\\
   
			\raisebox{0.3cm}{\rotatebox[origin=c]{90}{\footnotesize{{Source}}}}
			&\includegraphics[width=1.8cm]{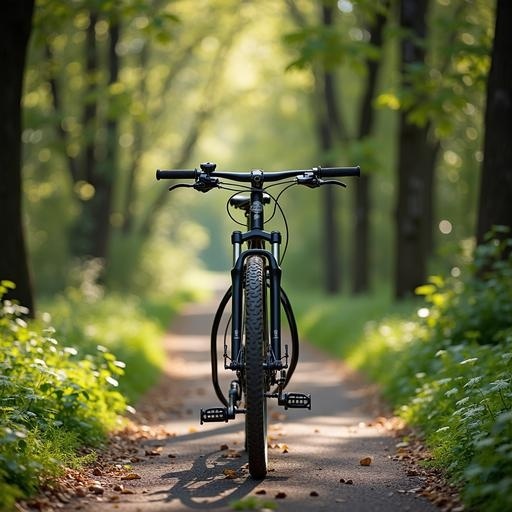}
			&\includegraphics[width=1.8cm]{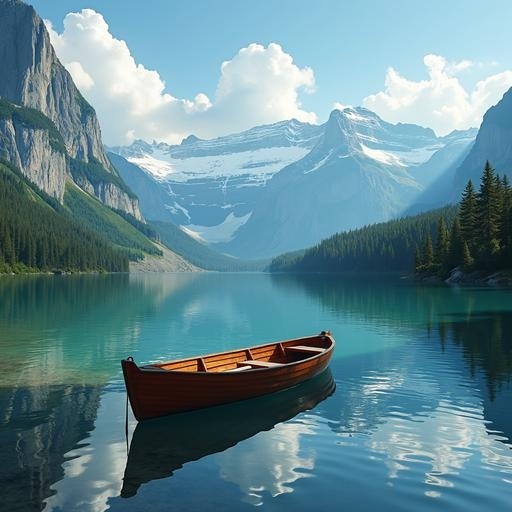}
			&\includegraphics[width=1.8cm]{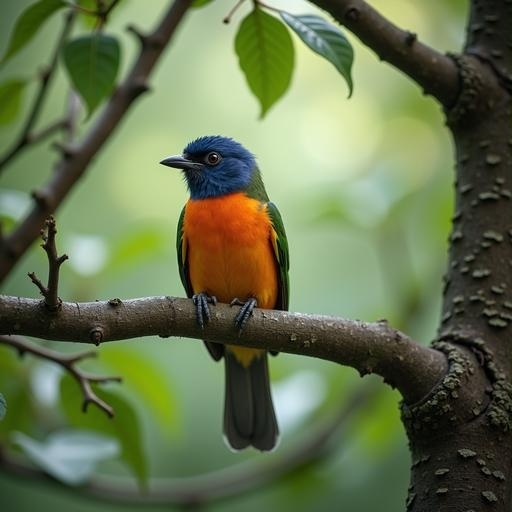}
			&\includegraphics[width=1.8cm]{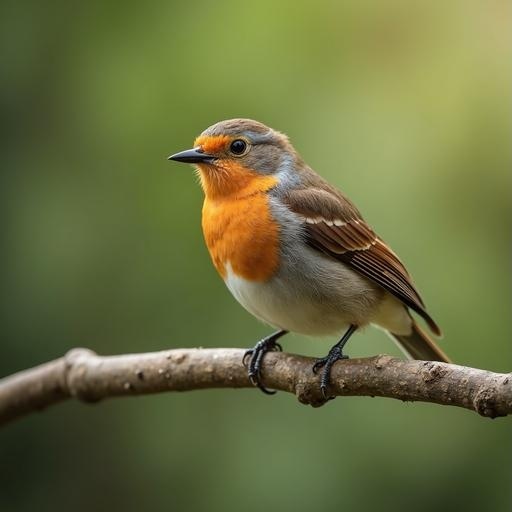}
			&\includegraphics[width=1.8cm]{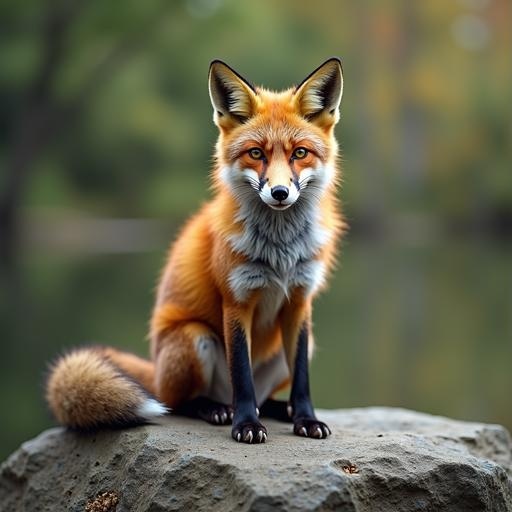}
			&\includegraphics[width=1.8cm]{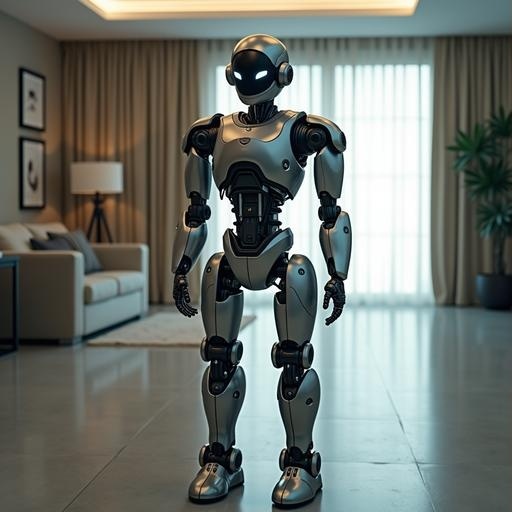}
			&\includegraphics[width=1.8cm]{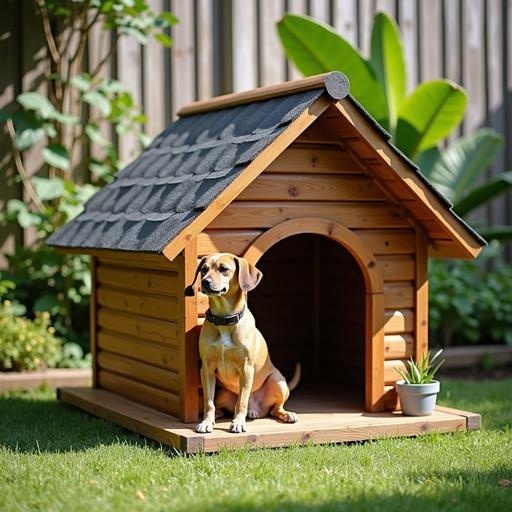}
			&\includegraphics[width=1.8cm]{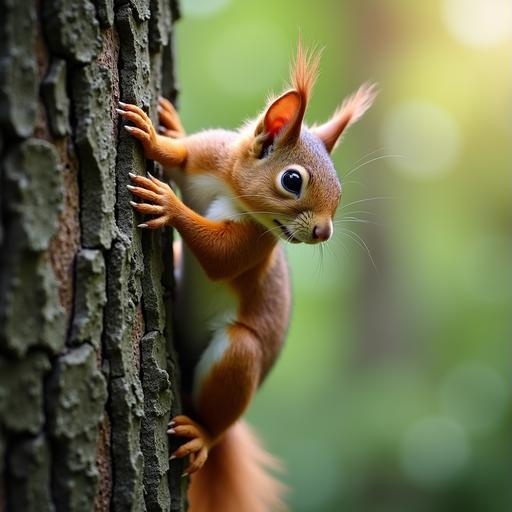}
			&\includegraphics[width=1.8cm]{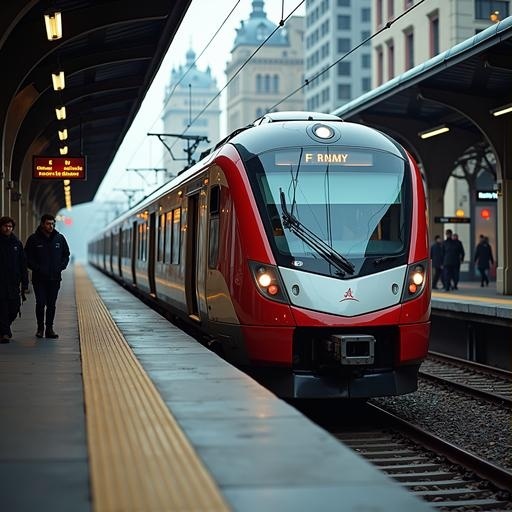}
			\\

			\raisebox{0.4cm}{\rotatebox[origin=c]{90}{\footnotesize{{StableFlow}}}}
			&\includegraphics[width=1.8cm]{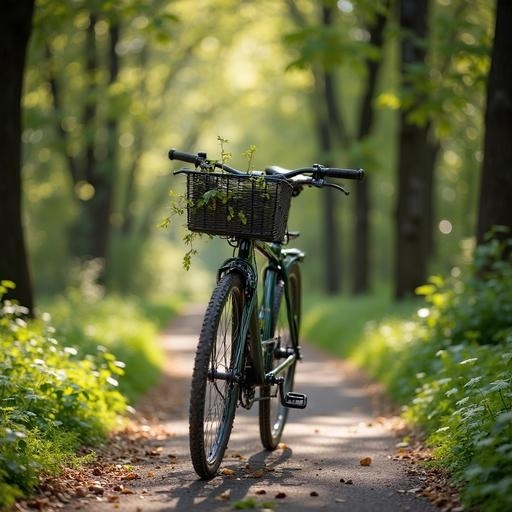}
			&\includegraphics[width=1.8cm]{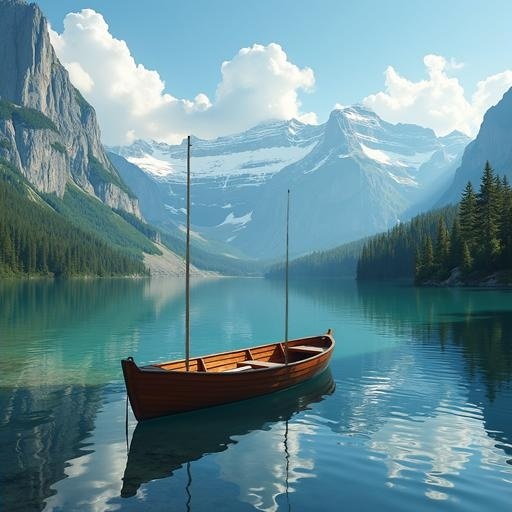}
			&\includegraphics[width=1.8cm]{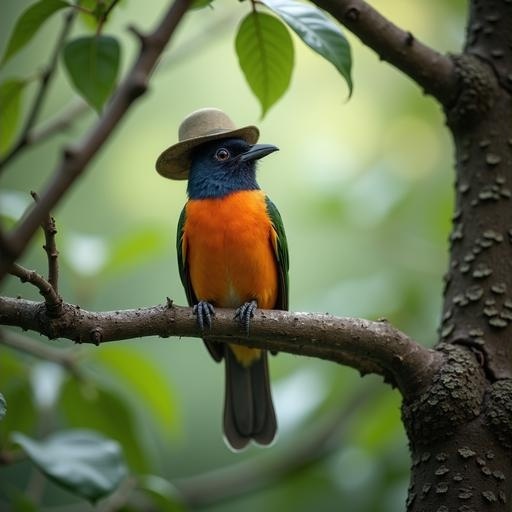}
			&\includegraphics[width=1.8cm]{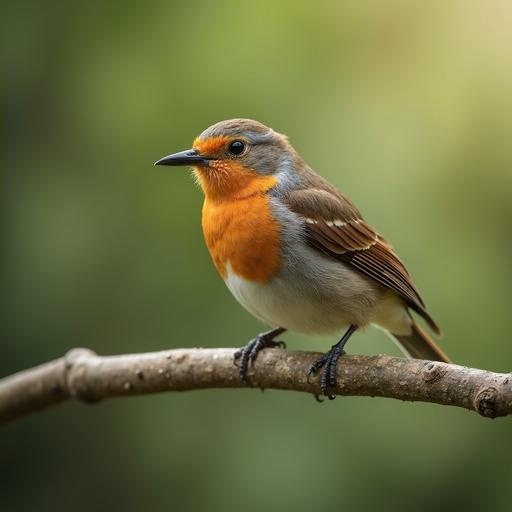}
			&\includegraphics[width=1.8cm]{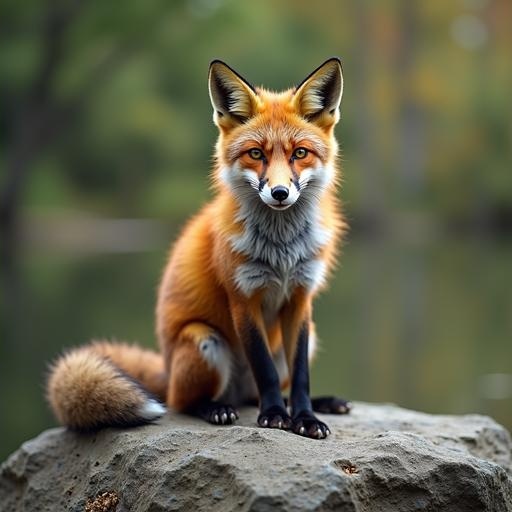}
			&\includegraphics[width=1.8cm]{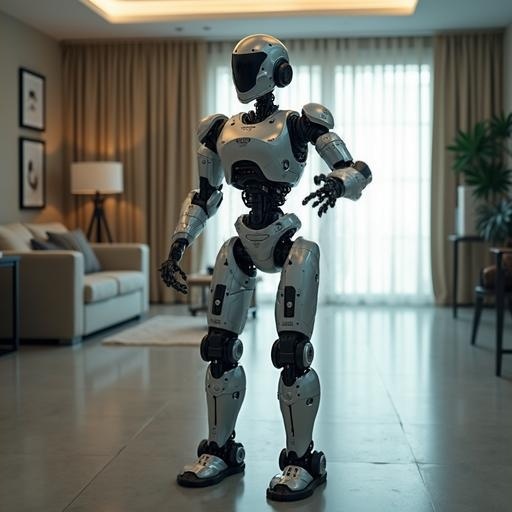}
			&\includegraphics[width=1.8cm]{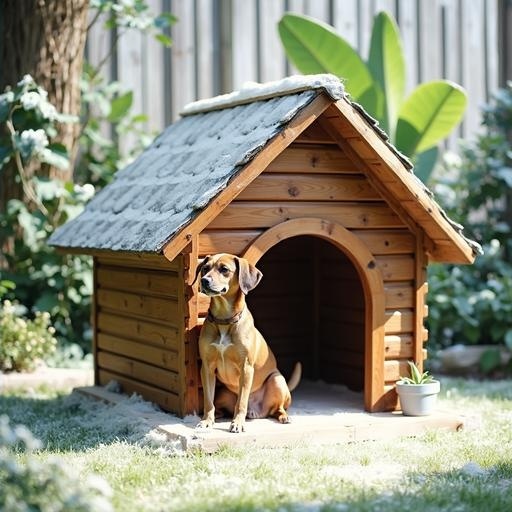}
			&\includegraphics[width=1.8cm]{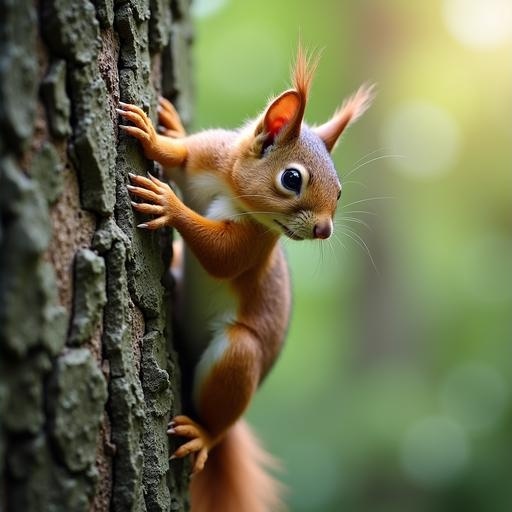}
			&\includegraphics[width=1.8cm]{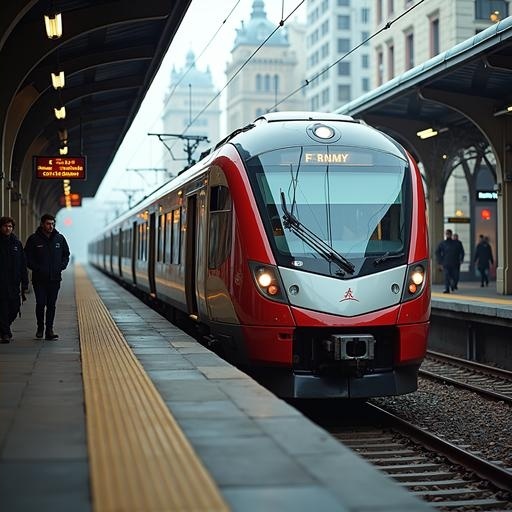}
			\\

			\raisebox{0.3cm}{\rotatebox[origin=c]{90}{\footnotesize{{TamingRF}}}}
			&\includegraphics[width=1.8cm]{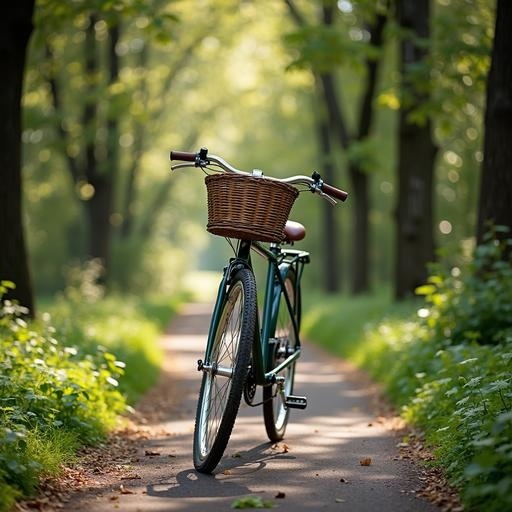}
			&\includegraphics[width=1.8cm]{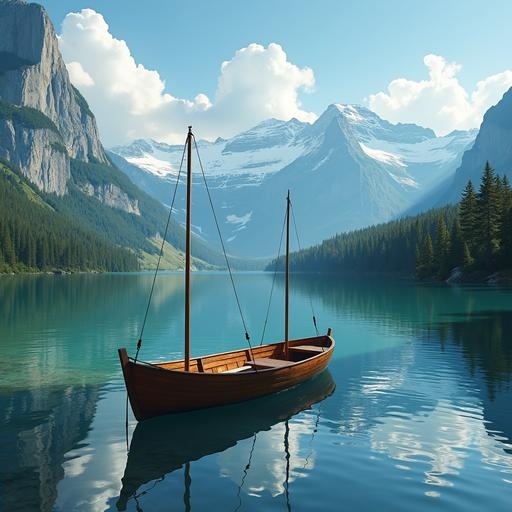}
			&\includegraphics[width=1.8cm]{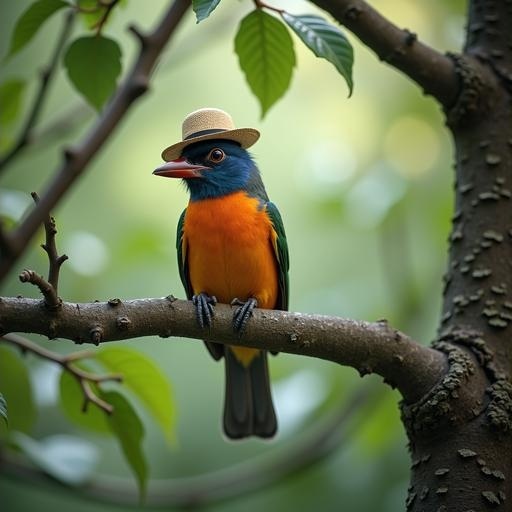}
			&\includegraphics[width=1.8cm]{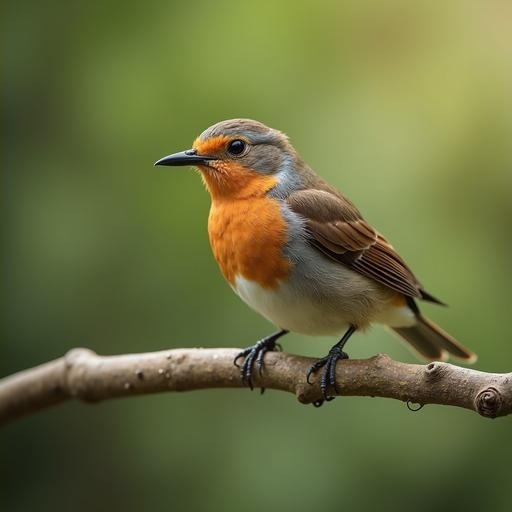}
			&\includegraphics[width=1.8cm]{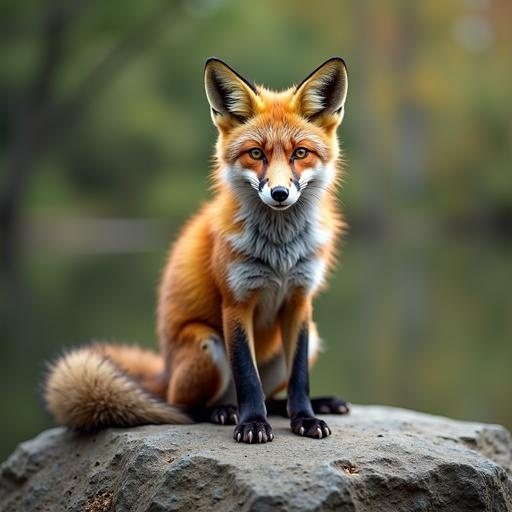}
			&\includegraphics[width=1.8cm]{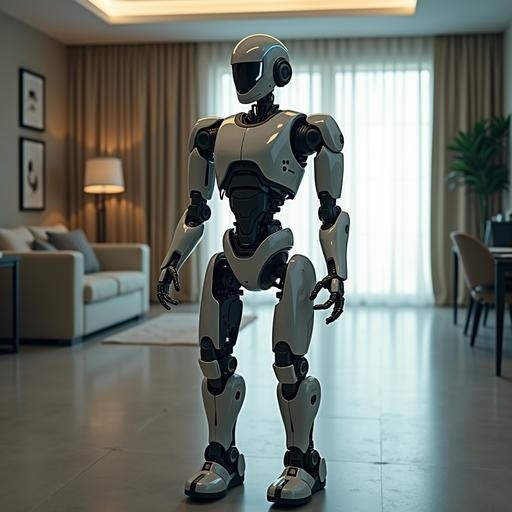}
			&\includegraphics[width=1.8cm]{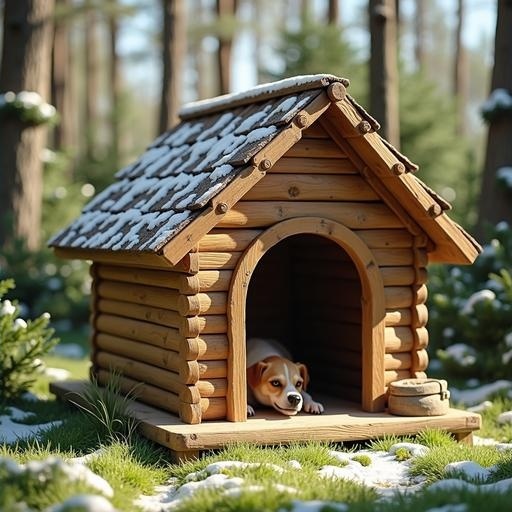}
			&\includegraphics[width=1.8cm]{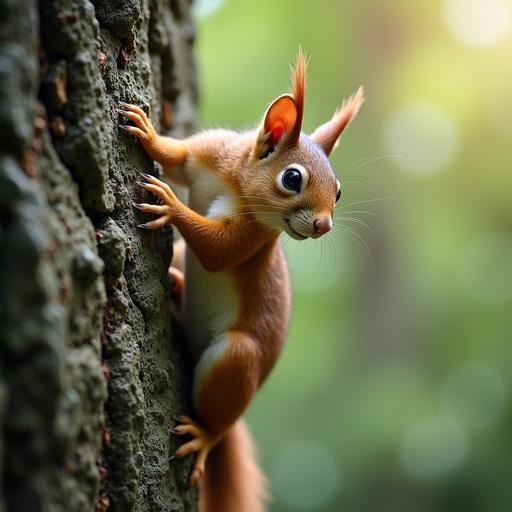}
			&\includegraphics[width=1.8cm]{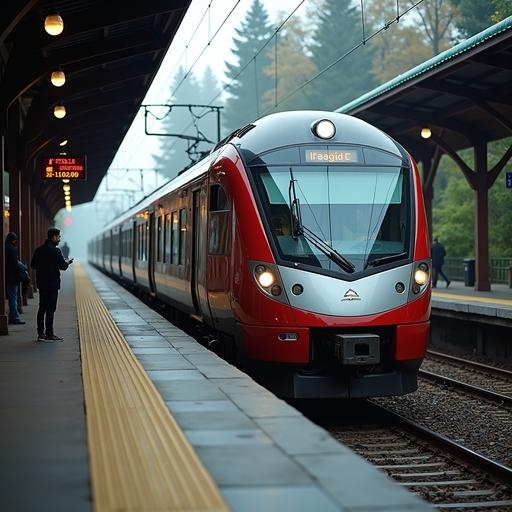}
			\\

			\raisebox{0.5cm}{\rotatebox[origin=c]{90}{\footnotesize{{MagicBrush}}}}
			&\includegraphics[width=1.8cm]{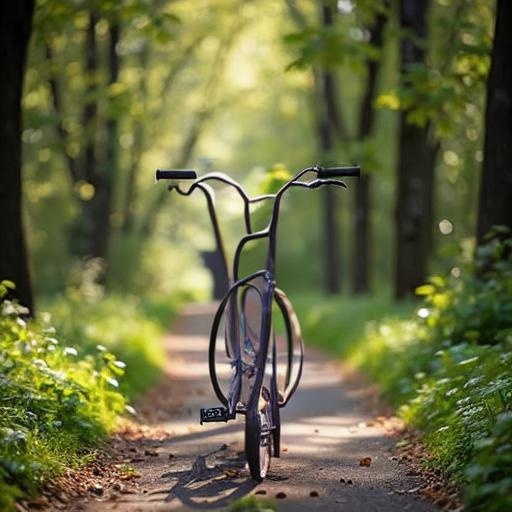}
			&\includegraphics[width=1.8cm]{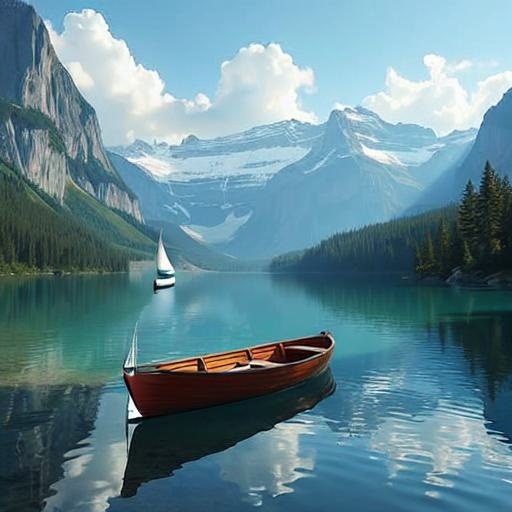}
			&\includegraphics[width=1.8cm]{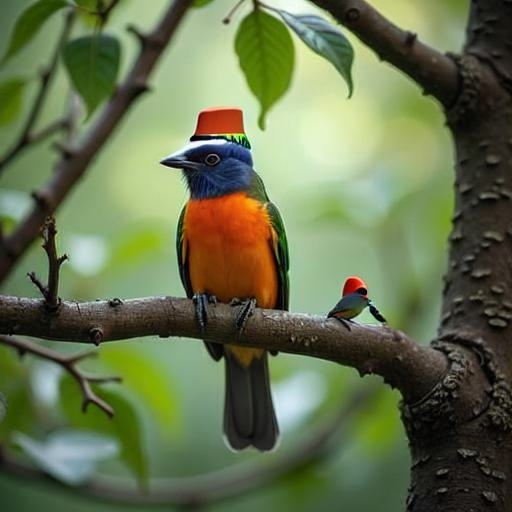}
			&\includegraphics[width=1.8cm]{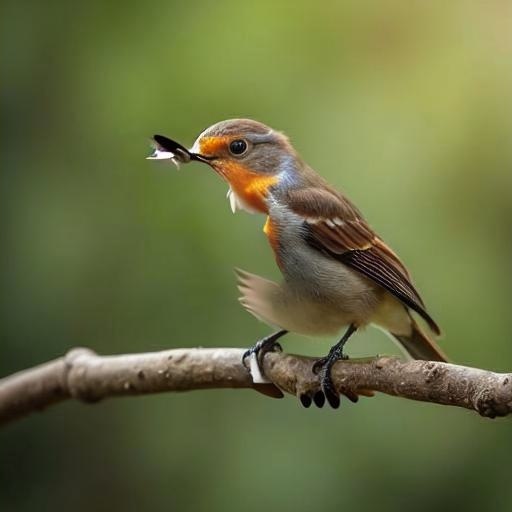}
			&\includegraphics[width=1.8cm]{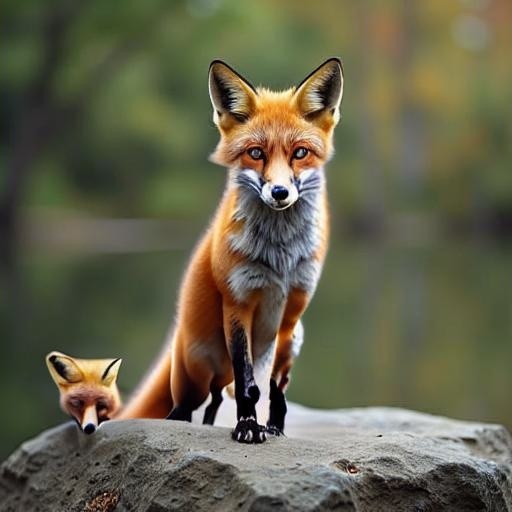}
			&\includegraphics[width=1.8cm]{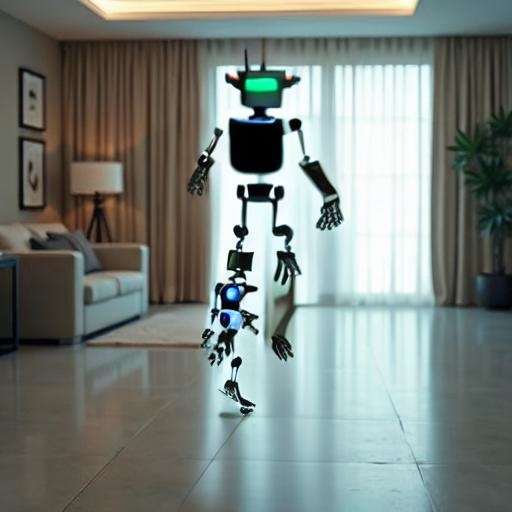}
			&\includegraphics[width=1.8cm]{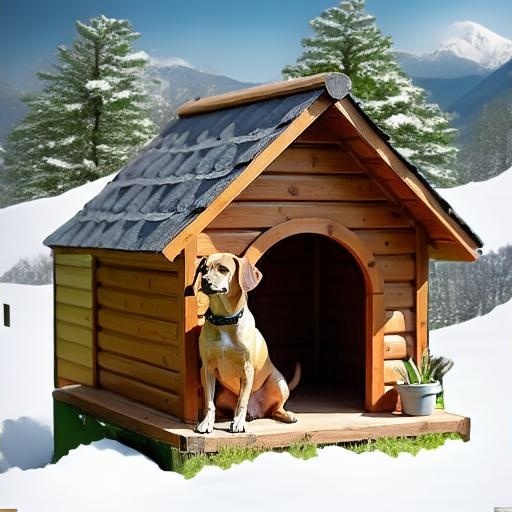}
			&\includegraphics[width=1.8cm]{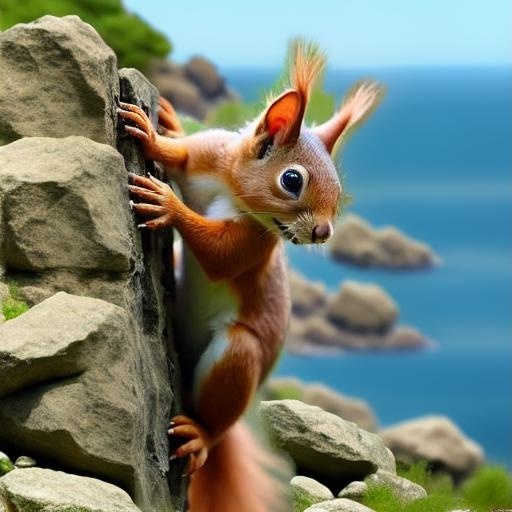}
			&\includegraphics[width=1.8cm]{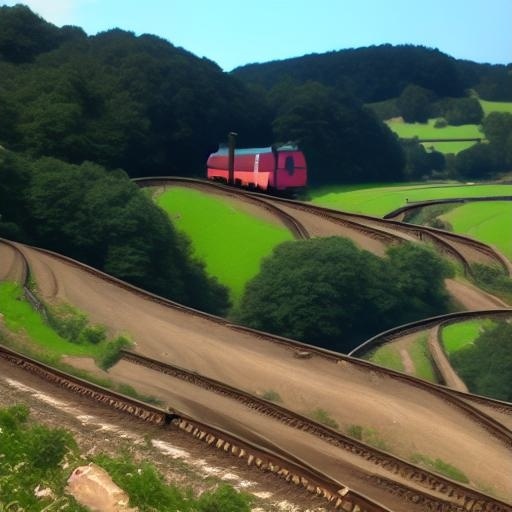}
			\\

			\raisebox{0.15cm}{\rotatebox[origin=c]{90}{\footnotesize{{OmniGen}}}}
			&\includegraphics[width=1.8cm]{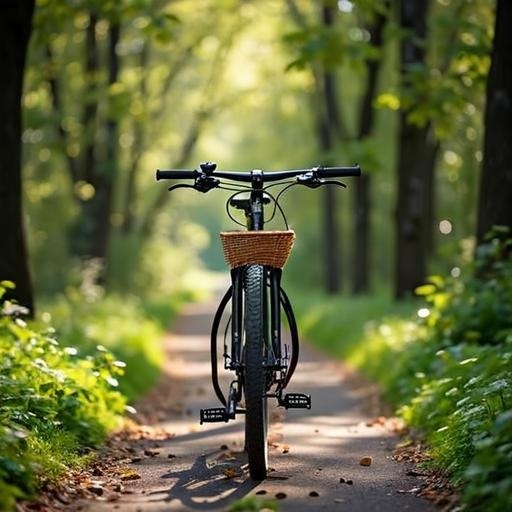}
			&\includegraphics[width=1.8cm]{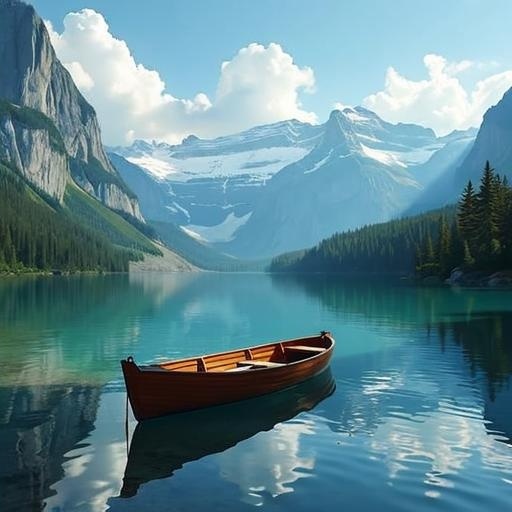}
			&\includegraphics[width=1.8cm]{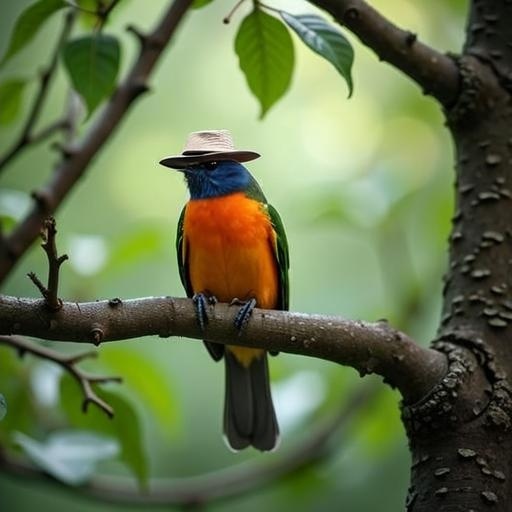}
			&\includegraphics[width=1.8cm]{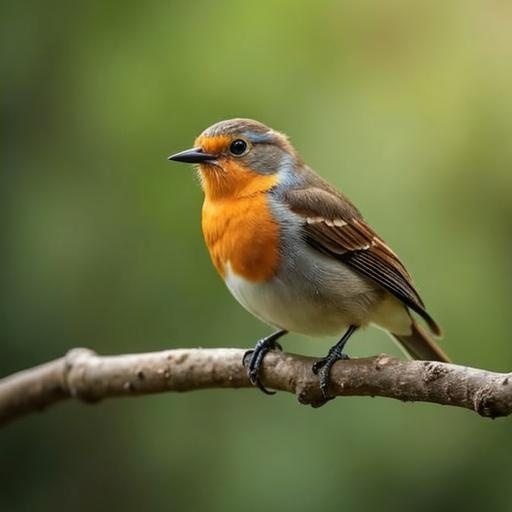}
			&\includegraphics[width=1.8cm]{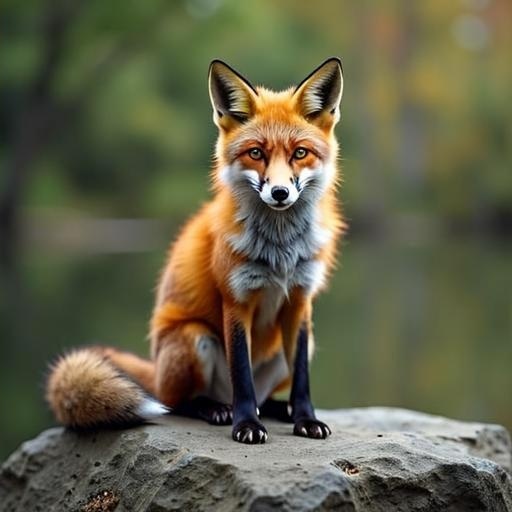}
			&\includegraphics[width=1.8cm]{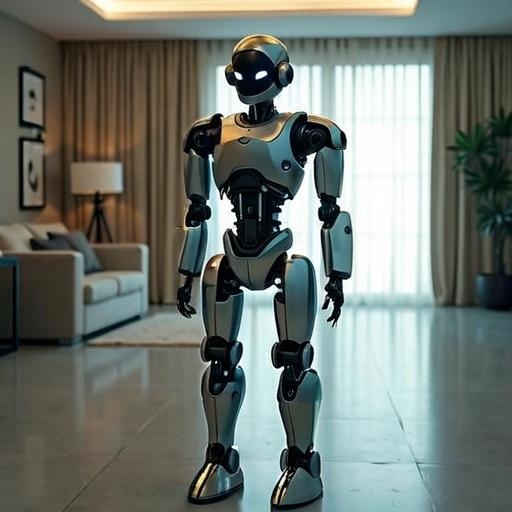}
			&\includegraphics[width=1.8cm]{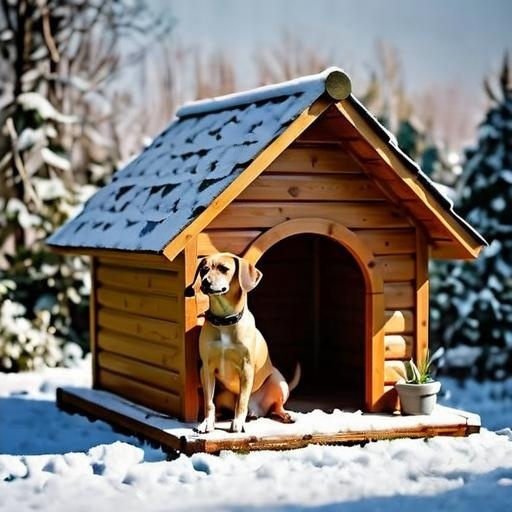}
			&\includegraphics[width=1.8cm]{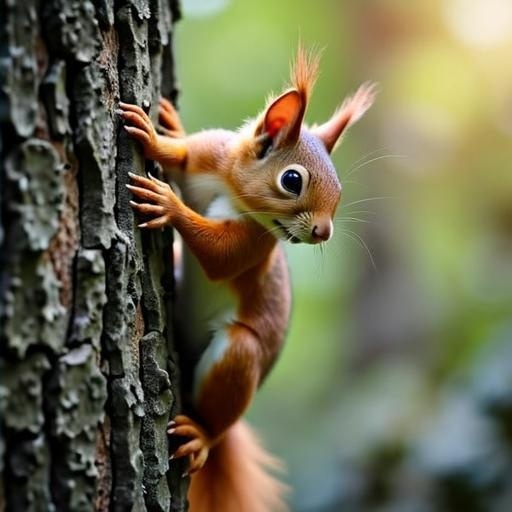}
			&\includegraphics[width=1.8cm]{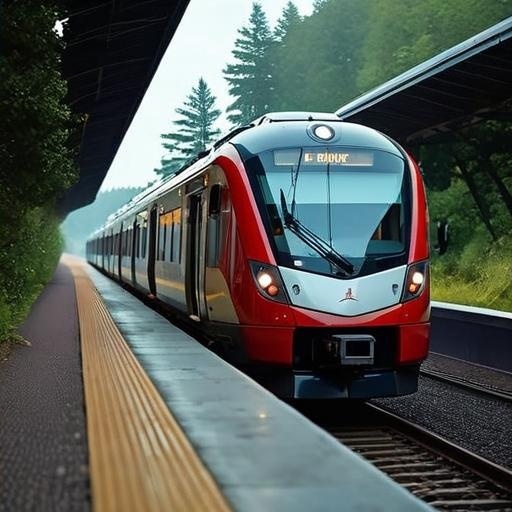}
			\\

			\raisebox{0.15cm}{\rotatebox[origin=c]{90}{\footnotesize{{Ours}}}}
			&\includegraphics[width=1.8cm]{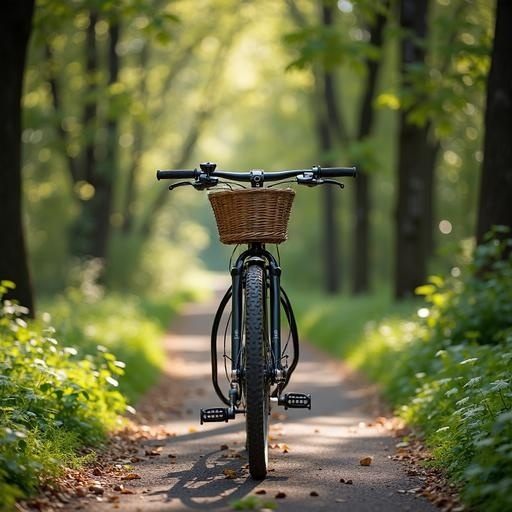}
			&\includegraphics[width=1.8cm]{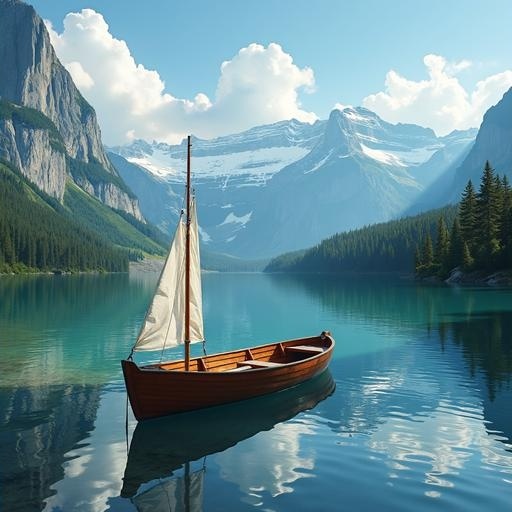}
			&\includegraphics[width=1.8cm]{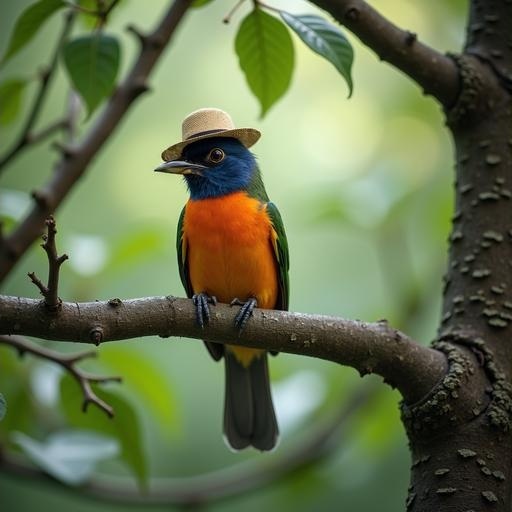}
			&\includegraphics[width=1.8cm]{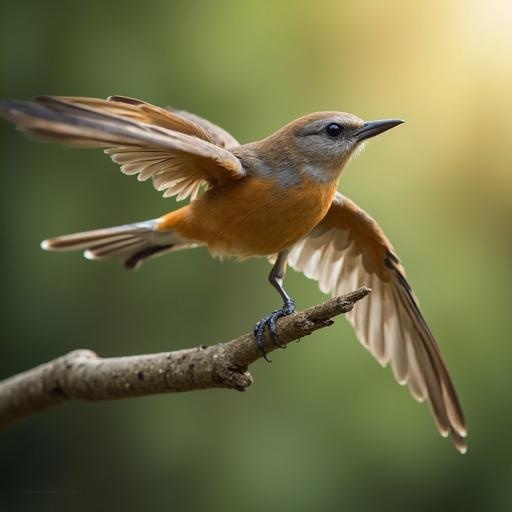}
			&\includegraphics[width=1.8cm]{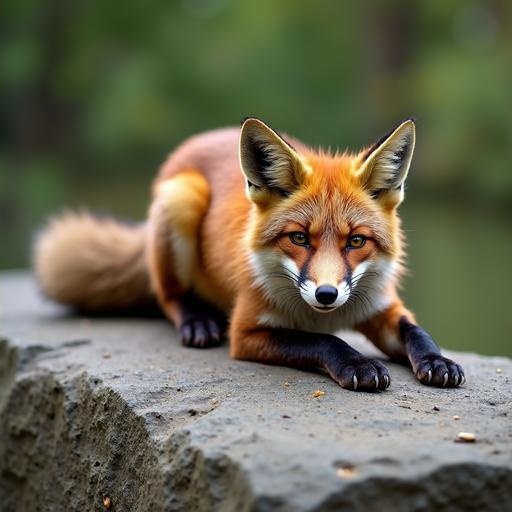}
			&\includegraphics[width=1.8cm]{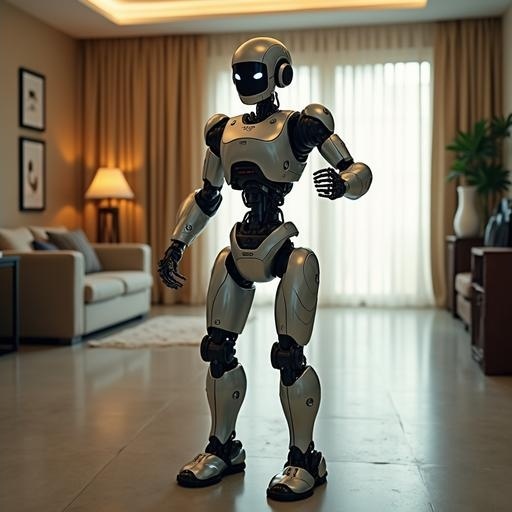}
			&\includegraphics[width=1.8cm]{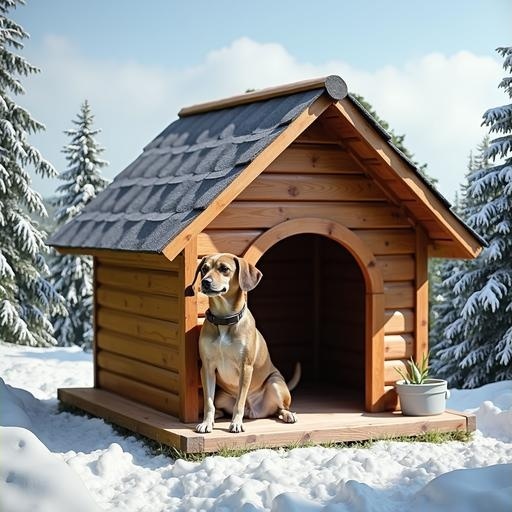}
			&\includegraphics[width=1.8cm]{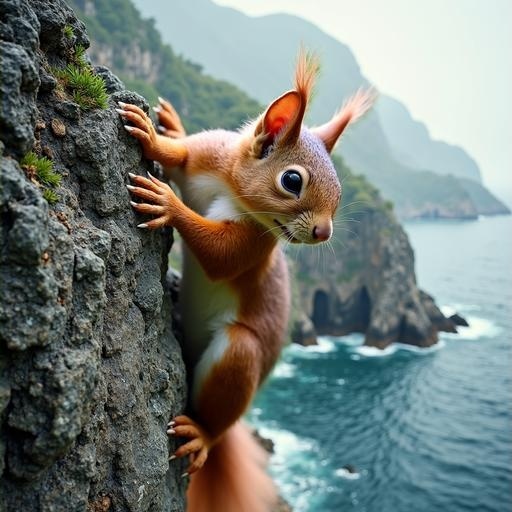}
			&\includegraphics[width=1.8cm]{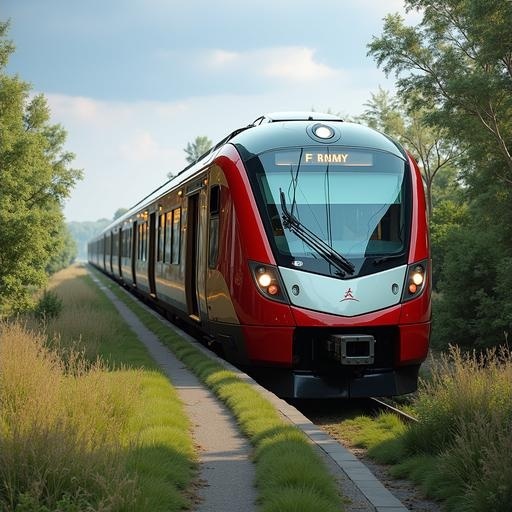}
			\\
			 & \scriptsize{\textit{``\textcolor{myblue}{Add a Basket}"}} & \scriptsize{\textit{``\textcolor{myblue}{Add a Sail}"}}& \scriptsize{\textit{``\textcolor{myblue}{Add a Hat}"}}& \scriptsize{\textit{``\textcolor{myblue}{Flying}"}}& \scriptsize{\textit{``\textcolor{myblue}{Stretching}"}}& \scriptsize{\textit{``\textcolor{myblue}{Dancing}"}}& \scriptsize{\textit{``\textcolor{myblue}{Snowy Hill}"}}& \scriptsize{\textit{``\textcolor{myblue}{Rocky Cliff}"}}& \scriptsize{\textit{``\textcolor{myblue}{Village Track}"}}
			\\
			
		\end{tabular}
	\end{center}
        \vspace{-1.9em}
	\caption{Qualitative comparison with training-free methods StableFlow~\cite{avrahami2024stable} and TamingRF~\cite{wang2024taming}, as well as general image editing models MagicBrush~\cite{zhang2023magicbrush} and OmniGen~\cite{xiao2024omnigen}. Our method achieves high-quality editing results while effectively preserving irrelevant regions.}
        \vspace{-1.0em}
	\label{fig:quail_comparsion}
\end{figure*}

Only minor modifications to Equation~\ref{eq:bg_replace} are needed to generalize value replacement to object moving and outpainting tasks. For object moving, it can be decomposed into a combination of region preservation and inpainting. Region preservation is achieved through value replacement, where the value of the foreground object before movement is used to replace the value at the corresponding position after movement, while irrelevant regions retain their own values. The missing area left by the original foreground object is then inpainted based on the editing text. For outpainting, the value of the low-resolution image is replaced in the target region of the larger image, while the remaining areas are generated according to the editing text. The algorithms for each editing task are provided in supplementary material.
\section{Experiments}
\label{sec:exp}

\noindent\textbf{Implementation Details.} We deploy our approach on FLUX.1-dev ($12$B). Following official recommendations, the guideline scale is set to $3.5$, and the number of denoising steps is set to $50$ by default. According to the quantitative results of position dependence in joint self-attention layers shown in Figure~\ref{fig:layer_dep_scatter}, we set the most position-relevant layers $\mathds{P}$ for the object addition task to $[1,2,4,26,30,54,55]$, applied across all denoising steps. For non-rigid editing, the more content-similarity-dependent layers $\mathds{C}$ are set to $[0,7,8,9,10,18,25,28,37,42,45,50,56]$, also applied across all denoising steps. For region-preserved editing (e.g., background replacement), value replacement is applied to all layers until the $45$th denoising steps.

\begin{table*}[t]
	\centering
	\small
	\setlength{\tabcolsep}{1.9pt}{
		\begin{tabular}{l|cccc|cccc|cccc}
			\hline
			\multicolumn{1}{ c }{{}}	& \multicolumn{4}{ c }{{\small{Object Addition}}} & \multicolumn{4}{ c }{{\small{Non-Rigid Editing}}} & \multicolumn{4}{ c }{{\small{Background Replacement}}} \\
			\hline
			\small{Methods} & \small{$\text{CLIP}_{img}$ $\uparrow$} & \small{$\text{CLIP}_{txt}$ $\uparrow$} & \small{$\text{CLIP}_{dir}$ $\uparrow$} & \small{PR $\uparrow$} & \small{$\text{CLIP}_{img}$ $\downarrow$} & \small{$\text{CLIP}_{txt}$ $\uparrow$} & \small{$\text{CLIP}_{dir}$ $\uparrow$} & \small{PR $\uparrow$} & \small{PSNR $\uparrow$} & \small{$\text{CLIP}_{txt}$ $\uparrow$} & \small{$\text{CLIP}_{dir}$ $\uparrow$} & \small{PR $\uparrow$} \\
			\hline
			StableFlow  & \small{0.964} & \small{0.319} & \small{0.173} & \small{12.2\%} & \small{0.969} & \small{0.307} & \small{0.124} & \small{11.1\%} & \small{17.14} & \small{0.283} & \small{0.123} & \small{1.4\%} \\
                TamingRF  & \small{0.958} & \small{0.320} & \small{0.175} & \small{31.9\%} & \small{0.961} & \small{0.308} & \small{0.120} & \small{13.2\%} & \small{16.32} & \small{0.284} & \small{0.134} & \small{1.9\%} \\
                MagicBrush  & \small{0.944} & \small{0.319} & \small{0.161} & \small{1.6\%} & \small{\textbf{0.933}} & \small{0.308} & \small{0.111} & \small{0.5\%} & \small{14.87} & \small{0.308} & \small{0.261} & \small{13.0\%} \\
                OmniGen  & \small{0.966} & \small{0.314} & \small{0.090} & \small{3.5\%} & \small{0.974} & \small{0.303} & \small{0.047} & \small{1.1\%} & \small{21.73} & \small{0.291} & \small{0.126} & \small{2.4\%} \\
                \rowcolor[HTML]{FFF2CC}
                Ours  & \textbf{\small{0.974}} & \textbf{\small{0.321}} & \textbf{\small{0.202}} & \textbf{\small{50.8\%}} & \small{0.940} & \textbf{\small{0.315}} & \textbf{\small{0.153}} & \textbf{\small{74.1\%}} & \textbf{\small{24.04}} & \textbf{\small{0.328}} & \textbf{\small{0.319}} & \textbf{\small{81.4\%}} \\

                \hline
		\end{tabular}
	}
        \vspace{-0.5em}
	\caption{Quantitative comparison with StableFlow~\cite{avrahami2024stable}, TamingRF~\cite{wang2024taming}, MagicBrush~\cite{zhang2023magicbrush} and OmniGen~\cite{xiao2024omnigen}. $\text{CLIP}_{img}$ measures the similarity between source image and edited image; $\text{CLIP}_{txt}$ measures the similarity between editing text and edited image; $\text{CLIP}_{dir}$ calculates the similarity between direction of text change and direction of image change, providing a more precise evaluation of editing effectiveness; for background editing, PSNR is computed on the foreground region before and after editing; PR represents the user preference rate.}
        \vspace{-1.0em}
	\label{tab:quant_comparsion}
\end{table*}

\subsection{Quantitative and Qualitative Comparison}
We compare our method with state-of-the-art general image editing approaches. Among them, StableFlow~\cite{avrahami2024stable} and TamingRF~\cite{wang2024taming} are training-free editing methods designed for FLUX, while MagicBrush~\cite{zhang2023magicbrush} and OmniGen~\cite{xiao2024omnigen} are pre-trained versatile image editing models. All methods are evaluated using their official implementations.

\begin{figure}[t]
	\begin{center}
		\setlength{\tabcolsep}{0.5pt}
		\begin{tabular}{m{0.3cm}<{\centering}m{1.55cm}<{\centering}|m{1.55cm}<{\centering}m{1.55cm}<{\centering}m{1.55cm}<{\centering}m{1.55cm}<{\centering}}
			& \scriptsize{Source Image} & \multicolumn{4}{ c }{{\scriptsize{Edited Image}}}
			\\

			\multirow{2}{*}{\raisebox{-0.75cm}{\rotatebox[origin=c]{90}{\footnotesize{{\textcolor{myblue}{Outpainting}}}}}}
			&\includegraphics[width=1.5cm]{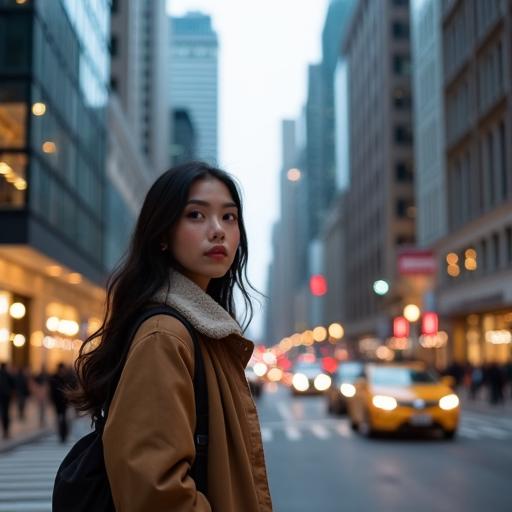}
			&\includegraphics[width=1.5cm]{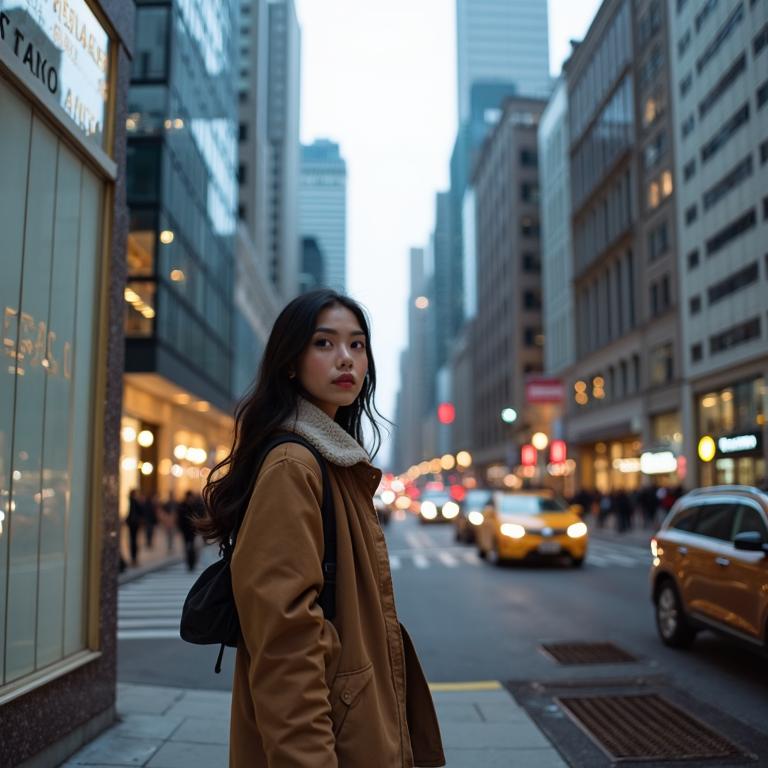}
			&\includegraphics[width=1.5cm]{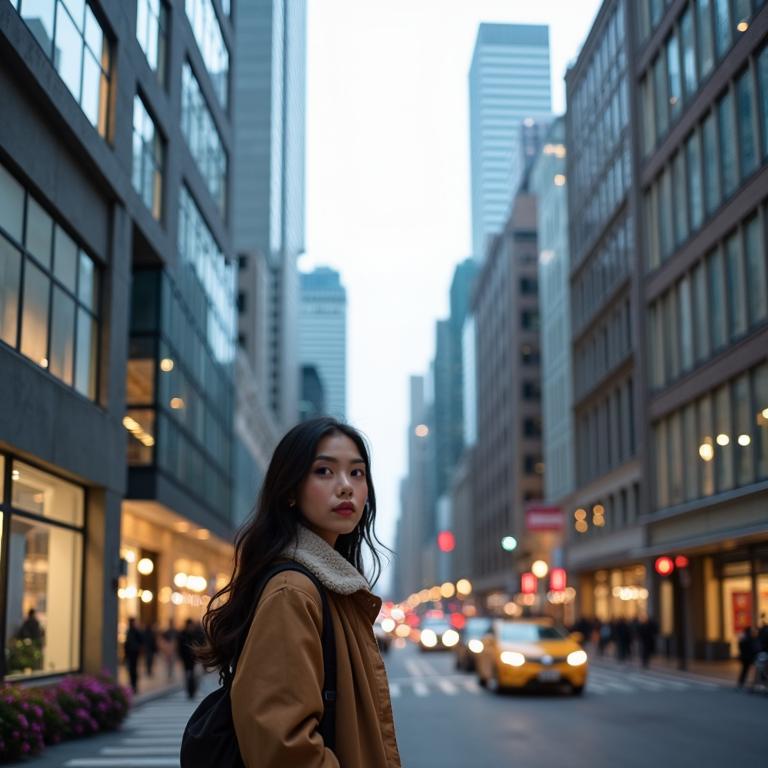}
			&\includegraphics[width=1.5cm]{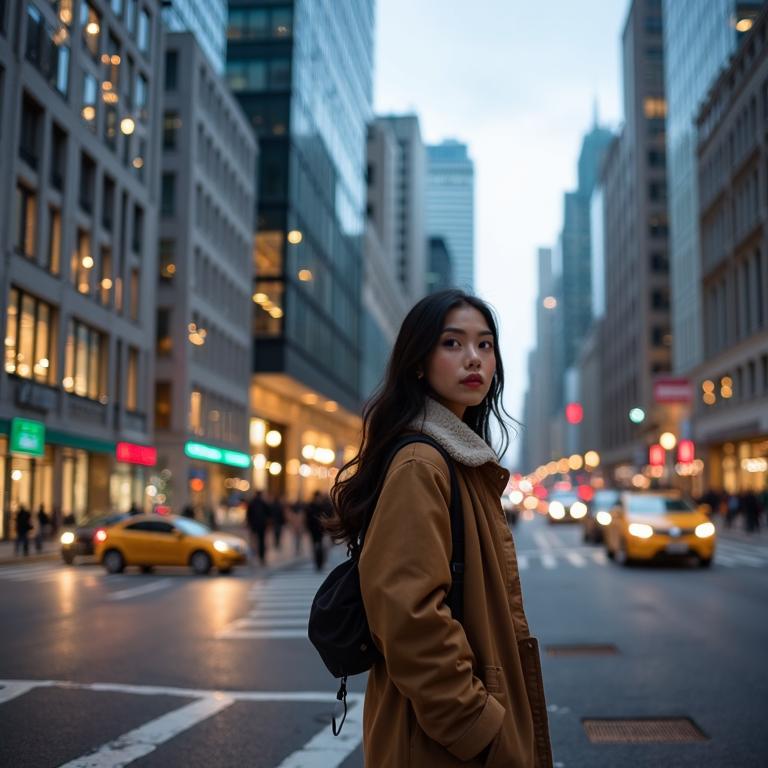}
			&\includegraphics[width=1.5cm]{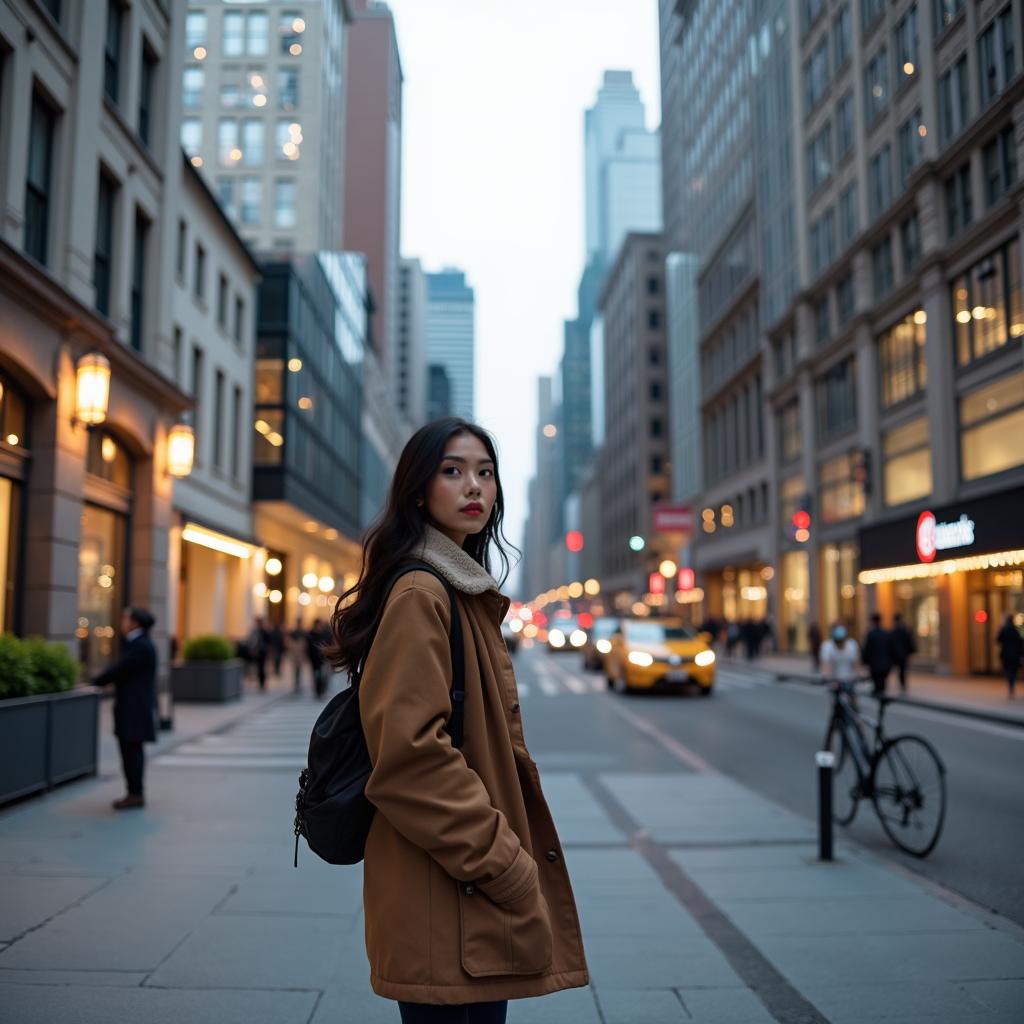}
			\\
			&\includegraphics[width=1.5cm]{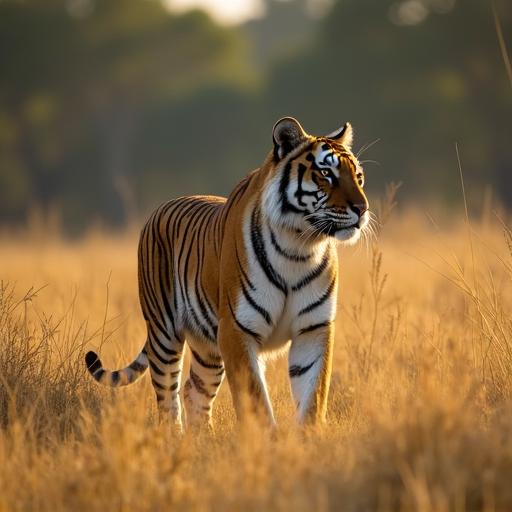}
			&\includegraphics[width=1.5cm]{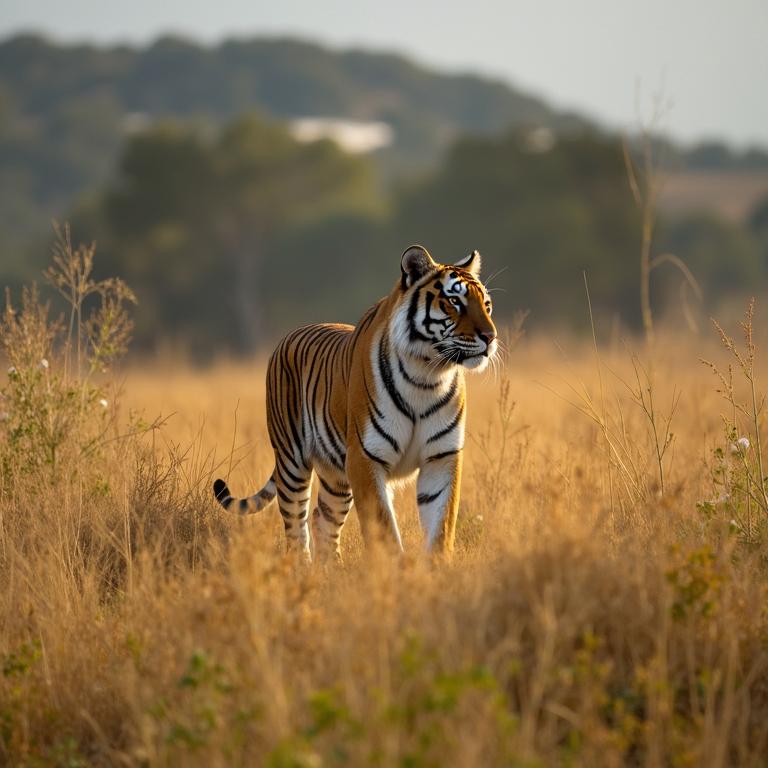}
			&\includegraphics[width=1.5cm]{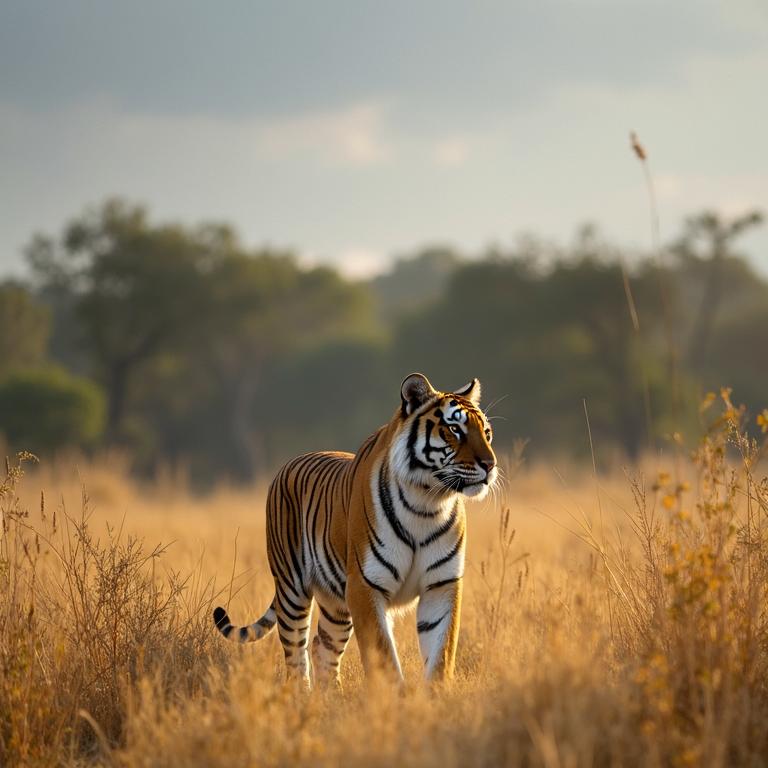}
			&\includegraphics[width=1.5cm]{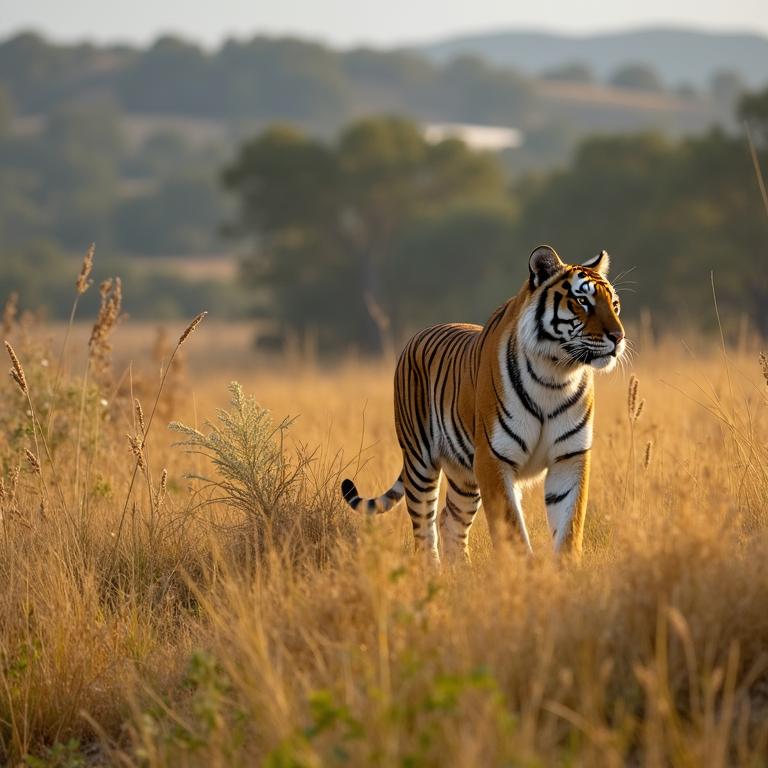}
			&\includegraphics[width=1.5cm]{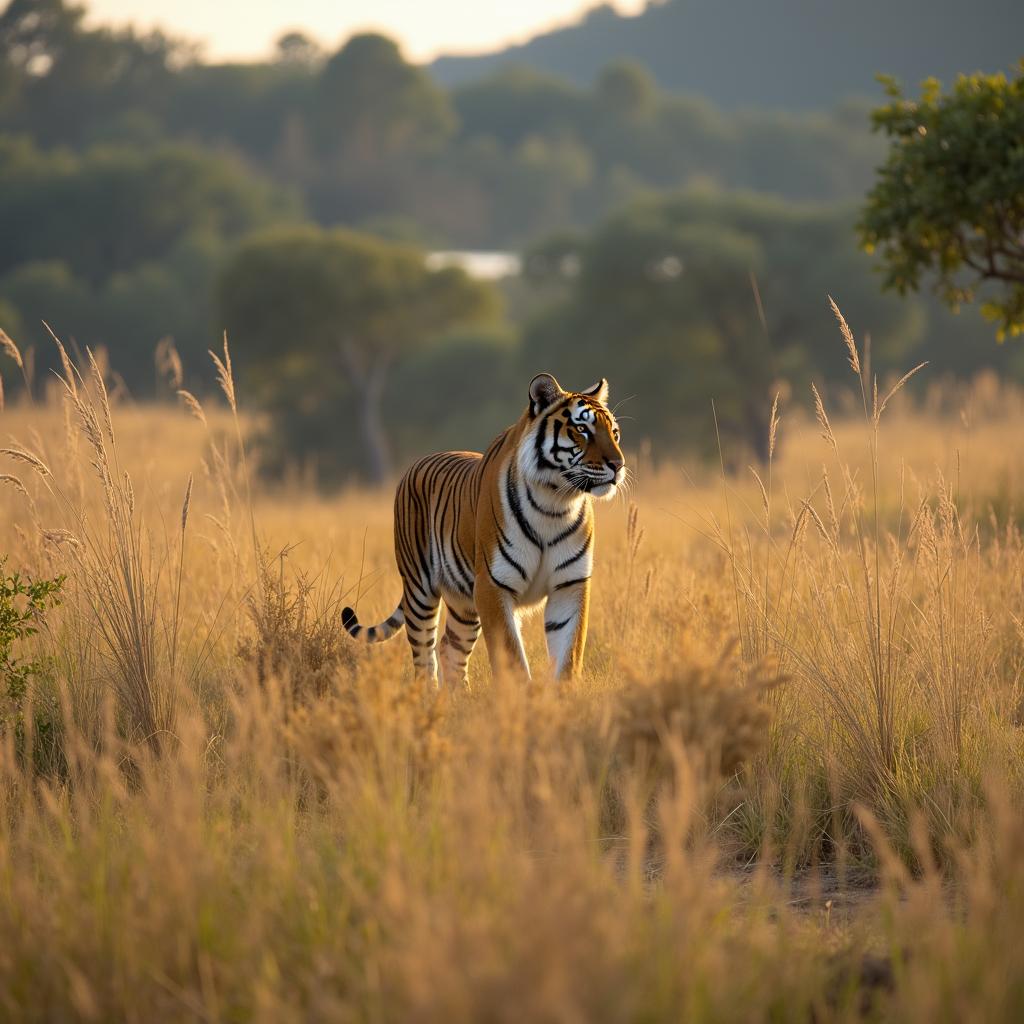}	
			\\

                \cdashline{1-6}
                \noalign{\vskip 0.15cm}
            
			\multirow{2}{*}{\raisebox{-0.75cm}{\rotatebox[origin=c]{90}{\footnotesize{{\textcolor{myblue}{Object Movement}}}}}}
			&\includegraphics[width=1.5cm]{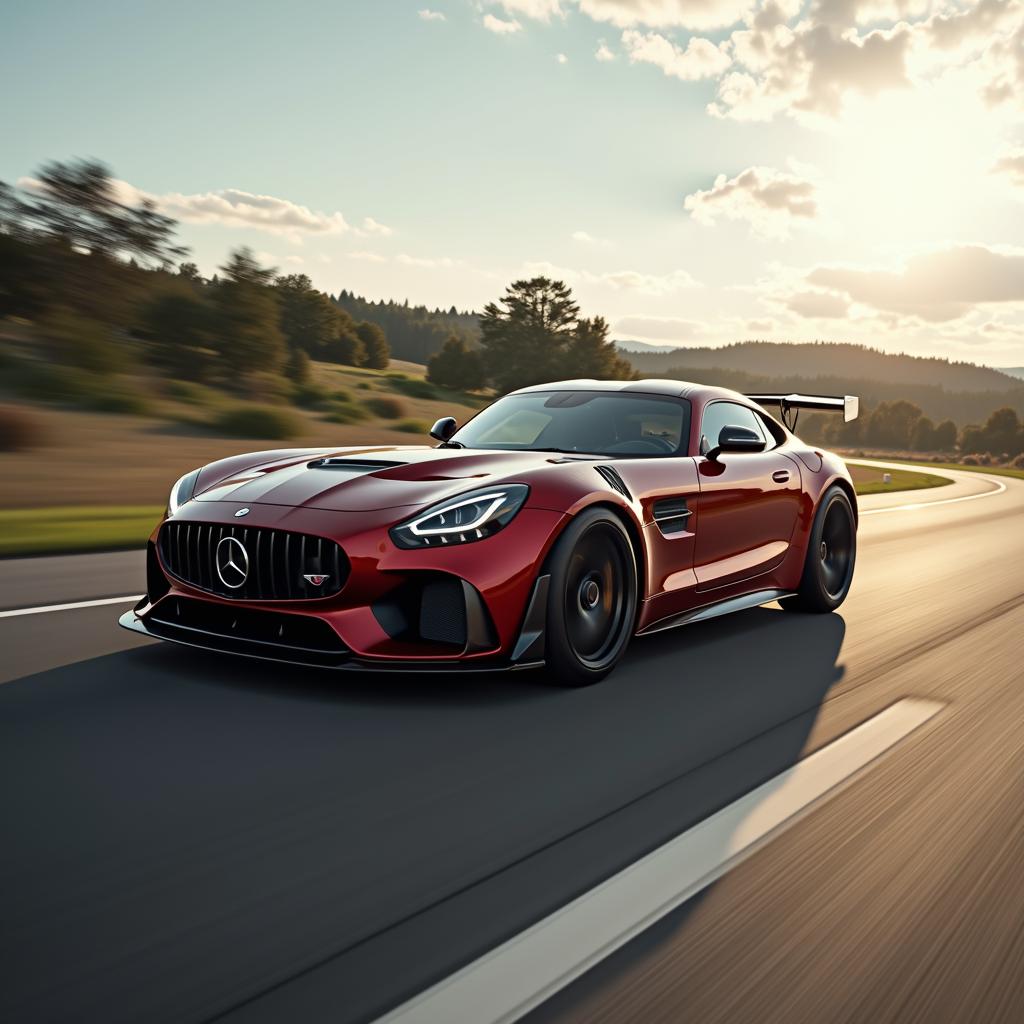}
			&\includegraphics[width=1.5cm]{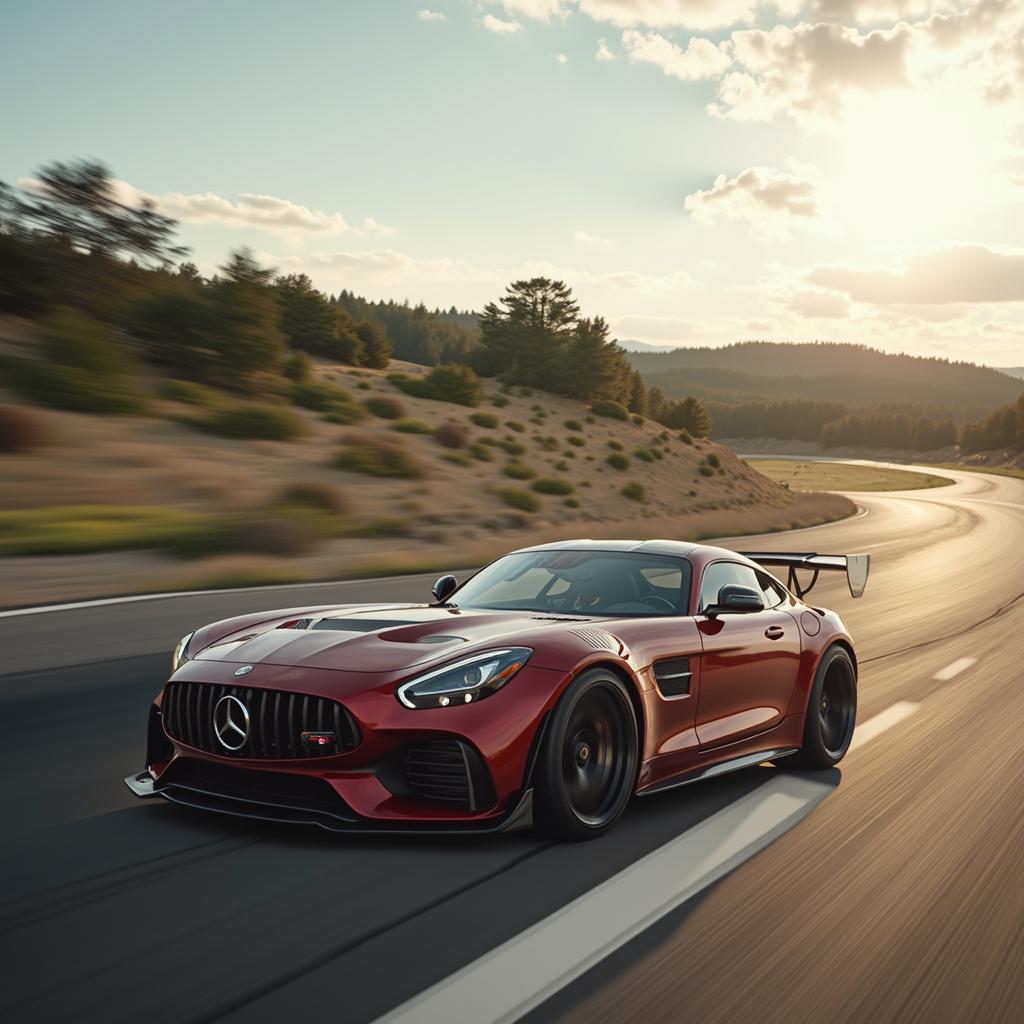}
			&\includegraphics[width=1.5cm]{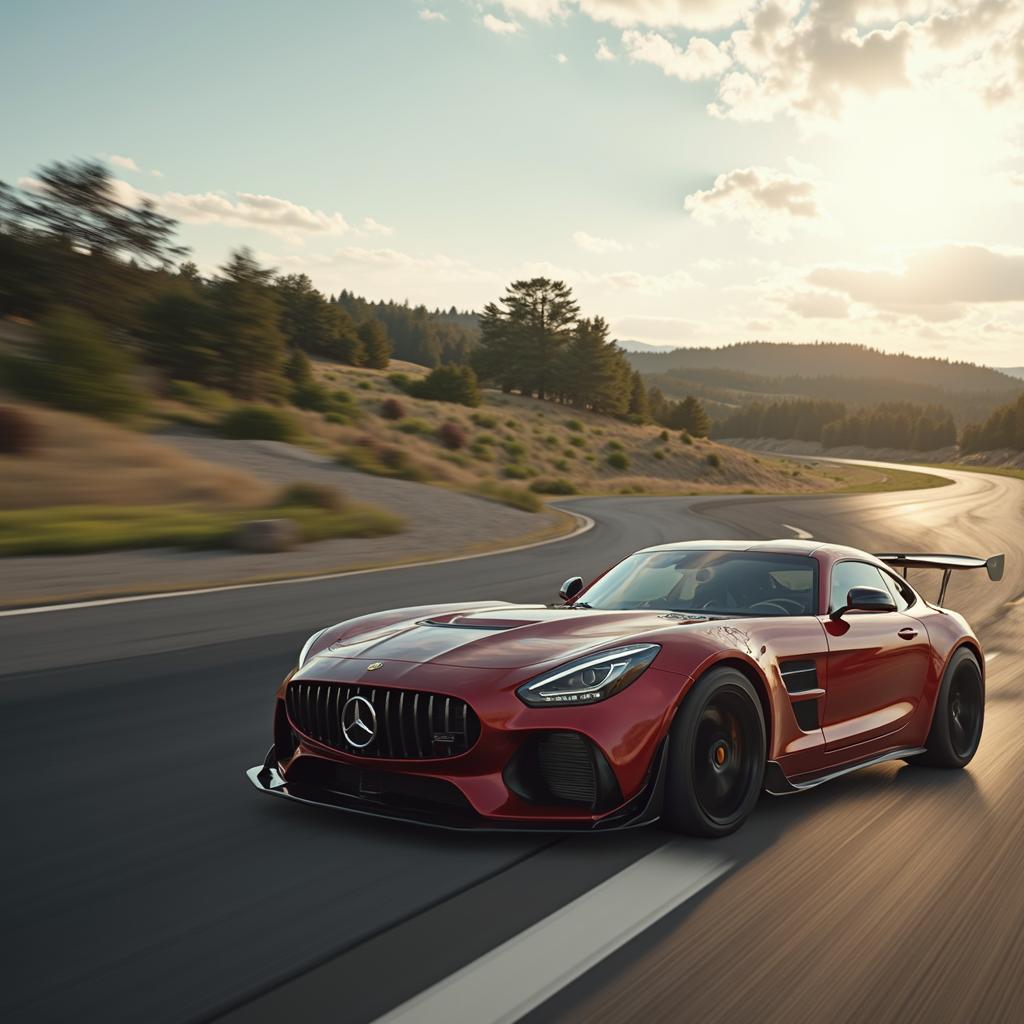}
			&\includegraphics[width=1.5cm]{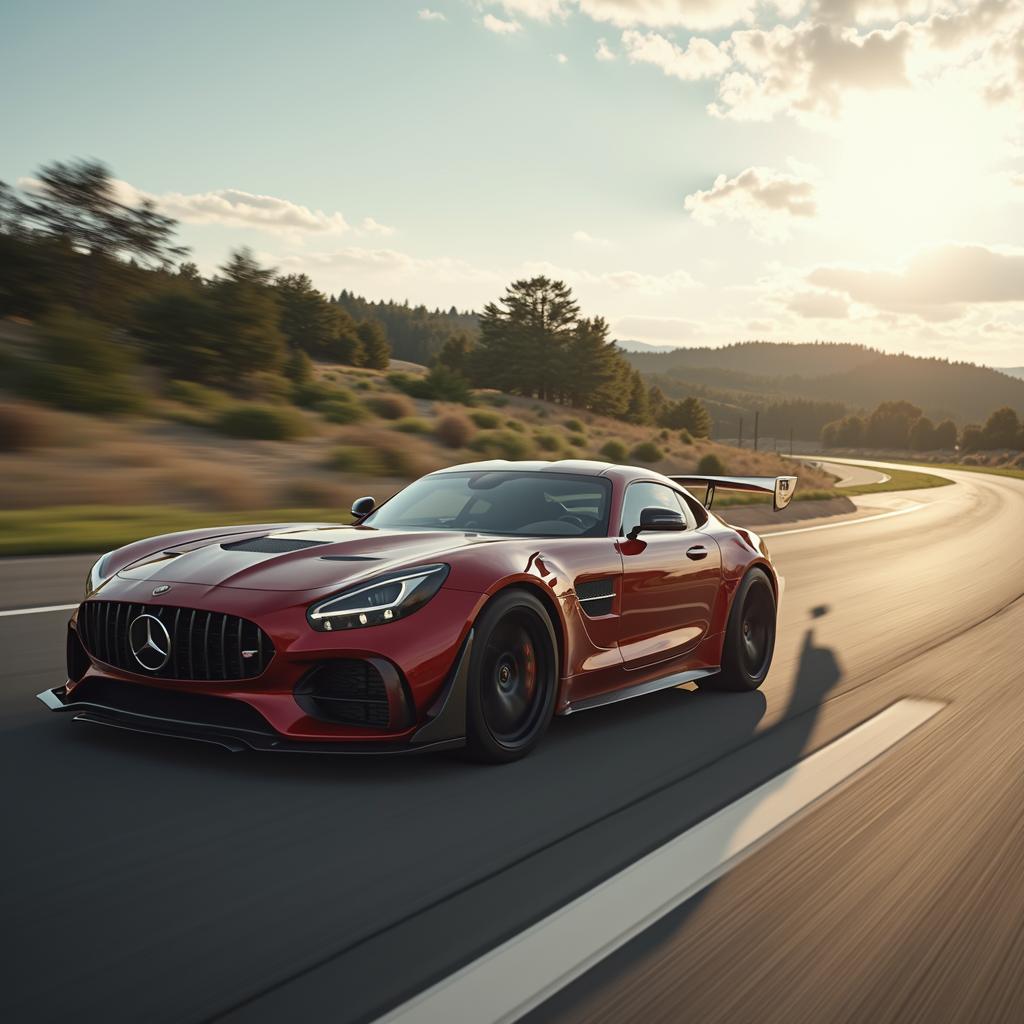}
			&\includegraphics[width=1.5cm]{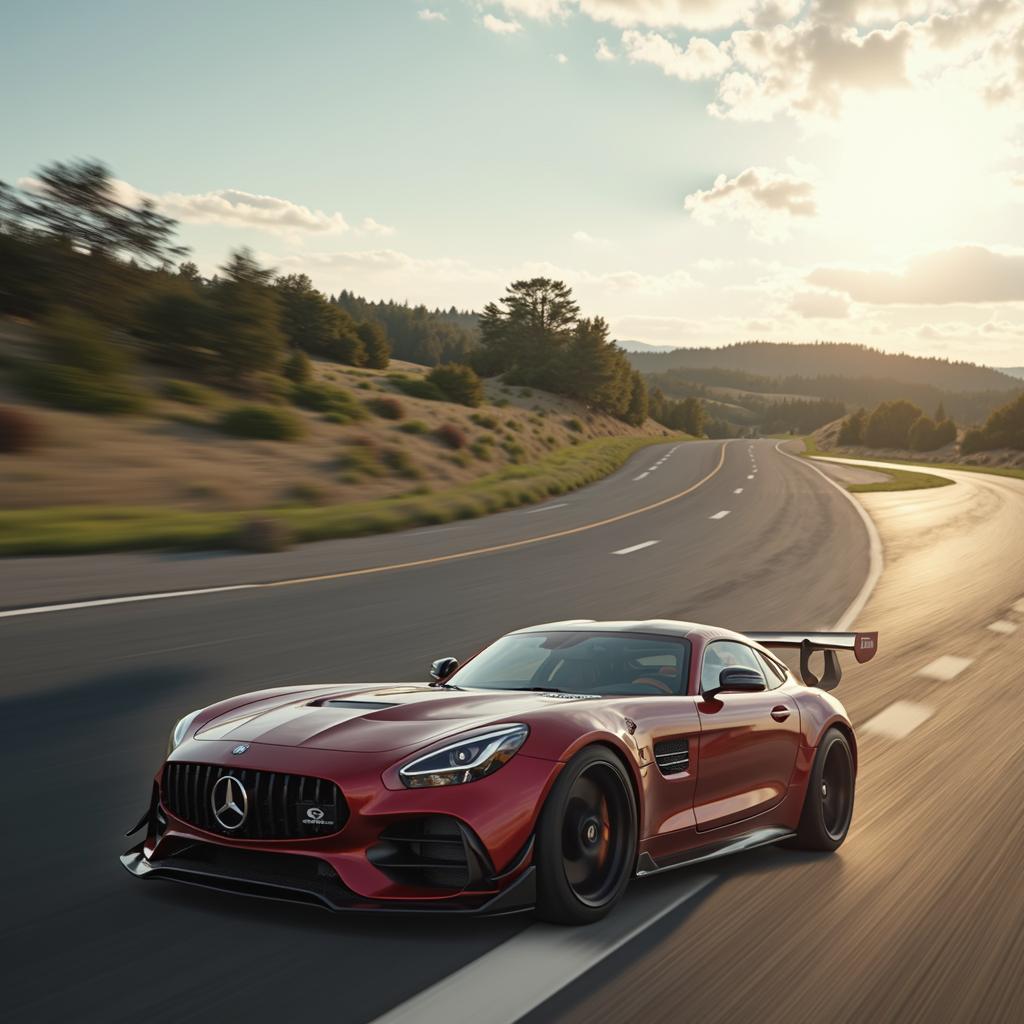}
			\\
			&\includegraphics[width=1.5cm]{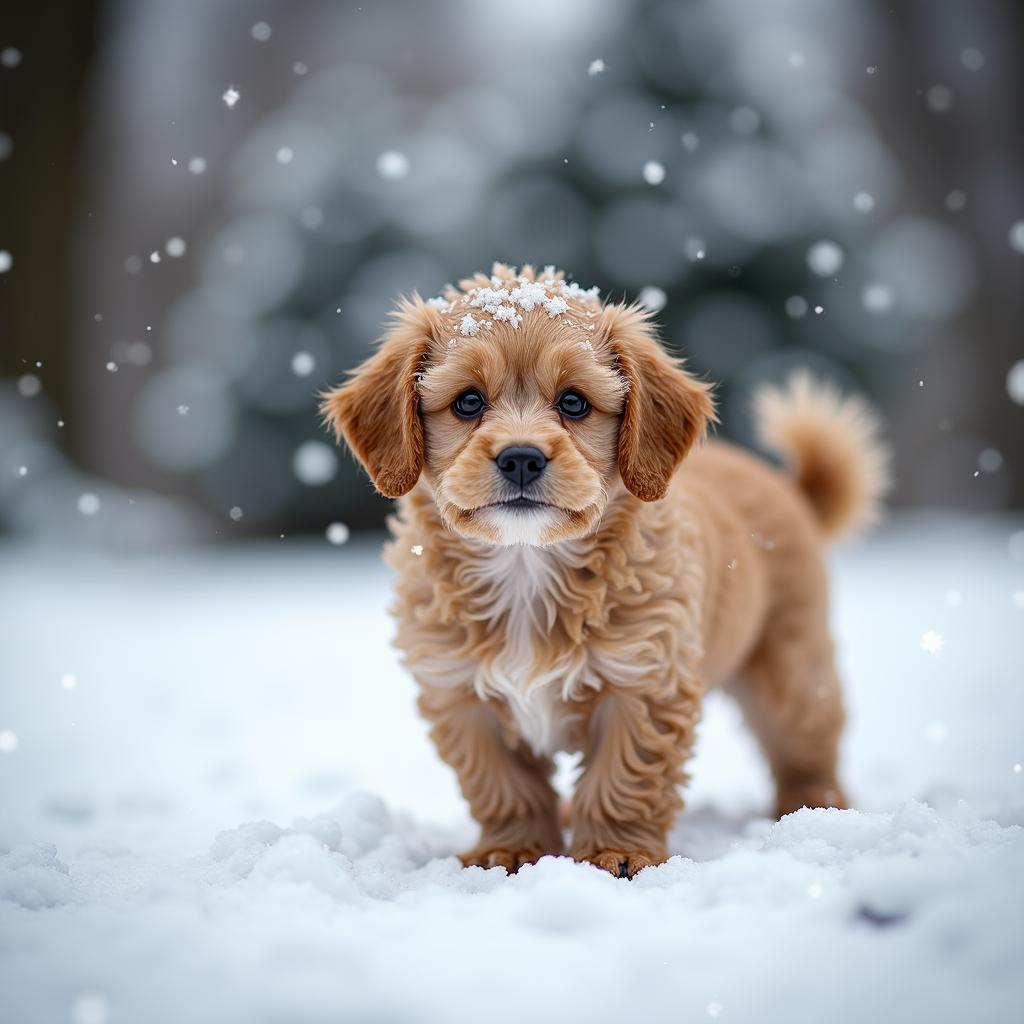}
			&\includegraphics[width=1.5cm]{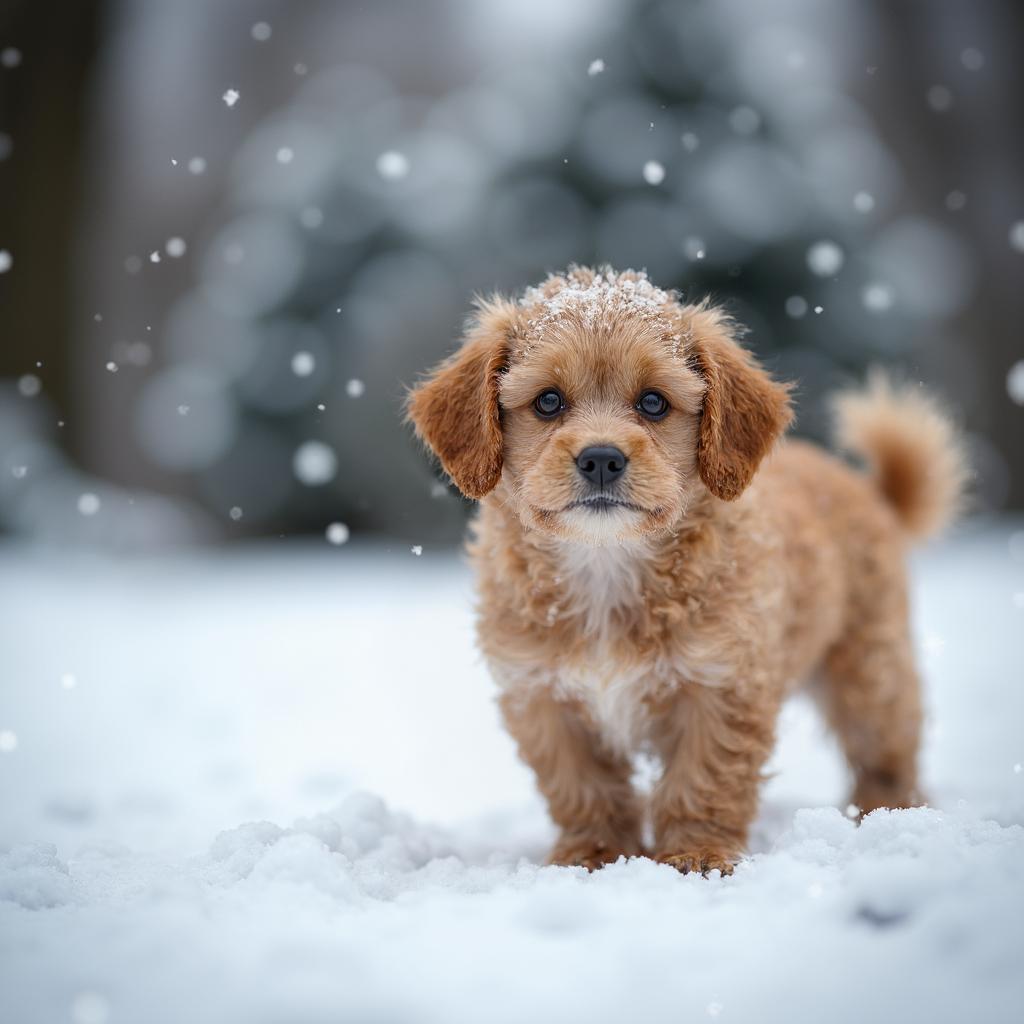}
			&\includegraphics[width=1.5cm]{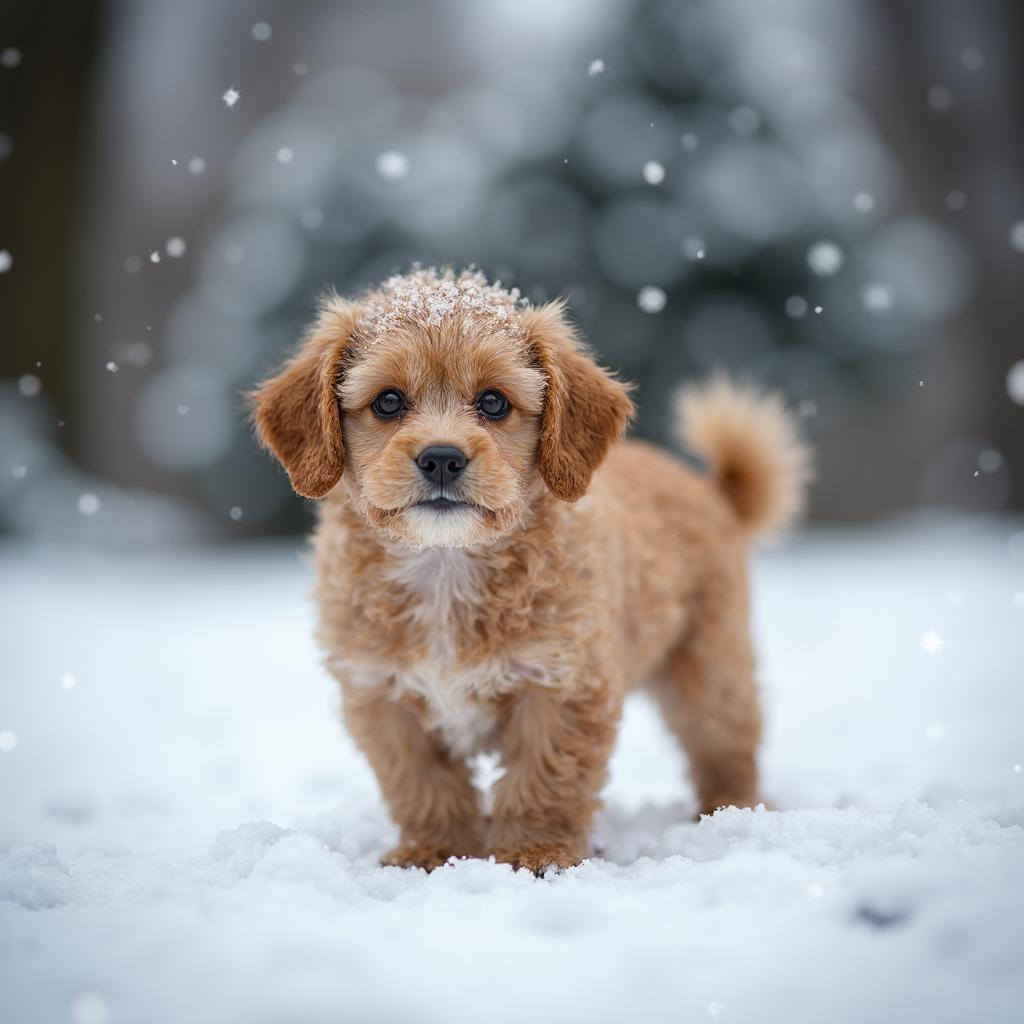}
			&\includegraphics[width=1.5cm]{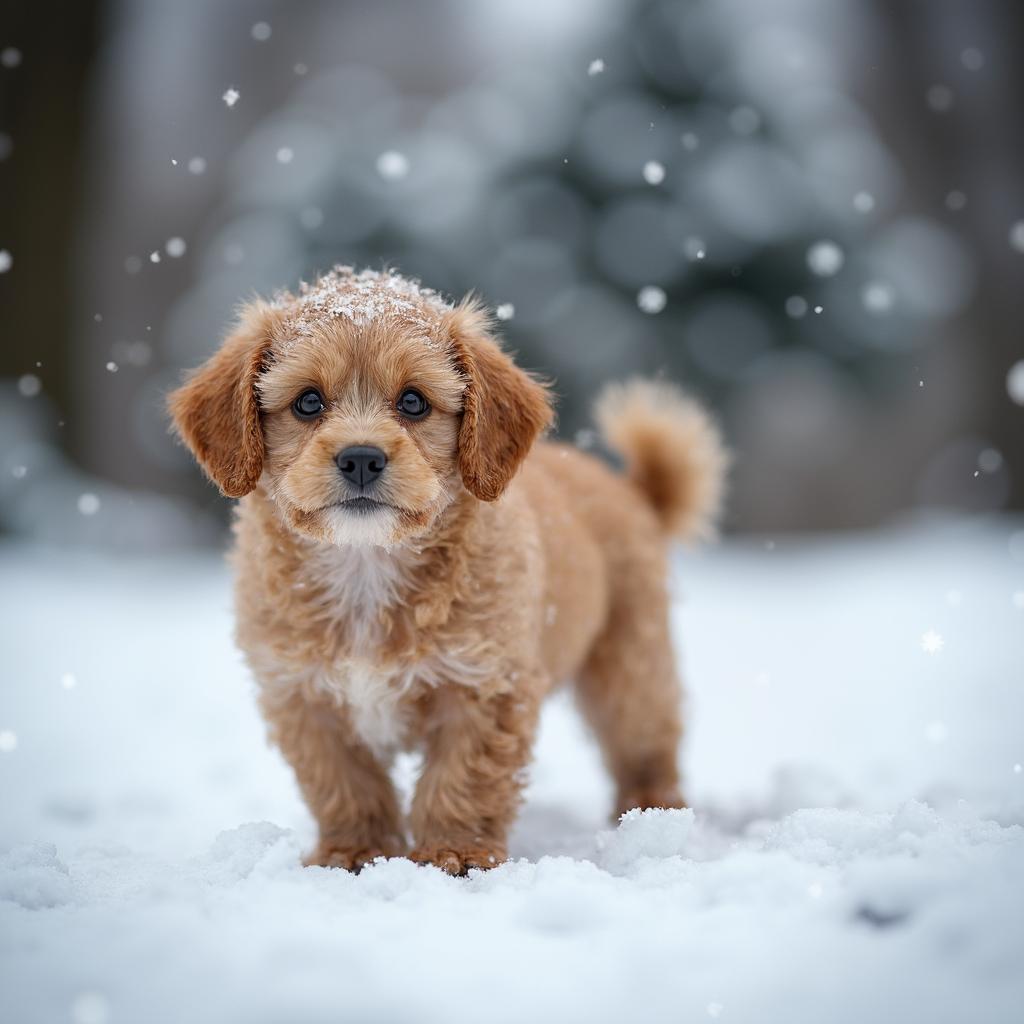}
			&\includegraphics[width=1.5cm]{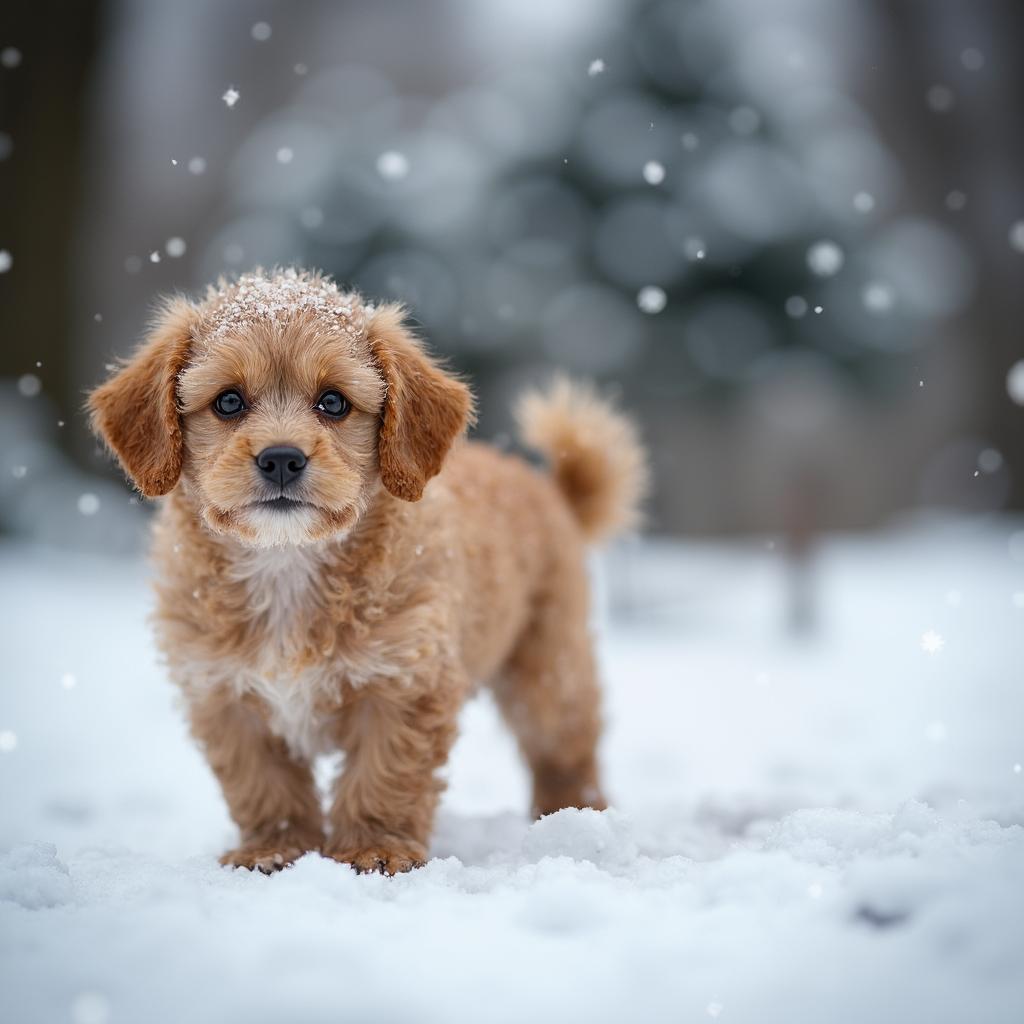}
			\\
			
		\end{tabular}
	\end{center}
        \vspace{-1.5em}
	\caption{Visual results of our method on region-preserved editing tasks such as object movement and outpainting.} 
	\label{fig:move_outpainting}
        \vspace{-1.0em}
\end{figure}

\begin{figure}[t]
	\begin{center}
		\setlength{\tabcolsep}{0.5pt}
		\begin{tabular}{m{0.3cm}<{\centering}m{1.55cm}<{\centering}m{1.55cm}<{\centering}m{1.55cm}<{\centering}m{1.55cm}<{\centering}m{1.55cm}<{\centering}}
			& \footnotesize{{\textcolor{myblue}{\textit{`hat'}}}} & \footnotesize{{\textcolor{myblue}{\textit{`orange'}}}} & \footnotesize{{\textcolor{myblue}{\textit{`running'}}}} & \footnotesize{{\textcolor{myblue}{\textit{`beach'}}}} & \footnotesize{{\textcolor{myblue}{\textit{`sunrise'}}}}
			\\

			\raisebox{0.15cm}{\rotatebox[origin=c]{90}{\scriptsize{Source Image}}}
			&\includegraphics[width=1.5cm]{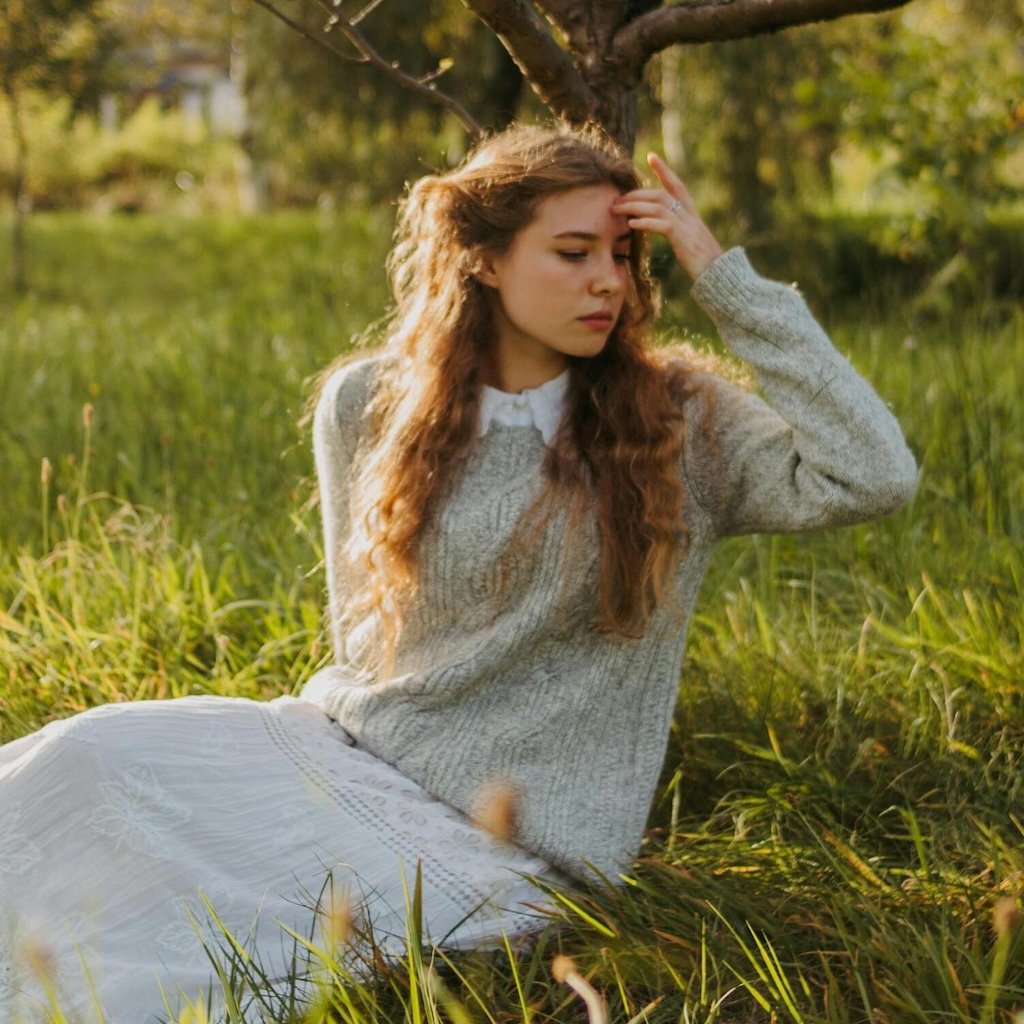}
			&\includegraphics[width=1.5cm]{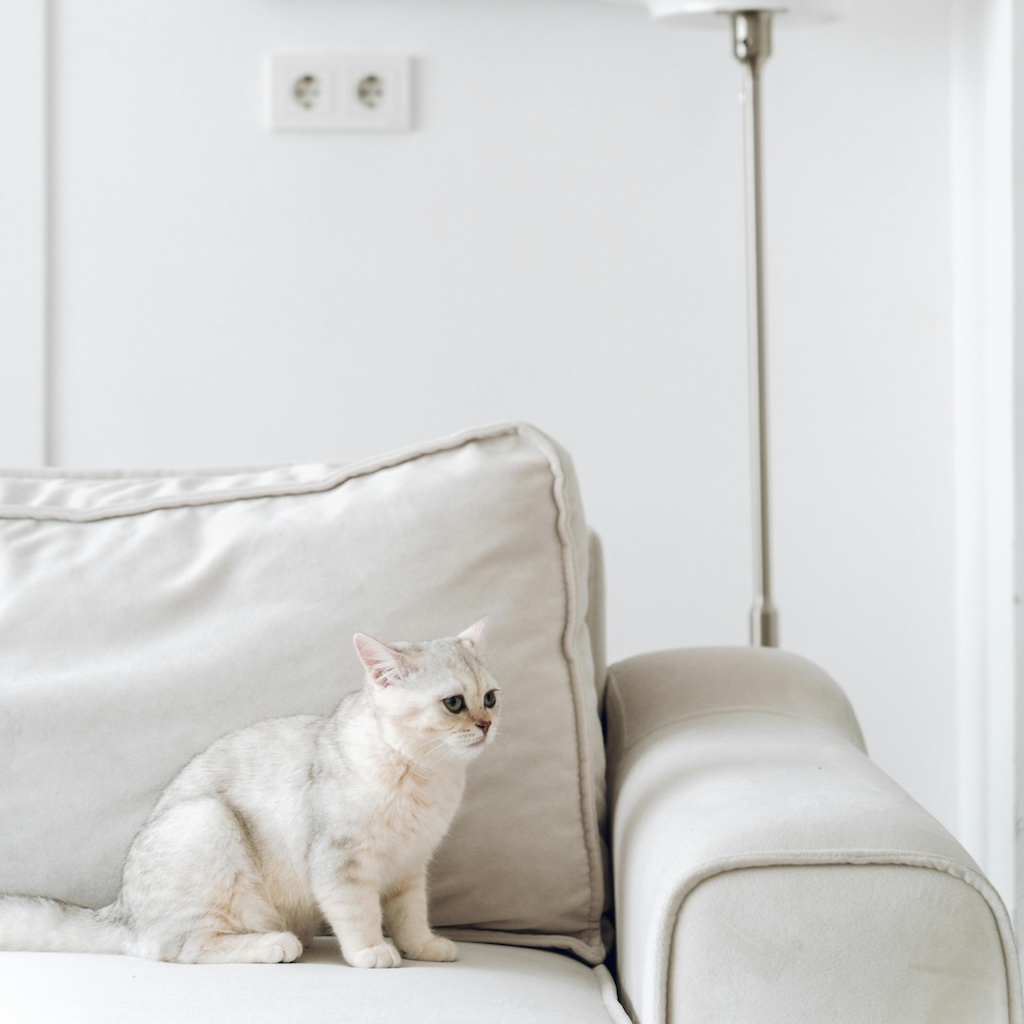}
			&\includegraphics[width=1.5cm]{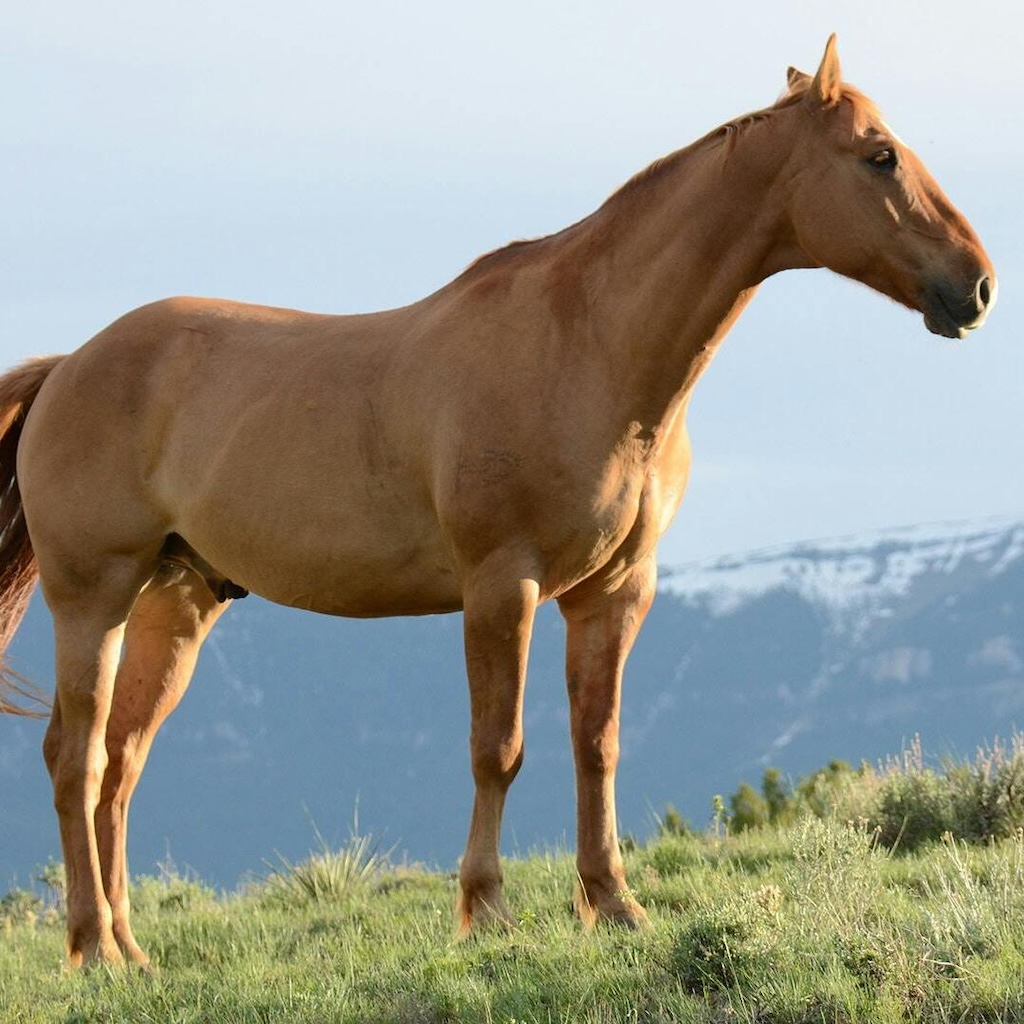}
			&\includegraphics[width=1.5cm]{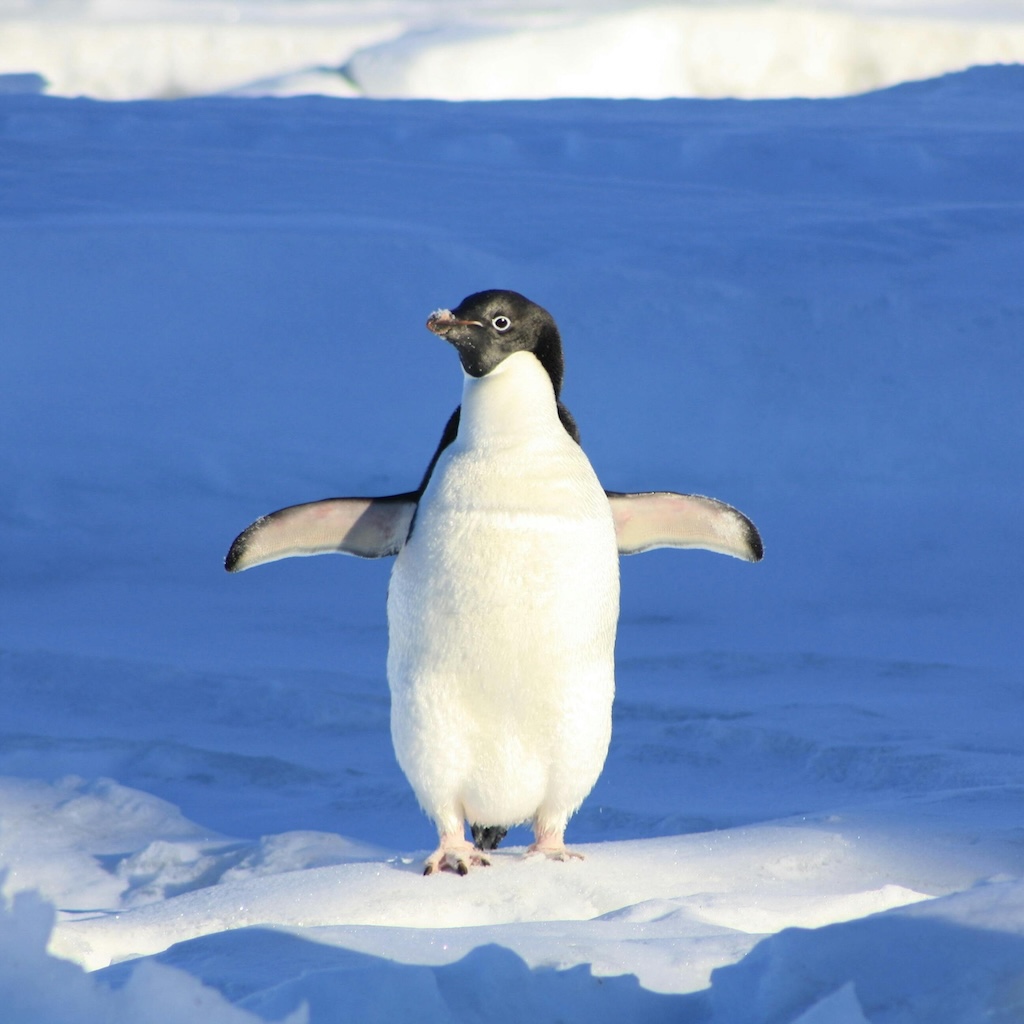}
			&\includegraphics[width=1.5cm]{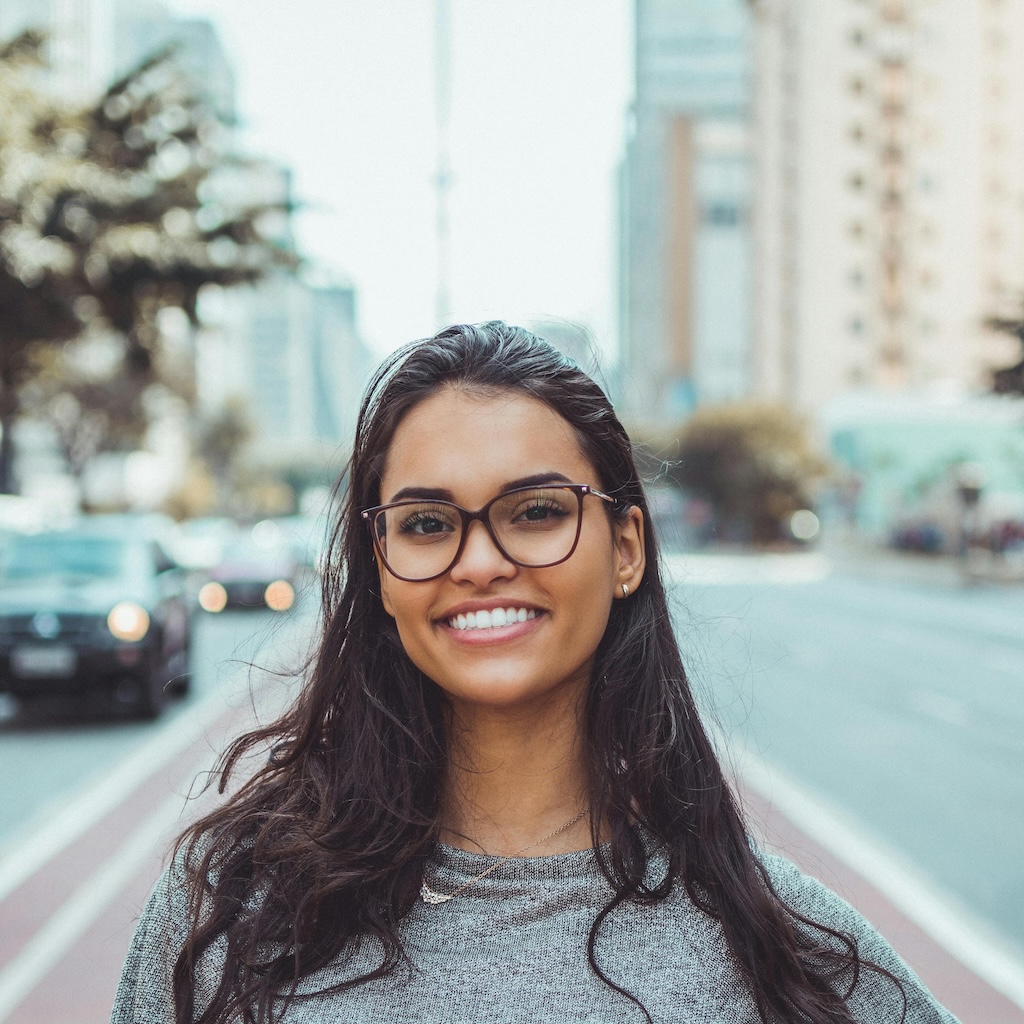}
			\\
                \raisebox{0.15cm}{\rotatebox[origin=c]{90}{\scriptsize{Edited Image}}}
			&\includegraphics[width=1.5cm]{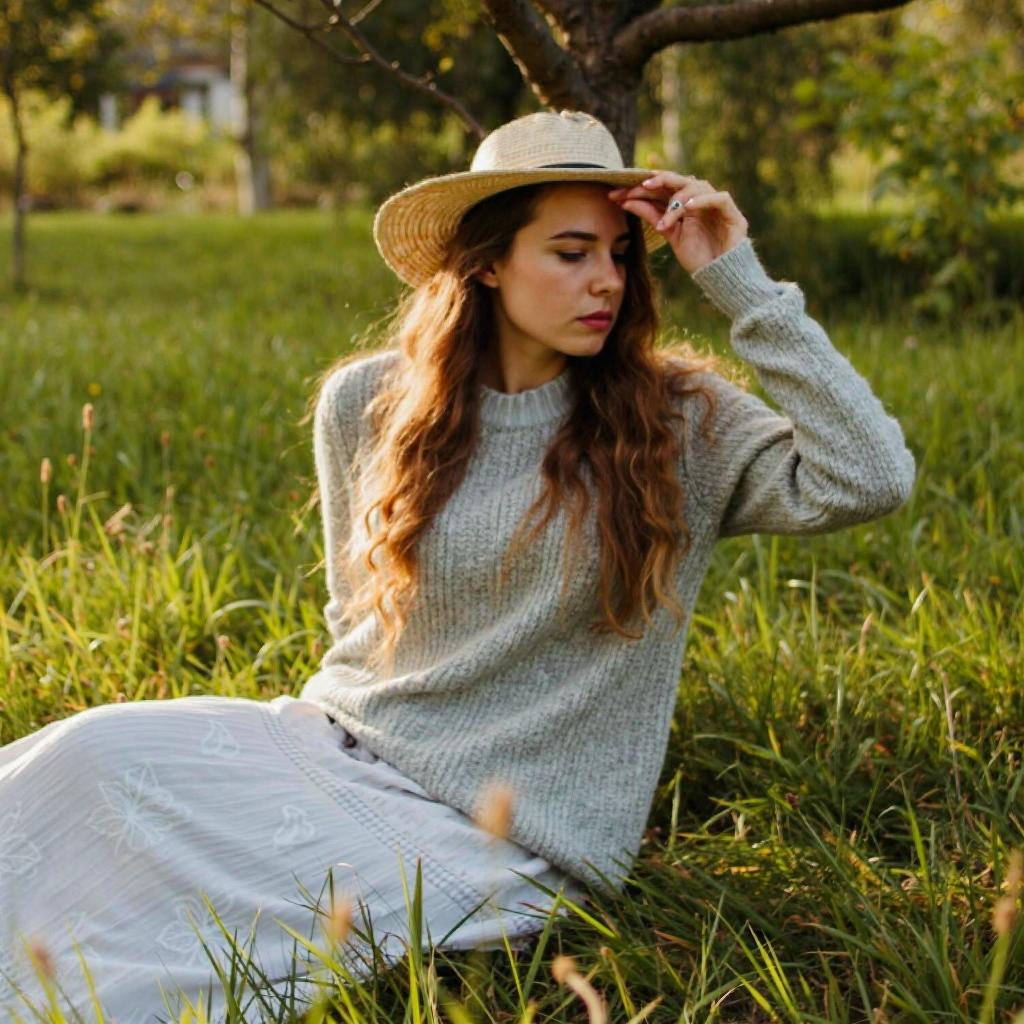}
			&\includegraphics[width=1.5cm]{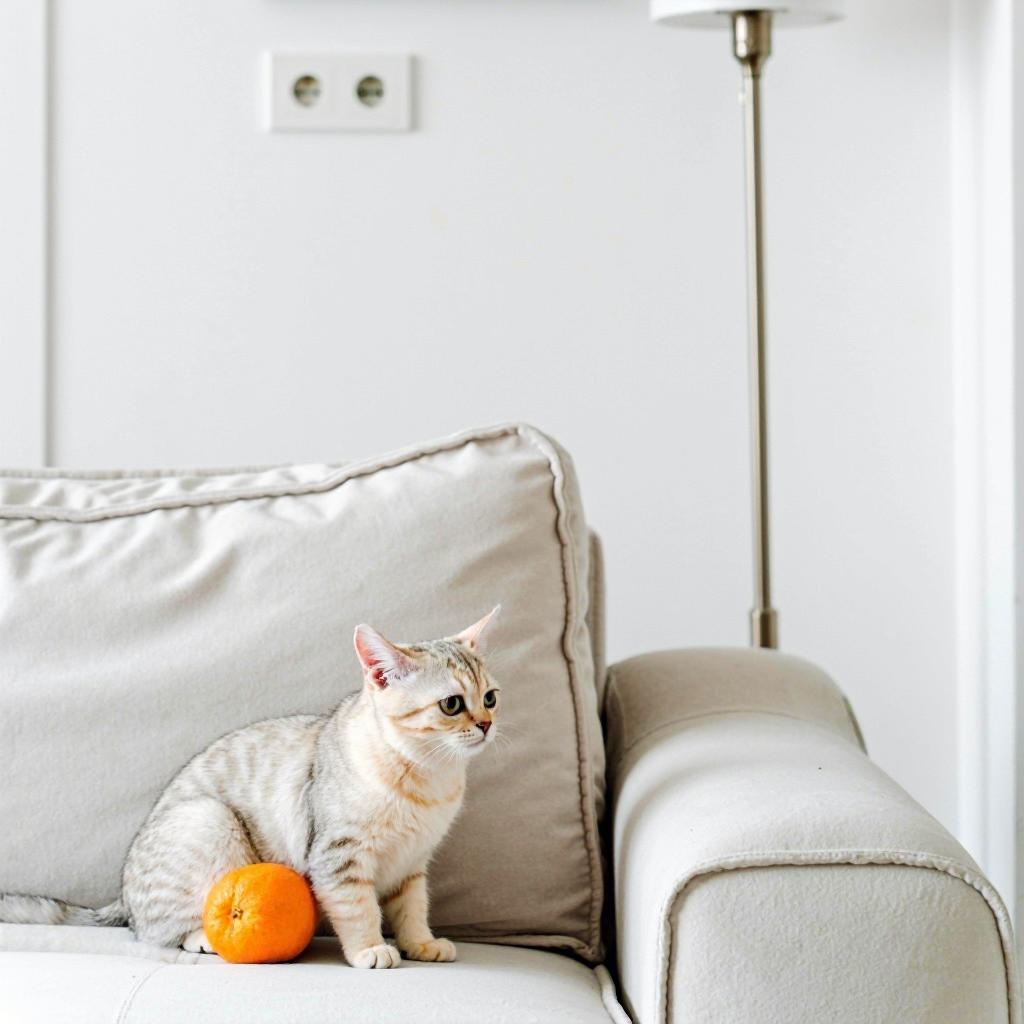}
			&\includegraphics[width=1.5cm]{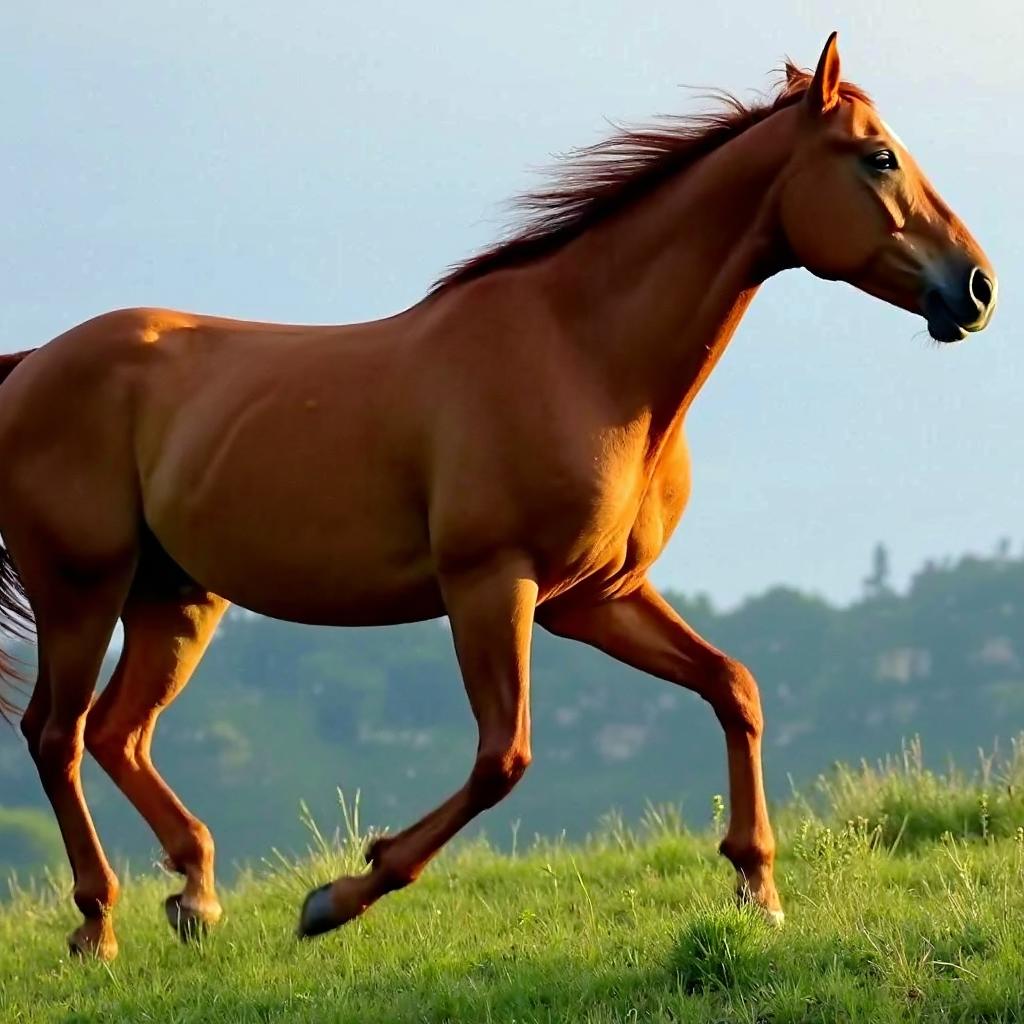}
			&\includegraphics[width=1.5cm]{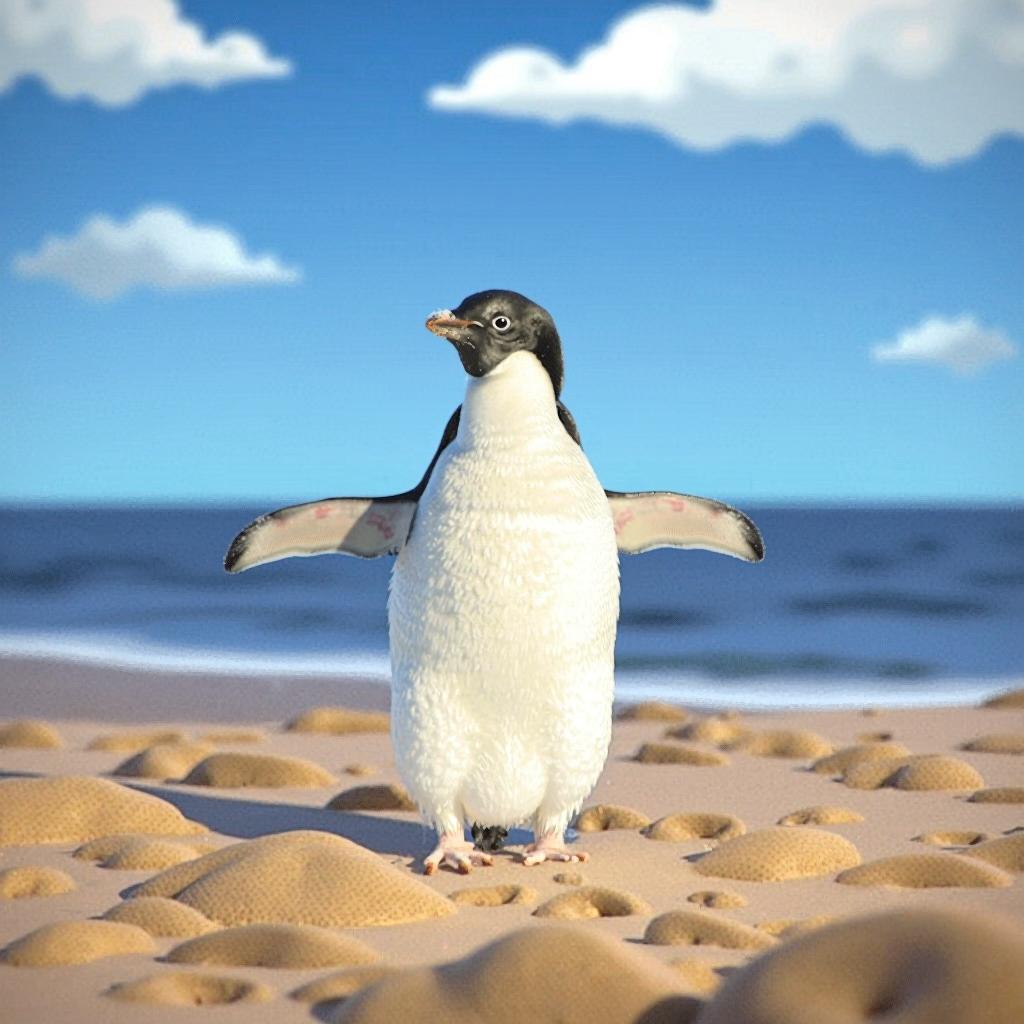}
			&\includegraphics[width=1.5cm]{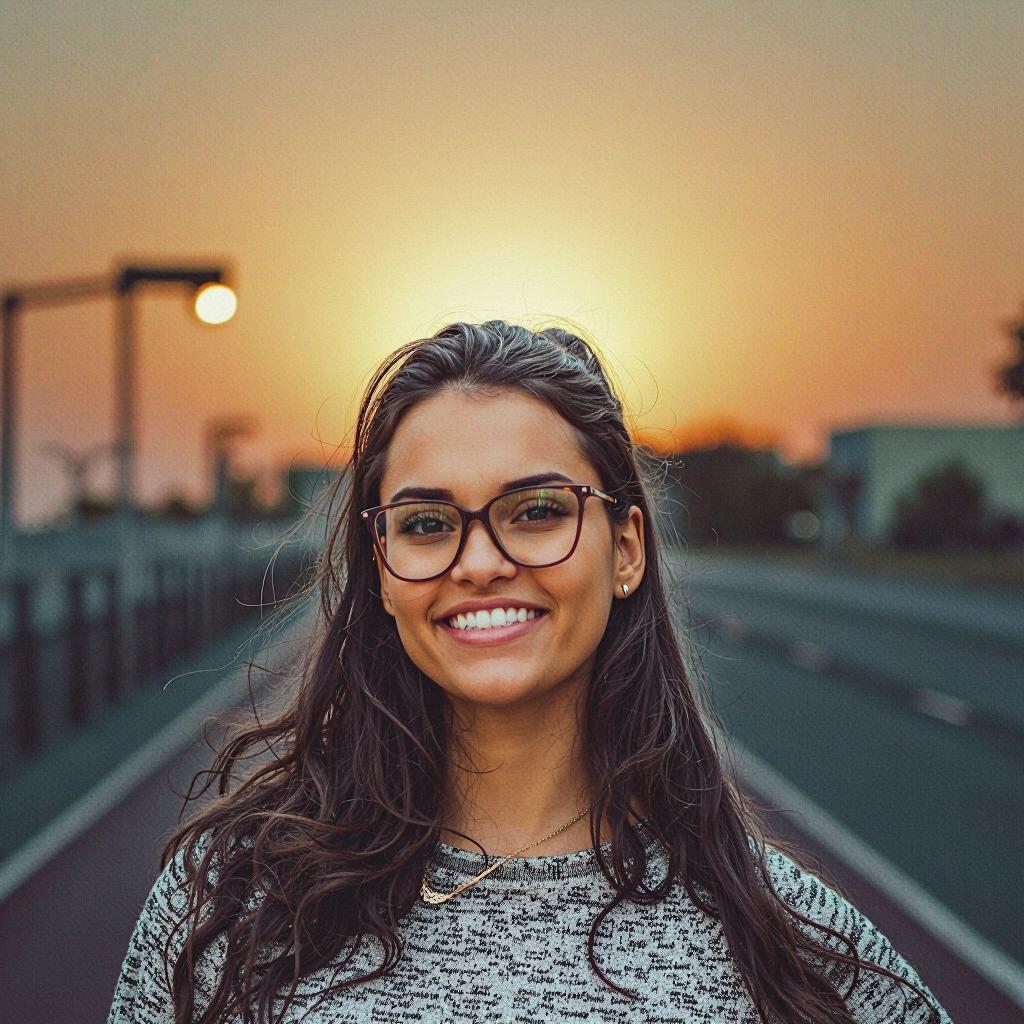}
			\\
			
		\end{tabular}
	\end{center}
        \vspace{-1.5em}
	\caption{Visual results of our method on real images.} 
	\label{fig:real_img}
        \vspace{-1.0em}
\end{figure}

In Figure~\ref{fig:quail_comparsion}, we qualitatively compare our method with these baselines on object addition, non-rigid editing, and background replacement tasks. For object addition, our method not only achieves high-quality editing results but also demonstrates the best ability to preserve irrelevant regions. For non-rigid editing, StableFlow, TamingRF, and OmniGen struggle to produce meaningful deformations, often resulting in near-duplication of the source image. MagicBrush attempts to deform the input image but fails to maintain the original appearance, whereas our method effectively balances object deformation and appearance transfer. For background replacement, our approach delivers the most visually compelling background changes while preserving the foreground object intact. Since these methods cannot handle object movement and outpainting, we present our visual results in Figure~\ref{fig:move_outpainting}. In Figure~\ref{fig:real_img}, we further provide our editing results on real images, which are inverted to the initial noise using the inverse Euler ODE solver.

For quantitative evaluation, we used ChatGPT~\cite{openai2024chatgpt} to generate $200$ diverse pairs of source and editing texts for each task, resulting in a total of $600$ evaluation samples. This approach not only facilitates the computation of quantitative metrics but also removes the impact of inversion from the evaluation, allowing a clearer demonstration of our contributions. The evaluation metrics include: $\text{CLIP}_{img}$, which measures the similarity between the source and edited images; $\text{CLIP}_{txt}$~\cite{radford2021learning}, which evaluates the alignment between the editing text and the edited image; and $\text{CLIP}_{dir}$~\cite{gal2022stylegan}, which computes the similarity between the direction of text changes and image changes. The $\text{CLIP}_{dir}$ metric provides a more precise assessment of editing effectiveness since $\text{CLIP}_{txt}$ considers not only the editing instructions but also descriptions of the source image, such as objects, attributes, and background. For background editing, we replace $\text{CLIP}_{img}$ with the PSNR between the foreground regions before and after editing, as it better reflects foreground preservation. Additionally, we conducted a user study involving $37$ participants with a background in computer vision. Each participant was presented with 60 randomly shuffled result sequences ($20$ per task) and was asked to select the best option by considering both the editing effectiveness and content preservation of the source image. The preference rate (PR) from this user study serves as another key evaluation metric.

Consistent with the qualitative comparison, the quantitative results in Table~\ref{tab:quant_comparsion} further demonstrate the superiority of our approach. For the $\text{CLIP}_{img}$ metric in non-rigid editing, we consider a lower value within a reasonable range to be preferable, as this task involves object deformation. StableFlow, TamingRF, and OmniGen struggle with this type of editing (as confirmed by $\text{CLIP}_{dir}$), leading to near-duplicate outputs of the source image and thus high $\text{CLIP}_{img}$ scores. While MagicBrush achieves the lowest $\text{CLIP}_{img}$ value, it loses significant details of the source image’s appearance. Our method effectively balances object deformation and appearance preservation. More aligned with human perception, the user study further highlights the strong advantages of our approach.

\subsection{Ablation Analysis}

\noindent\textbf{Correctness of Layer Selection for Attention Sharing Driven by Task Characteristics.} Combining our discovery of the layer-specific roles in RoPE-based MMDiT, we customize the layer selection mechanism for attention sharing based on the characteristics of different editing tasks. To verify its effectiveness, we design the following two settings: (1) Swapping the layers for object addition and non-rigid editing, i.e., performing key-value attention sharing for object addition in $\mathds{C}$ layers that rely more on content similarity, while executing it for non-rigid editing in $\mathds{P}$ layers that are more position-dependent. (2) Applying attention sharing across all layers for both object addition and non-rigid editing. The quantitative results are provided in Table~\ref{tab:ablation_add_non_rigid}. For object addition, performing attention sharing only in content-similarity-dependent layers $\mathds{C}$ significantly reduces the preservation of unrelated regions. When applied across all layers, our Reasoning-before-Generation strategy prevents suppression of the added object; however, since content-similarity-dependent layers also participate in attention sharing, the similarity of unrelated regions decreases. For non-rigid editing, applying attention sharing in position-dependent layers $\mathds{P}$ or across all layers causes the edited image to query information from the source image based on spatial position, leading to results that resemble direct copying. The visual results in Figures~\ref{fig:ablation_add_object} and~\ref{fig:ablation_non_rigid} further support this analysis. For instance, in Figure~\ref{fig:ablation_add_object}, applying attention sharing across all layers results in the loss of the girl's headband, as well as deformations in the deer's head and antlers.

\begin{table}[h]
    \centering
    \begin{subtable}[h]{\columnwidth}
        \centering
        \small
        \setlength{\tabcolsep}{8.6pt}{
        \begin{tabular}{m{1.9cm}<{\centering}|lll}
            \hline      
            Settings & \small{$\text{CLIP}_{img}$ $\uparrow$} & \small{$\text{CLIP}_{txt}$ $\uparrow$} & \small{$\text{CLIP}_{dir}$ $\uparrow$} \\ \hline
            \rowcolor[HTML]{FFF2CC}
            Ours  & \small{0.974} & \small{0.321} & \small{0.202} \\
            $\mathds{C}$ Layers  & \small{0.926} & \small{0.327} & \small{0.198} \\
            All Layers  & \small{0.969} & \small{0.319} & \small{0.202} \\
            w/o Reasoning  & \small{0.990} & \small{0.302} & \small{0.089} \\ \hline
        \end{tabular}
            }
        \caption{Object Addition}
        \label{tab:object_add_ablation}
    \end{subtable}
    \hspace{1em}

    \begin{subtable}[h]{\columnwidth}
        \centering
        \small
        \setlength{\tabcolsep}{8.6pt}{
        \begin{tabular}{m{1.9cm}<{\centering}|lll}
            \hline
            Settings & \small{$\text{CLIP}_{img}$ $\downarrow$} & \small{$\text{CLIP}_{txt}$ $\uparrow$} & \small{$\text{CLIP}_{dir}$ $\uparrow$} \\ \hline
            \rowcolor[HTML]{FFF2CC}
            Ours  & \small{0.940} & \small{0.315} & \small{0.153} \\
            $\mathds{P}$ Layers  & \small{0.992} & \small{0.300} & \small{0.057} \\
            All Layers  & \small{0.996} & \small{0.296} & \small{0.008} \\ \hline
        \end{tabular}
            }
        \caption{Non-Rigid Editing}
        \label{tab:non_rigid_ablation}
    \end{subtable}
    \vspace{-0.5em}
    \caption{Quantitative ablation on layer selection and reasoning-before-generation.}
    \label{tab:ablation_add_non_rigid}
    \vspace{-0.5em}
\end{table}

\begin{figure}[t]
	\begin{center}
		\setlength{\tabcolsep}{0.5pt}
		\begin{tabular}{m{0.3cm}<{\centering}m{1.55cm}<{\centering}|m{1.55cm}<{\centering}m{1.55cm}<{\centering}m{1.55cm}<{\centering}m{1.55cm}<{\centering}}
			& \scriptsize{Source Image} & \scriptsize{$\mathds{C}$ Layers} & \scriptsize{All Layers} & \scriptsize{w/o Reasoning} & \scriptsize{Ours}
			\\

			\raisebox{0.05cm}{\rotatebox[origin=c]{90}{\footnotesize{{\textcolor{myblue}{\textit{`balloon'}}}}}}
			&\includegraphics[width=1.5cm]{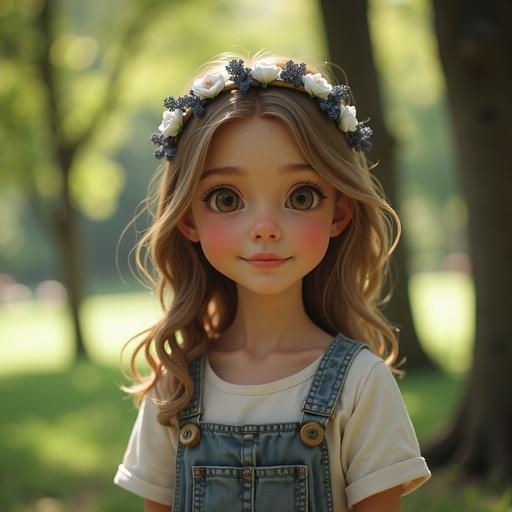}
			&\includegraphics[width=1.5cm]{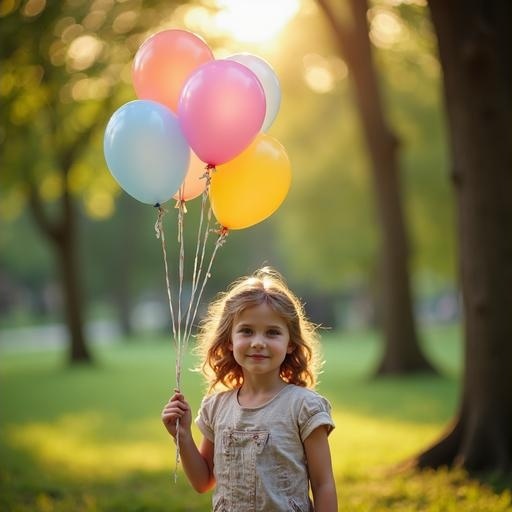}
			&\includegraphics[width=1.5cm]{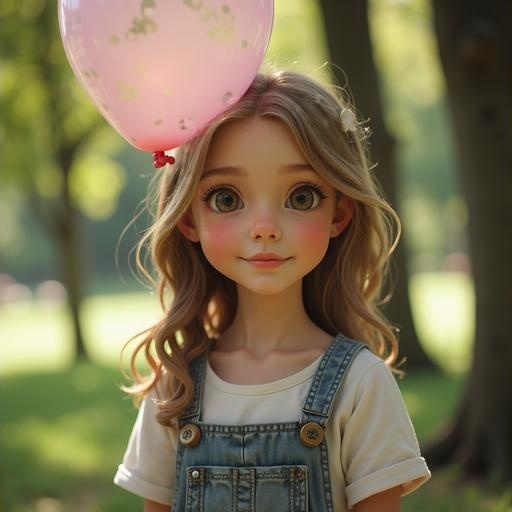}
			&\includegraphics[width=1.5cm]{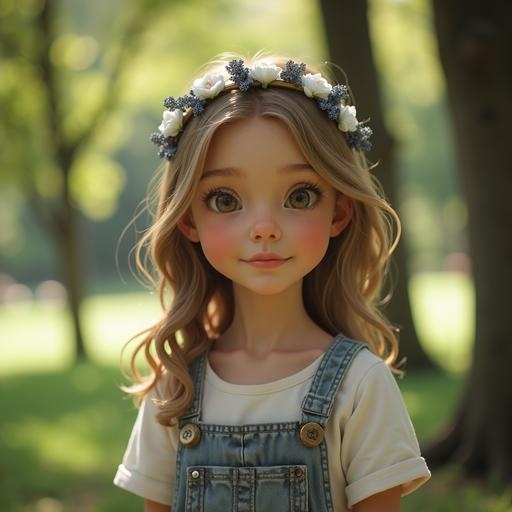}
			&\includegraphics[width=1.5cm]{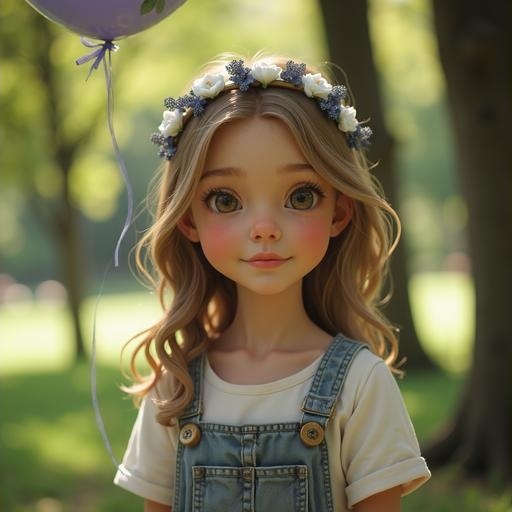}
			\\
                \raisebox{0.05cm}{\rotatebox[origin=c]{90}{\footnotesize{{\textcolor{myblue}{\textit{`flower crown'}}}}}}
			&\includegraphics[width=1.5cm]{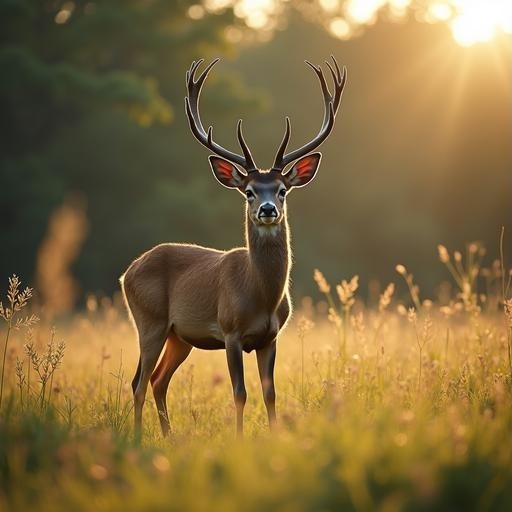}
			&\includegraphics[width=1.5cm]{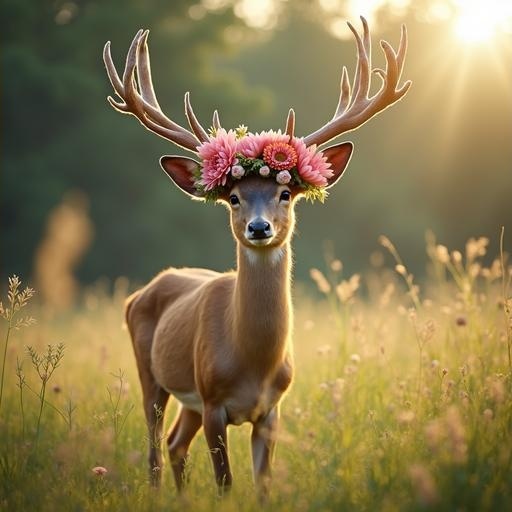}
			&\includegraphics[width=1.5cm]{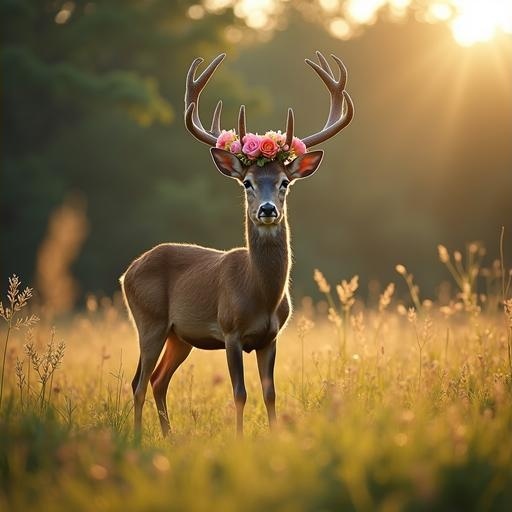}
			&\includegraphics[width=1.5cm]{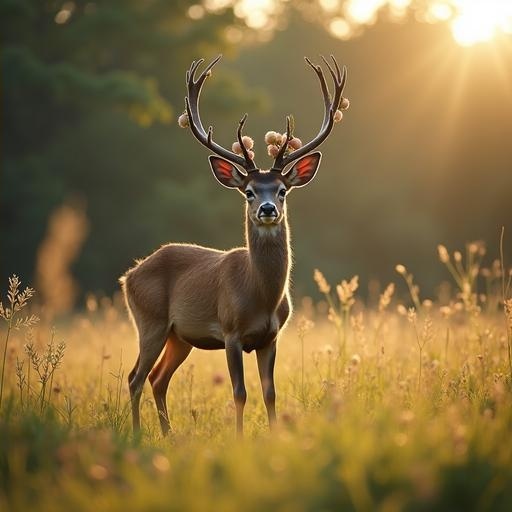}
			&\includegraphics[width=1.5cm]{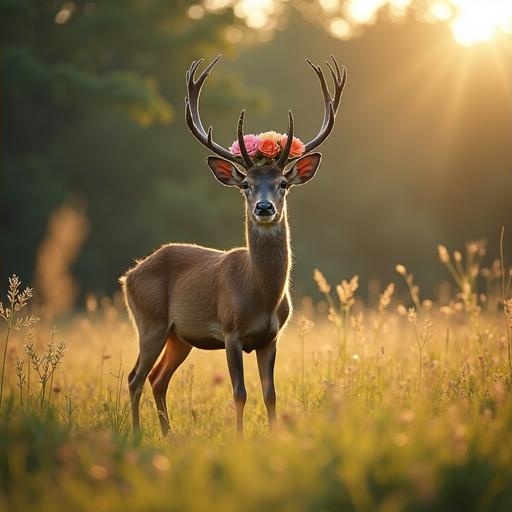}
			\\
			
		\end{tabular}
	\end{center}
        \vspace{-1.5em}
	\caption{Qualitative ablation on layer selection and reasoning-before-generation for object addition.} 
	\label{fig:ablation_add_object}
        \vspace{-1.5em}
\end{figure}

\noindent\textbf{Importance of Reasoning-before-Generation.} Table~\ref{tab:object_add_ablation} and Figure~\ref{fig:ablation_add_object} demonstrate that without the \textit{Reasoning-before-Generation} strategy, the added objects would be severely suppressed, as evidenced by the missing balloon and the incomplete flower crown in Figure~\ref{fig:ablation_add_object}.

\begin{figure}[t]
	\begin{center}
		\setlength{\tabcolsep}{0.5pt}
		\begin{tabular}{m{0.3cm}<{\centering}m{1.93cm}<{\centering}|m{1.93cm}<{\centering}m{1.93cm}<{\centering}m{1.93cm}<{\centering}}
			& \scriptsize{Source Image} & \scriptsize{$\mathds{P}$ Layers} & \scriptsize{All Layers} & \scriptsize{Ours}
			\\

			\raisebox{0.05cm}{\rotatebox[origin=c]{90}{\footnotesize{{\textcolor{myblue}{\textit{`flapping'}}}}}}
			&\includegraphics[width=1.85cm]{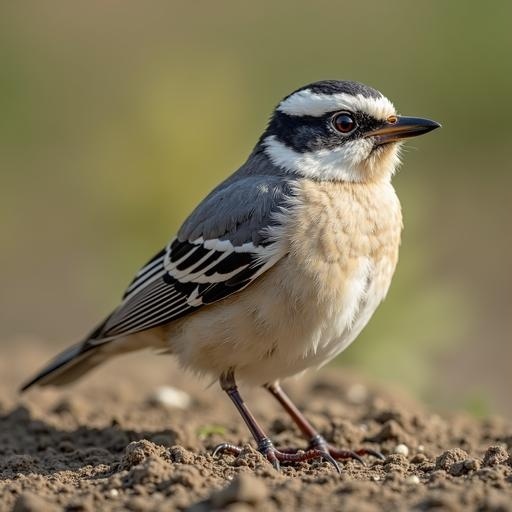}
			&\includegraphics[width=1.85cm]{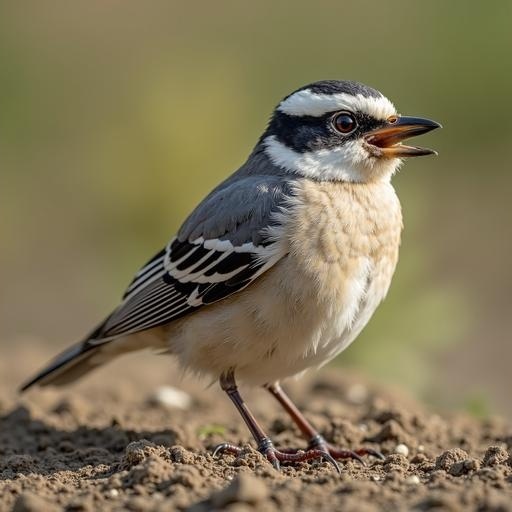}
			&\includegraphics[width=1.85cm]{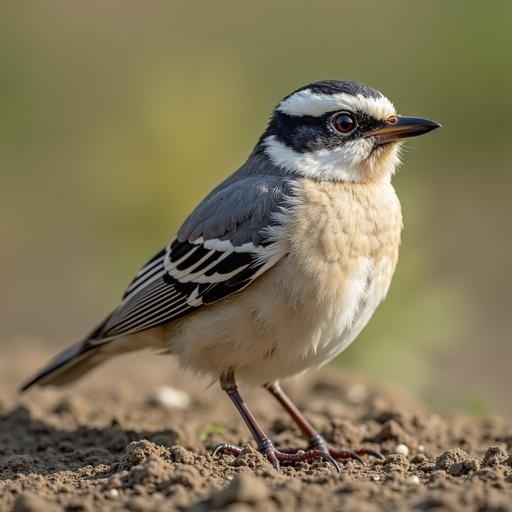}
			&\includegraphics[width=1.85cm]{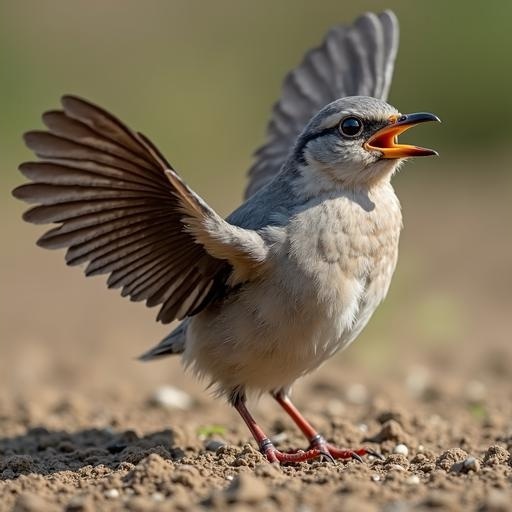}
			\\
                \raisebox{0.05cm}{\rotatebox[origin=c]{90}{\footnotesize{{\textcolor{myblue}{\textit{`jumping'}}}}}}
			&\includegraphics[width=1.85cm]{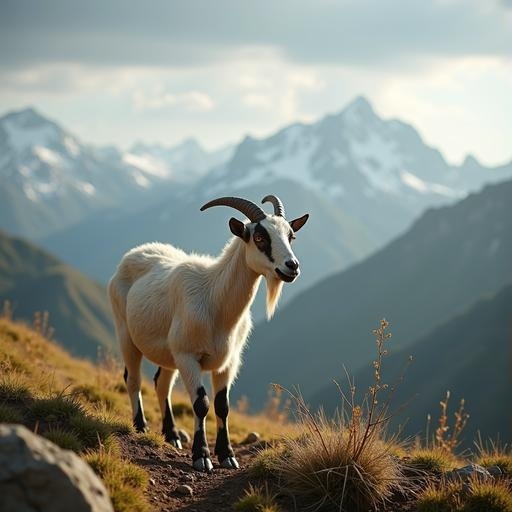}
			&\includegraphics[width=1.85cm]{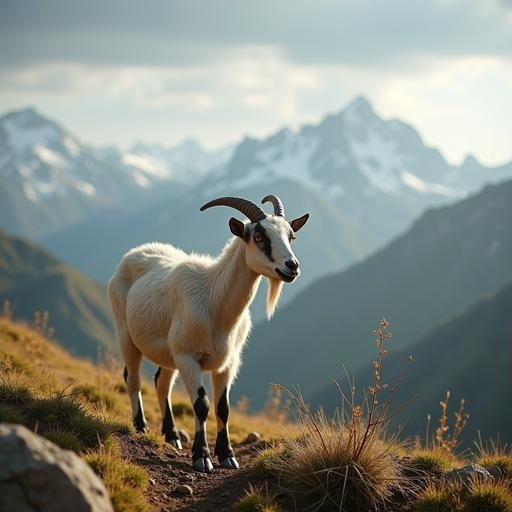}
			&\includegraphics[width=1.85cm]{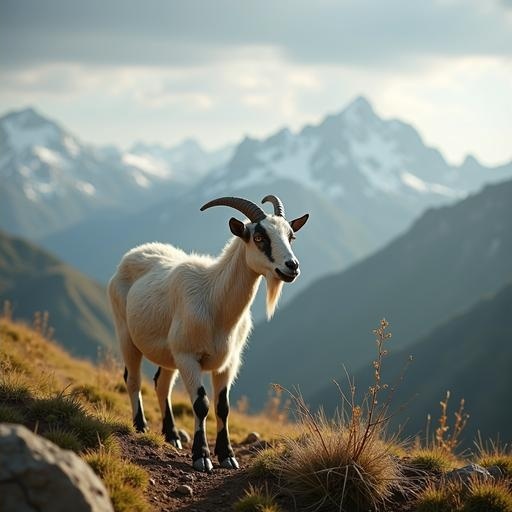}
			&\includegraphics[width=1.85cm]{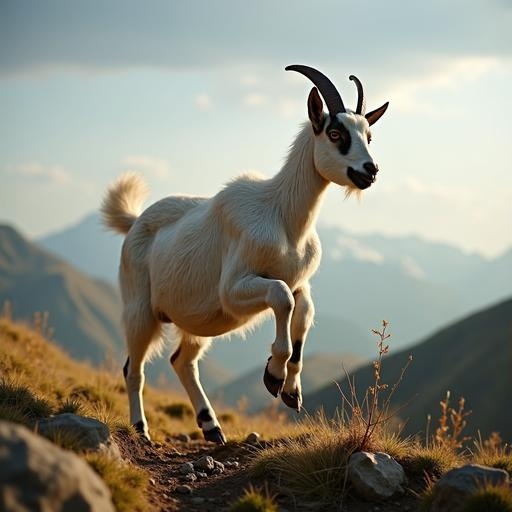}
			\\
			
		\end{tabular}
	\end{center}
        \vspace{-1.5em}
	\caption{Ablation on layer selection for non-rigid editing.} 
	\label{fig:ablation_non_rigid}
\end{figure}

\begin{figure}[t]
	\begin{center}
		\setlength{\tabcolsep}{0.5pt}
		\begin{tabular}{m{0.3cm}<{\centering}m{1.55cm}<{\centering}|m{1.55cm}<{\centering}m{1.55cm}<{\centering}m{1.55cm}<{\centering}m{1.55cm}<{\centering}}
			& \scriptsize{Source Image} & \scriptsize{Latent-$25$} & \scriptsize{Latent-$35$} & \scriptsize{Latent-$45$} & \scriptsize{Value-$45$}
			\\

			\raisebox{0.05cm}{\rotatebox[origin=c]{90}{\footnotesize{{\textcolor{myblue}{\textit{`desert'}}}}}}
			&\includegraphics[width=1.5cm]{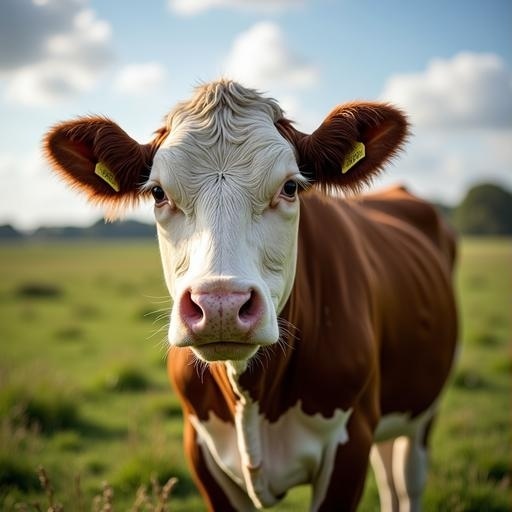}
			&\includegraphics[width=1.5cm]{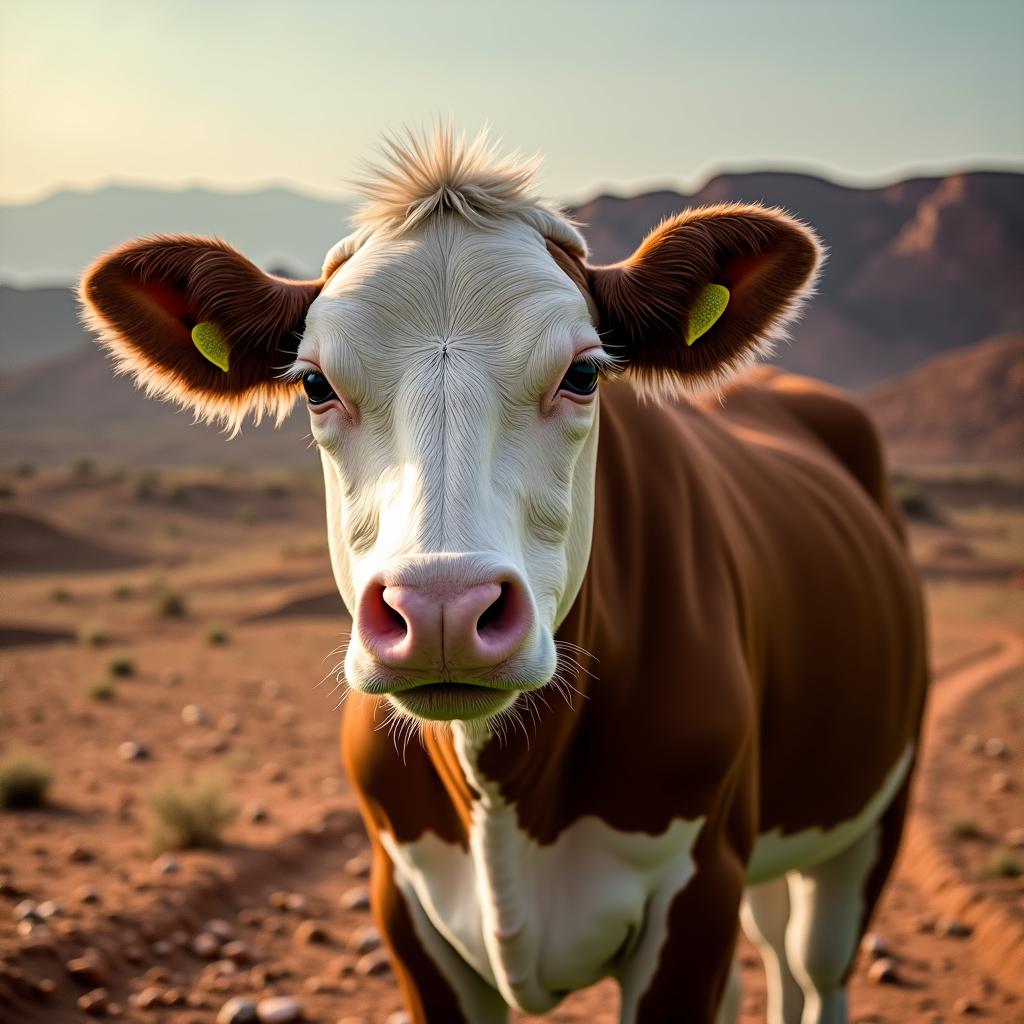}
			&\includegraphics[width=1.5cm]{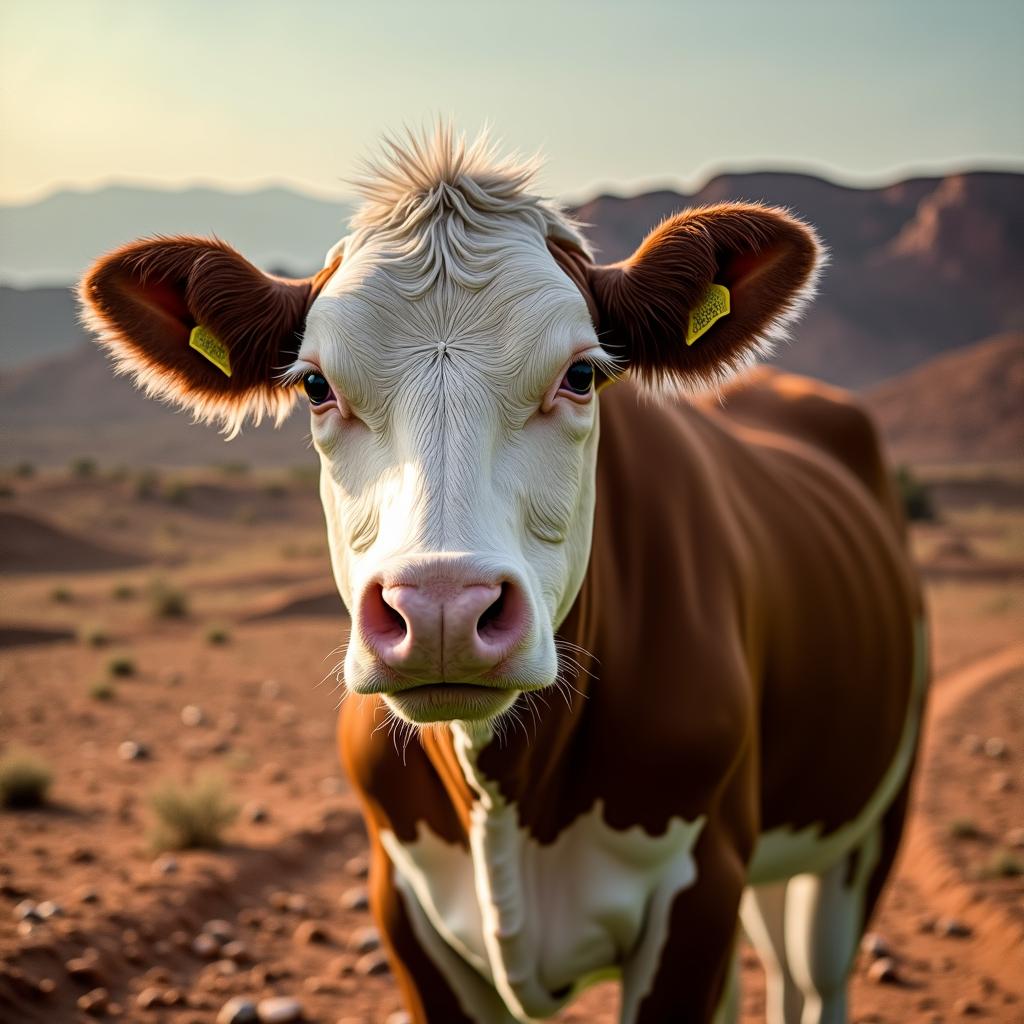}
			&\includegraphics[width=1.5cm]{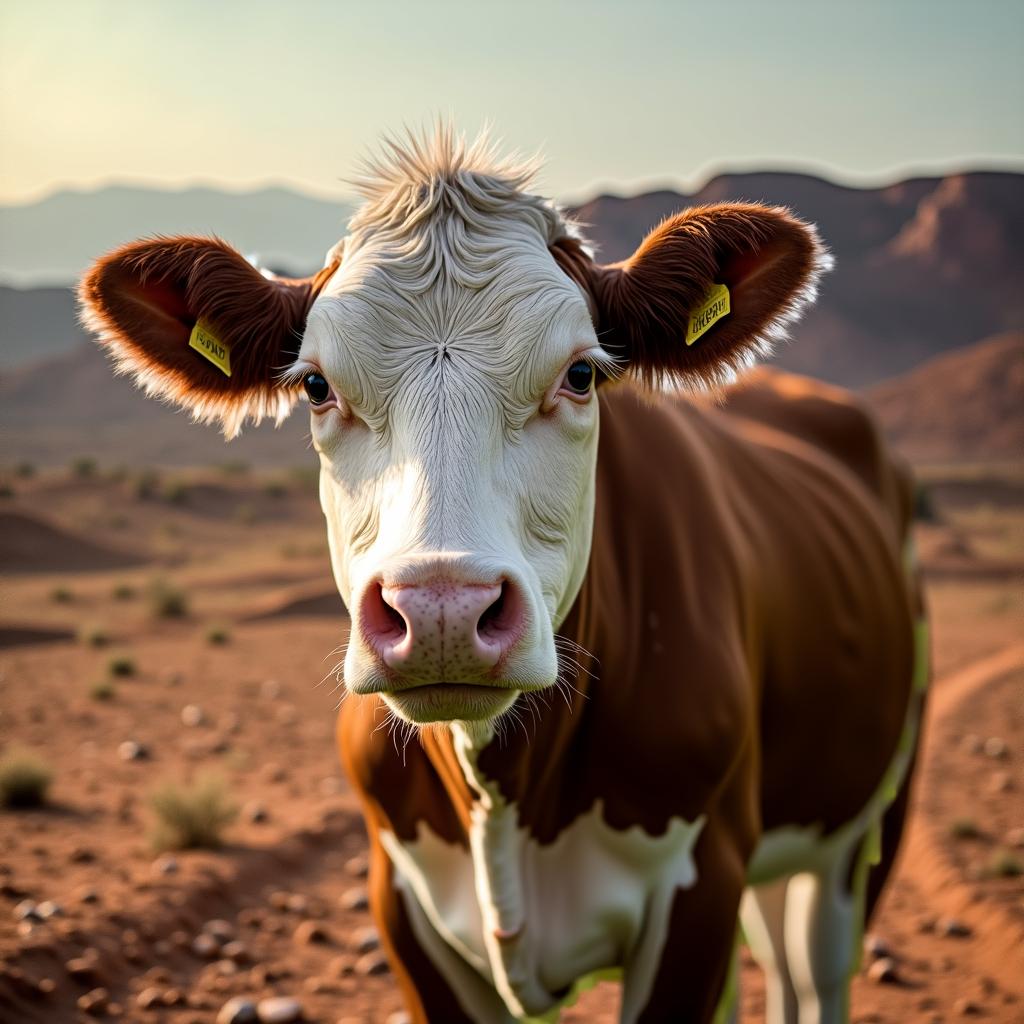}
			&\includegraphics[width=1.5cm]{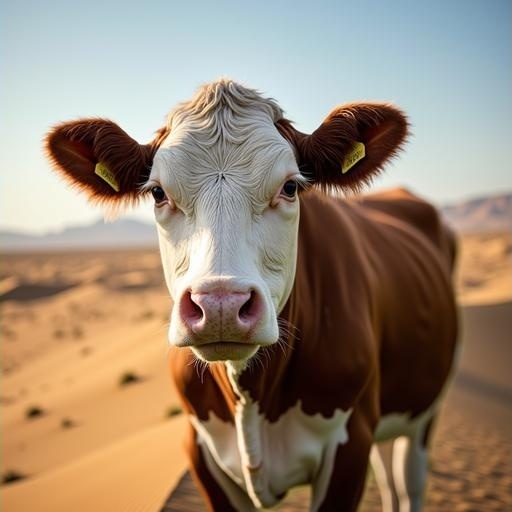}
			\\
                \raisebox{0.05cm}{\rotatebox[origin=c]{90}{\footnotesize{{\textcolor{myblue}{\textit{`snowy day'}}}}}}
			&\includegraphics[width=1.5cm]{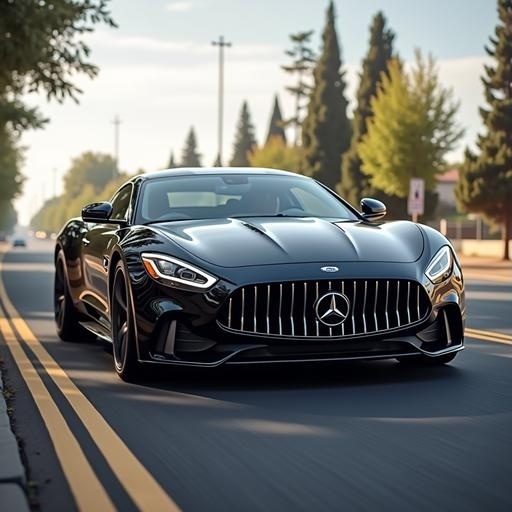}
			&\includegraphics[width=1.5cm]{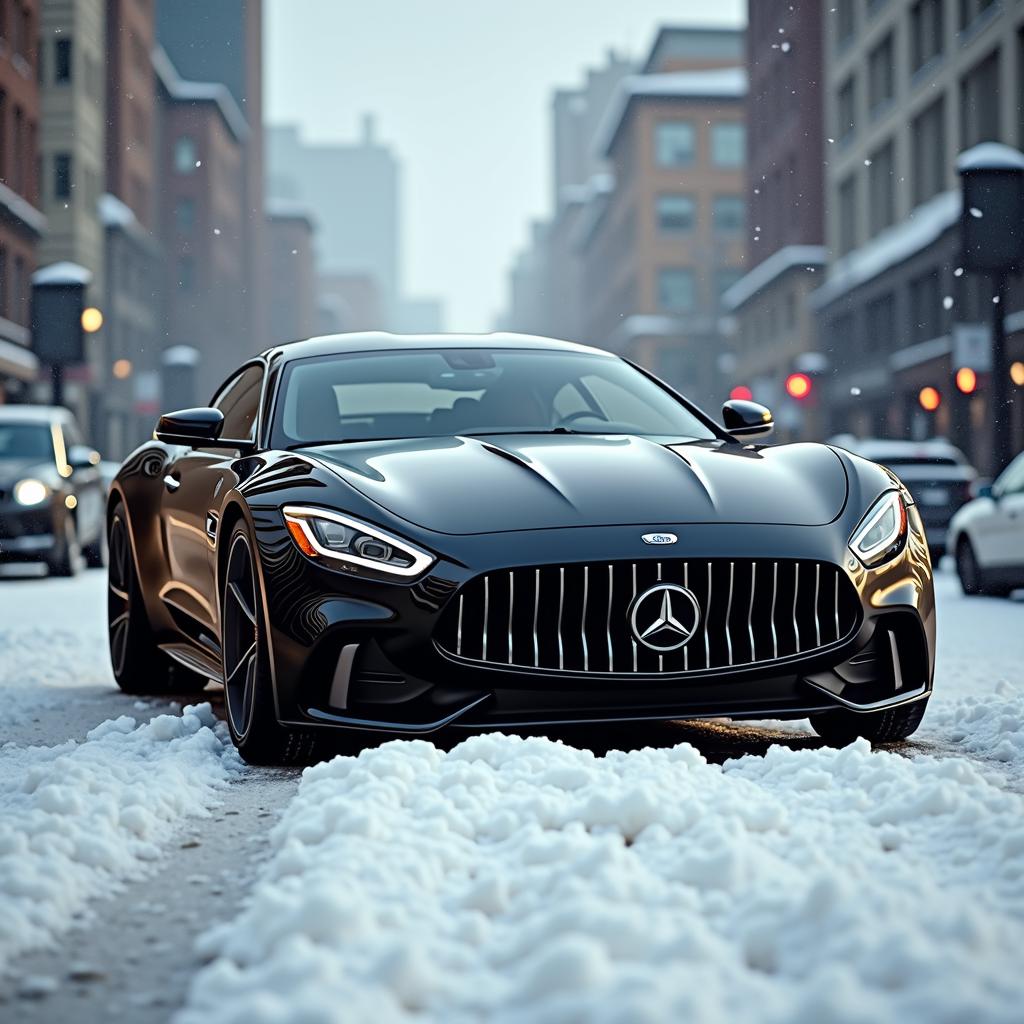}
			&\includegraphics[width=1.5cm]{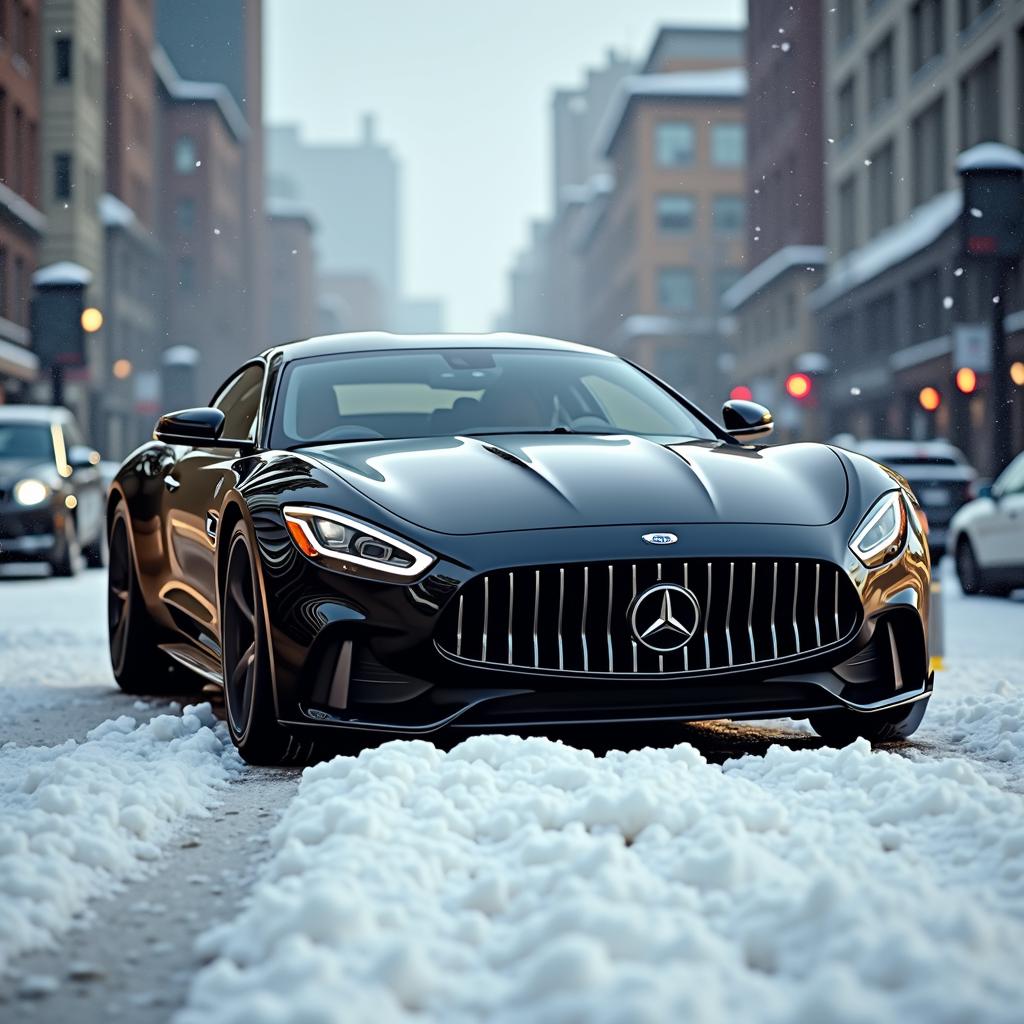}
			&\includegraphics[width=1.5cm]{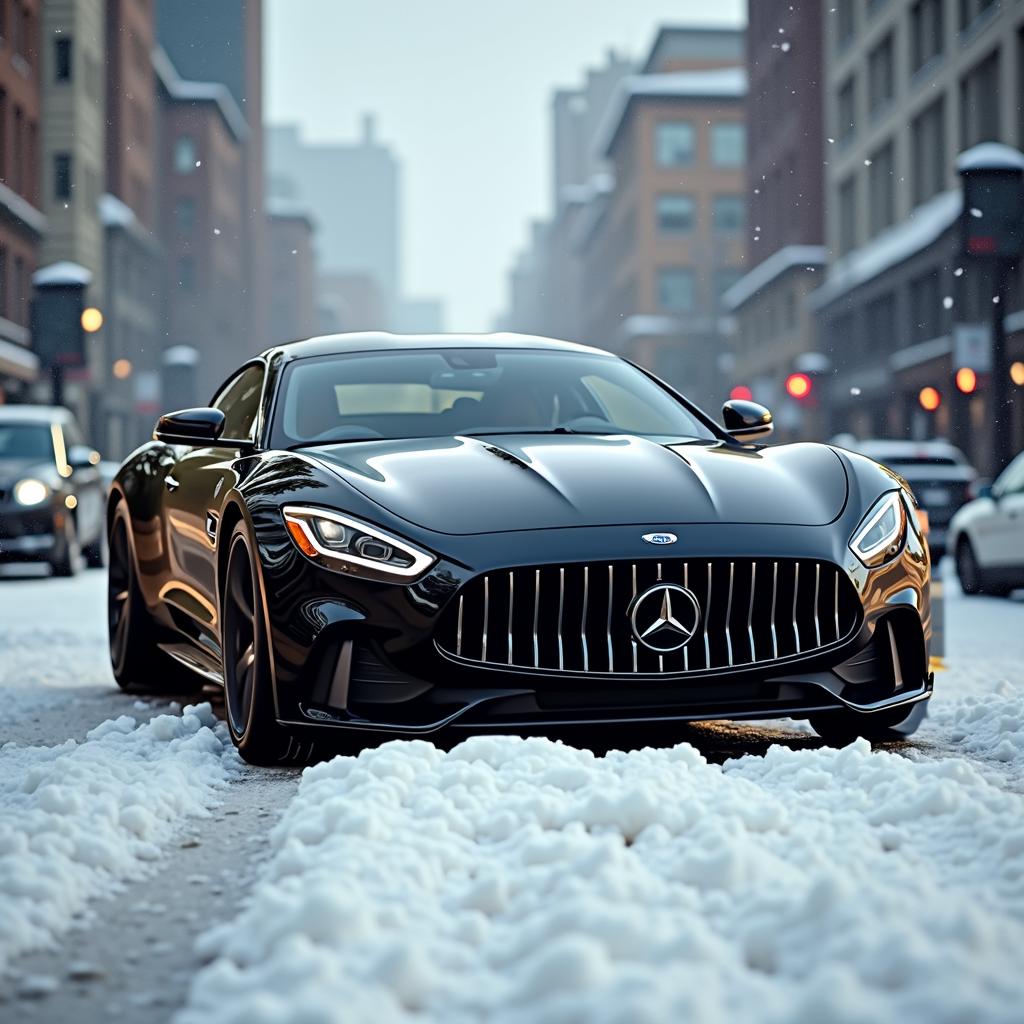}
			&\includegraphics[width=1.5cm]{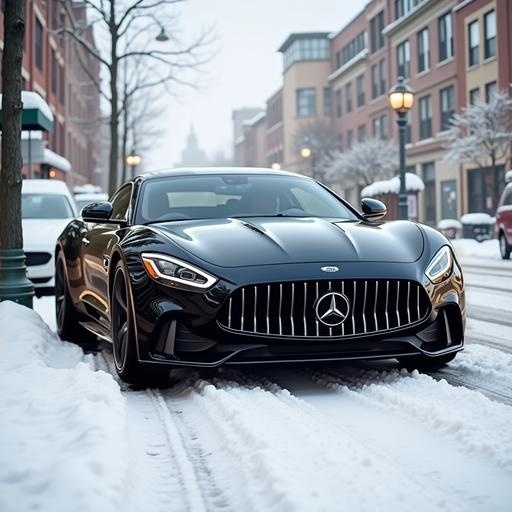}
			\\
			
		\end{tabular}
	\end{center}
        \vspace{-1.5em}
	\caption{Ablation of value vs. latent blending. Numbers represent total denoising steps during which blending is applied.}
        \vspace{-1.0em}
	\label{fig:ablation_bg}
\end{figure}

\noindent\textbf{Value Blending vs. Latent Blending.} For region-preserved editing (e.g., background replacement), we apply value attention sharing across all layers for the regions that need to be preserved (e.g., the foreground). This ensures spatially consistent copying effects in both position-dependent and content-similarity-dependent layers. In Figure~\ref{fig:ablation_bg}, we compare this value blending approach with the classical latent blending method~\cite{avrahami2023blended}. Clearly, stepwise value replacement within joint self-attention preserves the foreground while achieving better integration. In contrast, latent blending fails to maintain an unchanged foreground when applied with too few steps and introduces noticeable artifacts at the edges when applied with too many steps, such as the green tint around the cow and the sports car copying part of the road from the source image.
\section{Conclusions}
\label{sec:conclusion}

In this paper, we provide the first mechanistic analysis of layer-specific roles in RoPE-based MMDiT models, unveiling distinct layer-specific dependencies on positional encoding and content similarity during text-to-image generation. By systematically probing FLUX through strategic manipulations of RoPE, we identify patterns that challenge conventional assumptions about attention layer roles. Building upon these insights, we introduce a training-free, task-specific image editing framework that categorizes editing tasks based on their reliance on positional information or content similarity. We design tailored key-value injection strategies for each category, leading to a highly effective and flexible editing approach. Qualitative and quantitative evaluations consistently demonstrate the superiority of our method over existing state-of-the-art techniques. Future work may focus on reducing the gap in performance on real images, such as designing better inversion algorithms or leveraging the techniques proposed in this paper to generate a large-scale edited dataset for fine-tuning an instruction-based editing model.

{
    \small
    \bibliographystyle{ieeenat_fullname}
    \bibliography{main}
}

\clearpage
\maketitlesupplementary

\section{Algorithm}
The pseudo-code for performing object addition, non-rigid editing, background replacement, object movement, and outpainting using our method is provided in Algorithms~\ref{alg:obj_addition},~\ref{alg:non_rigid},~\ref{alg:bg_replace},~\ref{alg:object_move}, and~\ref{alg:outpainting}.

\begin{algorithm}[h]
	\SetAlgoLined
	\KwIn{A source prompt $\mathcal{P}_{src}$, A editing prompt $\mathcal{P}_{edit}$, a pretrained RoPE-based MMDiT text-to-image model $\varepsilon_\theta$, an image decoder $\mathcal{D}$, total sampling steps $T$ (default $50$), denoising step $R$ (default $7$) when applying Reasoning-before-Generation, the most position-dependent layers $\mathds{P}$.}
	\KwOut{An edited image $x^{edit}$ aligned with the editing prompt $\mathcal{P}_{edit}$.}
        \textbf{Initialize} $z_T^{src} \sim \mathcal{N}(0, 1)$, $z_T^{edit} \leftarrow z_T^{src}$, $z_{init} \leftarrow z_T^{src}$.
        
	\For{$t=T$ to $T-R+1$}{
            \For{$i$ in $\mathds{P}$}{
                $Q_{src}^{t-i}, K_{src}^{t-i}, V_{src}^{t-i} \leftarrow \varepsilon_\theta(z_t^{src}, t, \mathcal{P}_{src})$
                
                $Q_{edit}^{t-i}, K_{edit}^{t-i}, V_{edit}^{t-i} \leftarrow \varepsilon_\theta(z_t^{edit}, t, \mathcal{P}_{edit})$
                
                $Attn(Q_{edit}^{t-i},K_{src}^{t-i},V_{src}^{t-i})$
            }
            $z_{t-1}^{src} \leftarrow \varepsilon_\theta(z_t^{src}, t, \mathcal{P}_{src})$

            $z_{t-1}^{edit} \leftarrow \varepsilon_\theta(z_t^{edit}, t, \mathcal{P}_{edit}, Attn)$
        }
        The added object region mask $M_{obj}$ is reasoned out.

        $z_T^{src} \leftarrow z_{init}$, $z_T^{edit} \leftarrow z_{init}$

	\For{$t=T$ to $1$}{
            \For{$i$ in $\mathds{P}$}{
                $Q_{src}^{t-i}, K_{src}^{t-i}, V_{src}^{t-i} \leftarrow \varepsilon_\theta(z_t^{src}, t, \mathcal{P}_{src})$
                
                $Q_{edit}^{t-i}, K_{edit}^{t-i}, V_{edit}^{t-i} \leftarrow \varepsilon_\theta(z_t^{edit}, t, \mathcal{P}_{edit})$

                $K_{obj}^{t-i}=M_{obj}\times K_{edit}^{t-i}+(1-M_{obj})\times K_{src}^{t-i}$

                $V_{obj}^{t-i}=M_{obj}\times V_{edit}^{t-i}+(1-M_{obj})\times V_{src}^{t-i}$
                
                $Attn(Q_{edit}^{t-i},K_{obj}^{t-i},V_{obj}^{t-i})$
            }
            $z_{t-1}^{src} \leftarrow \varepsilon_\theta(z_t^{src}, t, \mathcal{P}_{src})$

            $z_{t-1}^{edit} \leftarrow \varepsilon_\theta(z_t^{edit}, t, \mathcal{P}_{edit}, Attn)$
        }
        
        $x^{src} \leftarrow \mathcal{D}(z_0^{src}) $
        
        $x^{edit} \leftarrow \mathcal{D}(z_0^{edit}) $
        
        \textbf{Return:} $x^{edit}$

	\caption{Object Addition}
	\label{alg:obj_addition}
\end{algorithm}

\begin{algorithm}[h]
	\SetAlgoLined
	\KwIn{A source prompt $\mathcal{P}_{src}$, A editing prompt $\mathcal{P}_{edit}$, a pretrained RoPE-based MMDiT text-to-image model $\varepsilon_\theta$, an image decoder $\mathcal{D}$, total sampling steps $T$ (default $50$), the more content-similarity-dependent layers $\mathds{C}$.}
	\KwOut{An edited image $x^{edit}$ aligned with the editing prompt $\mathcal{P}_{edit}$.}
        \textbf{Initialize} $z_T^{src} \sim \mathcal{N}(0, 1)$, $z_T^{edit} \leftarrow z_T^{src}$.
        
	\For{$t=T$ to $1$}{
            \For{$i$ in $\mathds{C}$}{
                $Q_{src}^{t-i}, K_{src}^{t-i}, V_{src}^{t-i} \leftarrow \varepsilon_\theta(z_t^{src}, t, \mathcal{P}_{src})$
                
                $Q_{edit}^{t-i}, K_{edit}^{t-i}, V_{edit}^{t-i} \leftarrow \varepsilon_\theta(z_t^{edit}, t, \mathcal{P}_{edit})$
                
                $Attn(Q_{edit}^{t-i},K_{src}^{t-i},V_{src}^{t-i})$
            }
            $z_{t-1}^{src} \leftarrow \varepsilon_\theta(z_t^{src}, t, \mathcal{P}_{src})$

            $z_{t-1}^{edit} \leftarrow \varepsilon_\theta(z_t^{edit}, t, \mathcal{P}_{edit}, Attn)$
        }
        
        $x^{src} \leftarrow \mathcal{D}(z_0^{src}) $
        
        $x^{edit} \leftarrow \mathcal{D}(z_0^{edit}) $
        
        \textbf{Return:} $x^{edit}$

	\caption{Non-Rigid Editing}
	\label{alg:non_rigid}
\end{algorithm}

\begin{algorithm}[h]
	\SetAlgoLined
	\KwIn{A source prompt $\mathcal{P}_{src}$, A editing prompt $\mathcal{P}_{edit}$, a pretrained RoPE-based MMDiT text-to-image model $\varepsilon_\theta$, an image decoder $\mathcal{D}$, total sampling steps $T$ (default $50$), denoising step $B$ (default $45$) to stop value blending, total number of layers $L$, the foreground mask automatically derived by SAM2 $M_{fg}^{sam}$.}
	\KwOut{An edited image $x^{edit}$ aligned with the editing prompt $\mathcal{P}_{edit}$.}
        \textbf{Initialize} $z_T^{src} \sim \mathcal{N}(0, 1)$, $z_T^{edit} \leftarrow z_T^{src}$.
        
	\For{$t=T$ to $T-B+1$}{
            \For{$i=1$ to $L$}{
                $Q_{src}^{t-i}, K_{src}^{t-i}, V_{src}^{t-i} \leftarrow \varepsilon_\theta(z_t^{src}, t, \mathcal{P}_{src})$
                
                $Q_{edit}^{t-i}, K_{edit}^{t-i}, V_{edit}^{t-i} \leftarrow \varepsilon_\theta(z_t^{edit}, t, \mathcal{P}_{edit})$

                $V_{fg}^{t-i}=M_{fg}^{sam}\times V_{src}^{t-i}+(1-M_{fg}^{sam})\times V_{edit}^{t-i}$
                
                $Attn(Q_{edit}^{t-i},K_{edit}^{t-i},V_{fg}^{t-i})$
            }
            $z_{t-1}^{src} \leftarrow \varepsilon_\theta(z_t^{src}, t, \mathcal{P}_{src})$

            $z_{t-1}^{edit} \leftarrow \varepsilon_\theta(z_t^{edit}, t, \mathcal{P}_{edit}, Attn)$
        }

	\For{$t=T-B$ to $1$}{
            $z_{t-1}^{src} \leftarrow \varepsilon_\theta(z_t^{src}, t, \mathcal{P}_{src})$

            $z_{t-1}^{edit} \leftarrow \varepsilon_\theta(z_t^{edit}, t, \mathcal{P}_{edit})$
        }
        
        $x^{src} \leftarrow \mathcal{D}(z_0^{src}) $
        
        $x^{edit} \leftarrow \mathcal{D}(z_0^{edit}) $
        
        \textbf{Return:} $x^{edit}$

	\caption{Background Replacement}
	\label{alg:bg_replace}
\end{algorithm}

\begin{algorithm}[h]
	\SetAlgoLined
	\KwIn{A source prompt $\mathcal{P}_{src}$, A editing prompt $\mathcal{P}_{edit}$, a pretrained RoPE-based MMDiT text-to-image model $\varepsilon_\theta$, an image decoder $\mathcal{D}$, total sampling steps $T$ (default $50$), denoising step $B$ (default $45$) to stop value blending, total number of layers $L$, the coordinate $c$ of the movement direction, the function $MAP$ that maps the source object value to a specified location based on $c$ and copies the unaffected region.}
	\KwOut{An edited image $x^{edit}$ aligned with the editing prompt $\mathcal{P}_{edit}$.}
        \textbf{Initialize} $z_T^{src} \sim \mathcal{N}(0, 1)$, $z_T^{edit} \leftarrow z_T^{src}$.
        
	\For{$t=T$ to $T-B+1$}{
            \For{$i=1$ to $L$}{
                $Q_{src}^{t-i}, K_{src}^{t-i}, V_{src}^{t-i} \leftarrow \varepsilon_\theta(z_t^{src}, t, \mathcal{P}_{src})$
                
                $Q_{edit}^{t-i}, K_{edit}^{t-i}, V_{edit}^{t-i} \leftarrow \varepsilon_\theta(z_t^{edit}, t, \mathcal{P}_{edit})$

                $V_{move}^{t-i}=MAP(V_{edit}^{t-i}, V_{src}^{t-i}, c)$
                
                $Attn(Q_{edit}^{t-i},K_{edit}^{t-i},V_{move}^{t-i})$
            }
            $z_{t-1}^{src} \leftarrow \varepsilon_\theta(z_t^{src}, t, \mathcal{P}_{src})$

            $z_{t-1}^{edit} \leftarrow \varepsilon_\theta(z_t^{edit}, t, \mathcal{P}_{edit}, Attn)$
        }

	\For{$t=T-B$ to $1$}{
            $z_{t-1}^{src} \leftarrow \varepsilon_\theta(z_t^{src}, t, \mathcal{P}_{src})$

            $z_{t-1}^{edit} \leftarrow \varepsilon_\theta(z_t^{edit}, t, \mathcal{P}_{edit})$
        }
        
        $x^{src} \leftarrow \mathcal{D}(z_0^{src}) $
        
        $x^{edit} \leftarrow \mathcal{D}(z_0^{edit}) $
        
        \textbf{Return:} $x^{edit}$

	\caption{Object Movement}
	\label{alg:object_move}
\end{algorithm}

\begin{algorithm}[h]
	\SetAlgoLined
	\KwIn{A source prompt $\mathcal{P}_{src}$, A editing prompt $\mathcal{P}_{edit}$, a pretrained RoPE-based MMDiT text-to-image model $\varepsilon_\theta$, an image decoder $\mathcal{D}$, total sampling steps $T$ (default $50$), denoising step $B$ (default $45$) to stop value blending, total number of layers $L$, the paste coordinates $c$ of the original image on the higher-resolution edited image, the function $PASTE$ copies the value of the original image to the corresponding position in the edited image based on $c$.}
	\KwOut{An edited image $x^{edit}$ aligned with the editing prompt $\mathcal{P}_{edit}$.}
        \textbf{Initialize} $z_T^{src} \sim \mathcal{N}(0, 1)$, $z_T^{edit} \leftarrow z_T^{src}$.
        
	\For{$t=T$ to $T-B+1$}{
            \For{$i=1$ to $L$}{
                $Q_{src}^{t-i}, K_{src}^{t-i}, V_{src}^{t-i} \leftarrow \varepsilon_\theta(z_t^{src}, t, \mathcal{P}_{src})$
                
                $Q_{edit}^{t-i}, K_{edit}^{t-i}, V_{edit}^{t-i} \leftarrow \varepsilon_\theta(z_t^{edit}, t, \mathcal{P}_{edit})$

                $V_{out}^{t-i}=PASTE(V_{edit}^{t-i}, V_{src}^{t-i}, c)$
                
                $Attn(Q_{edit}^{t-i},K_{edit}^{t-i},V_{out}^{t-i})$
            }
            $z_{t-1}^{src} \leftarrow \varepsilon_\theta(z_t^{src}, t, \mathcal{P}_{src})$

            $z_{t-1}^{edit} \leftarrow \varepsilon_\theta(z_t^{edit}, t, \mathcal{P}_{edit}, Attn)$
        }

	\For{$t=T-B$ to $1$}{
            $z_{t-1}^{src} \leftarrow \varepsilon_\theta(z_t^{src}, t, \mathcal{P}_{src})$

            $z_{t-1}^{edit} \leftarrow \varepsilon_\theta(z_t^{edit}, t, \mathcal{P}_{edit})$
        }
        
        $x^{src} \leftarrow \mathcal{D}(z_0^{src}) $
        
        $x^{edit} \leftarrow \mathcal{D}(z_0^{edit}) $
        
        \textbf{Return:} $x^{edit}$

	\caption{Outpainting}
	\label{alg:outpainting}
\end{algorithm}

\section{More Qualitative Results}

\begin{figure*}[t]
	\begin{center}
		\setlength{\tabcolsep}{0.5pt}
		\begin{tabular}{m{0.3cm}<{\centering}m{3.3cm}<{\centering}m{3.3cm}<{\centering}m{3.3cm}<{\centering}m{3.3cm}<{\centering}m{3.3cm}<{\centering}}
   
			\raisebox{0.3cm}{\rotatebox[origin=c]{90}{\normalsize{{Source}}}}
			&\includegraphics[width=3.15cm]{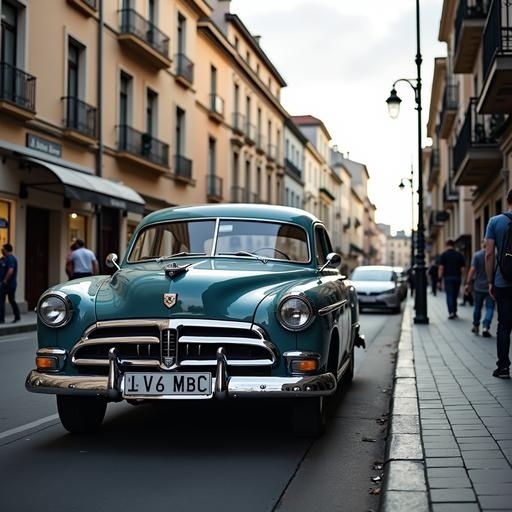}
			&\includegraphics[width=3.15cm]{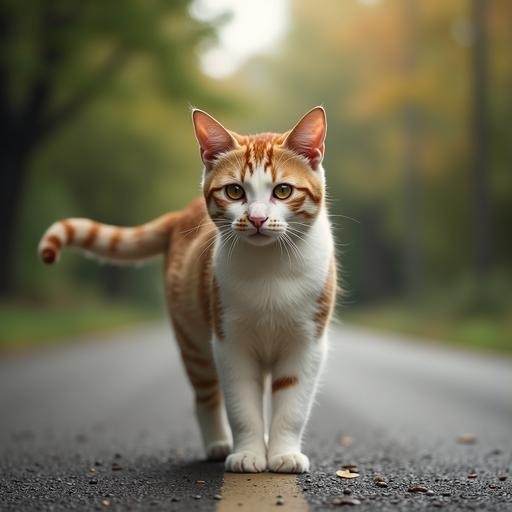}
			&\includegraphics[width=3.15cm]{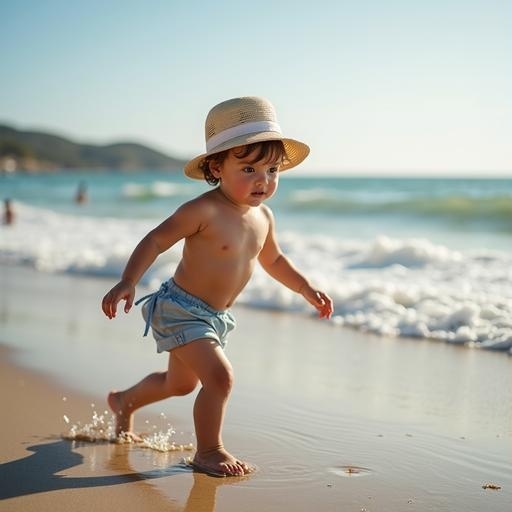}
			&\includegraphics[width=3.15cm]{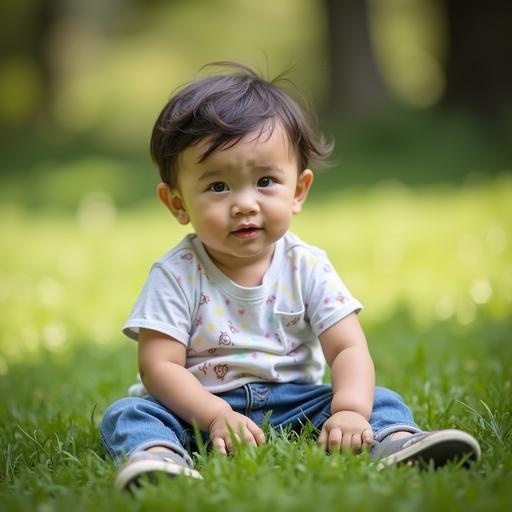}
			&\includegraphics[width=3.15cm]{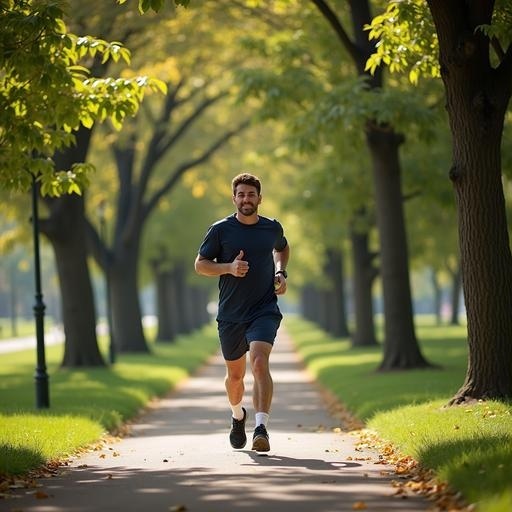}
			\\

			\raisebox{0.4cm}{\rotatebox[origin=c]{90}{\normalsize{{StableFlow}}}}
			&\includegraphics[width=3.15cm]{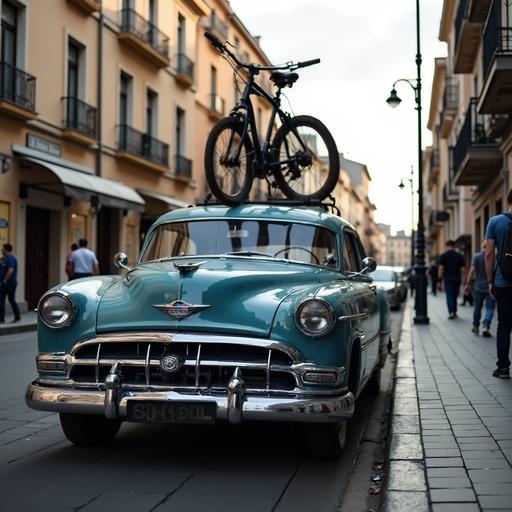}
			&\includegraphics[width=3.15cm]{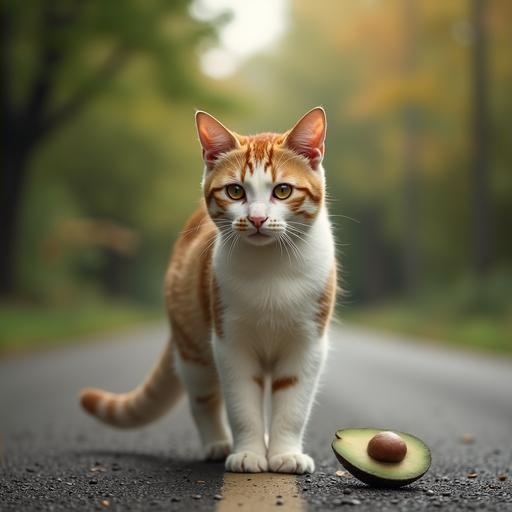}
			&\includegraphics[width=3.15cm]{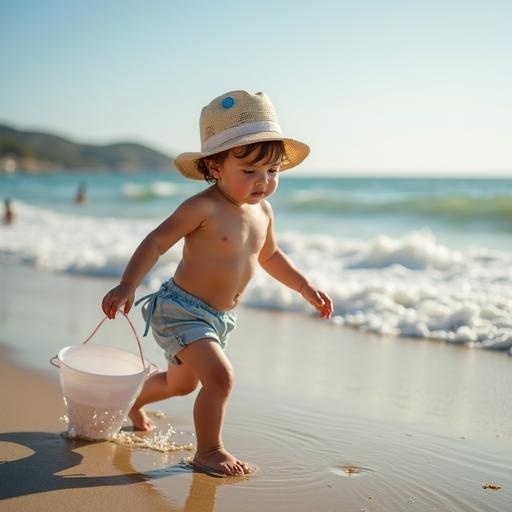}
			&\includegraphics[width=3.15cm]{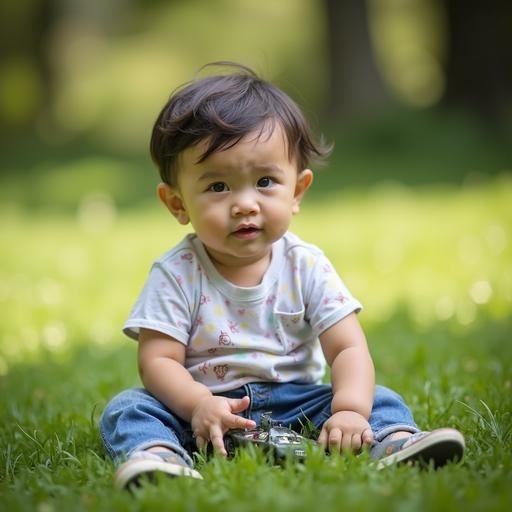}
			&\includegraphics[width=3.15cm]{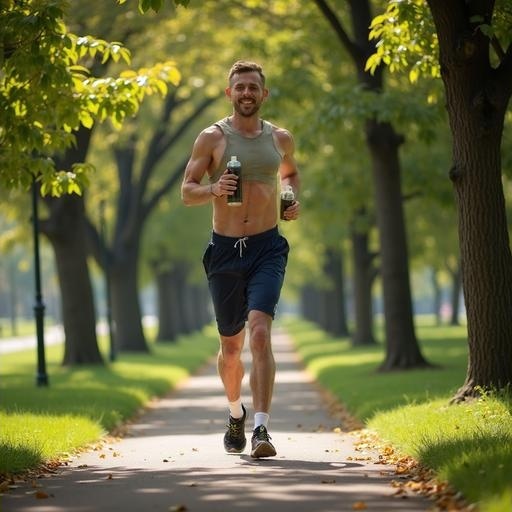}
			\\

			\raisebox{0.5cm}{\rotatebox[origin=c]{90}{\normalsize{{TamingRF}}}}
			&\includegraphics[width=3.15cm]{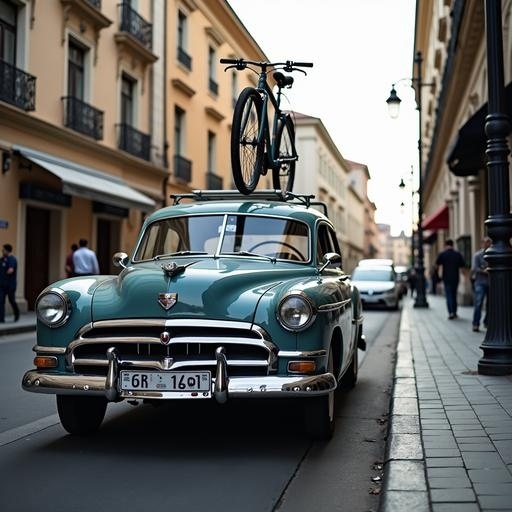}
			&\includegraphics[width=3.15cm]{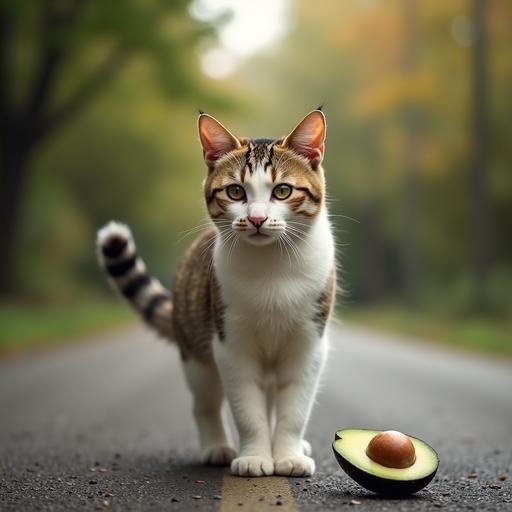}
			&\includegraphics[width=3.15cm]{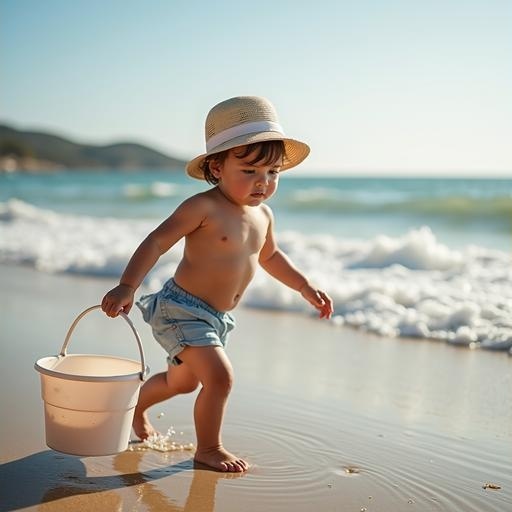}
			&\includegraphics[width=3.15cm]{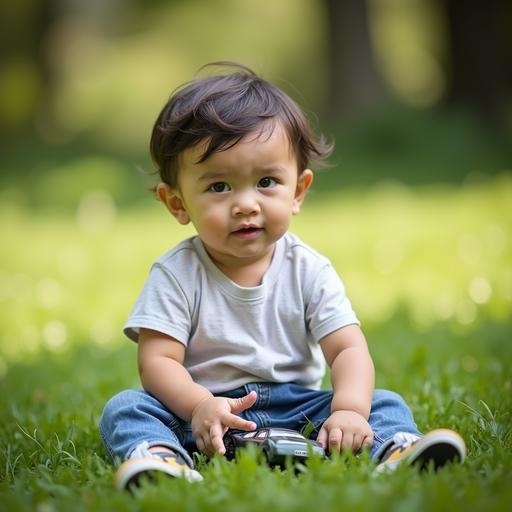}
			&\includegraphics[width=3.15cm]{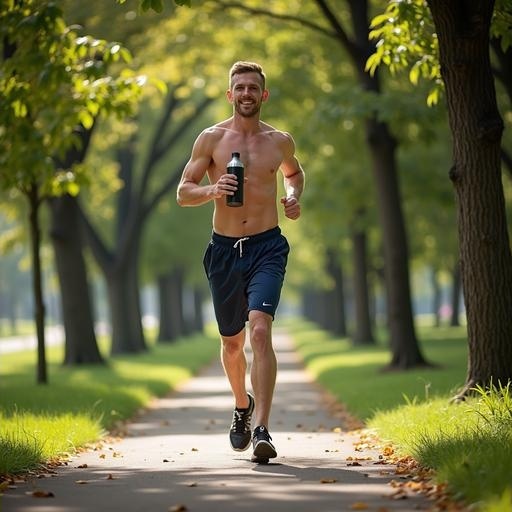}
			\\

			\raisebox{0.7cm}{\rotatebox[origin=c]{90}{\normalsize{{MagicBrush}}}}
			&\includegraphics[width=3.15cm]{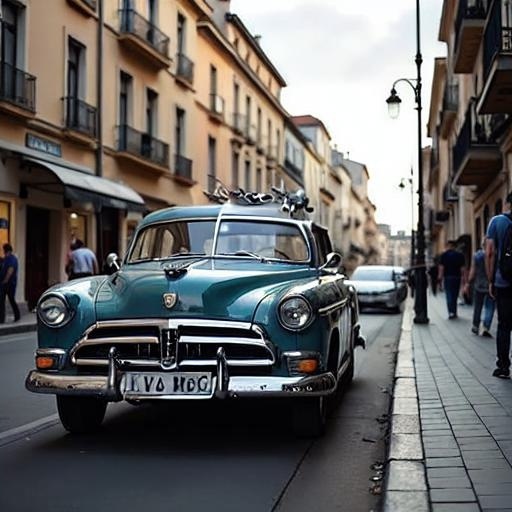}
			&\includegraphics[width=3.15cm]{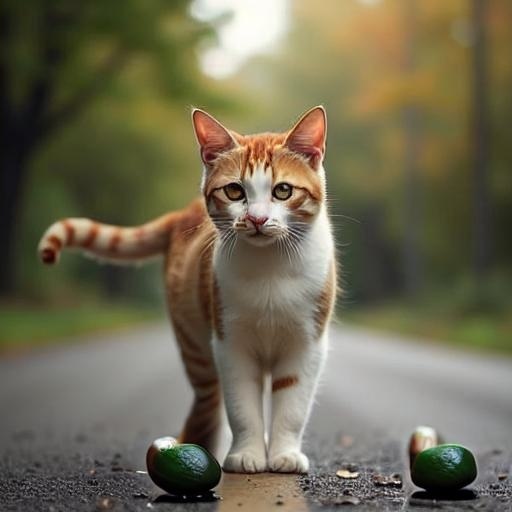}
			&\includegraphics[width=3.15cm]{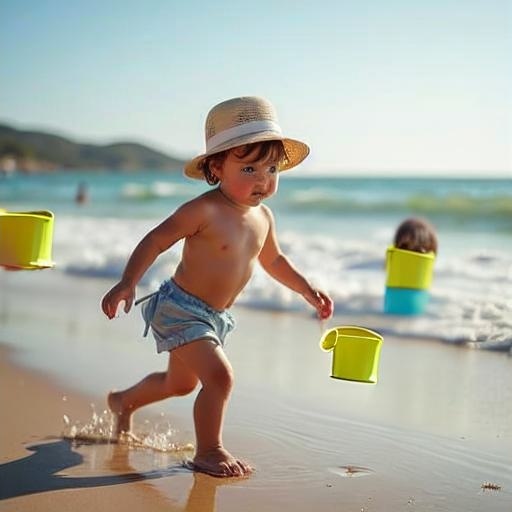}
			&\includegraphics[width=3.15cm]{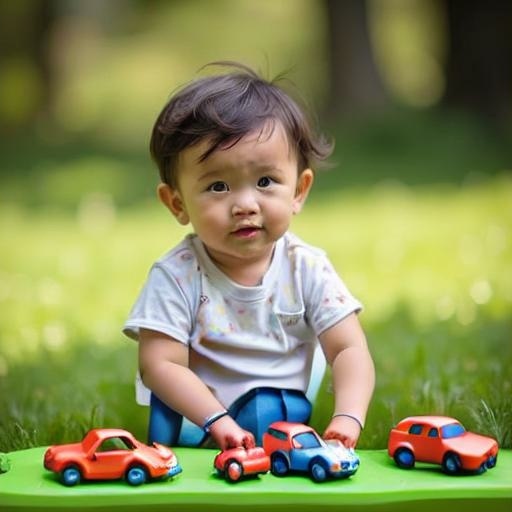}
			&\includegraphics[width=3.15cm]{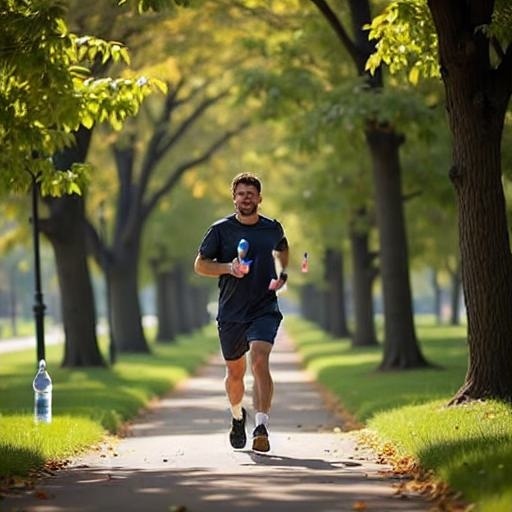}
			\\

			\raisebox{0.4cm}{\rotatebox[origin=c]{90}{\normalsize{{OmniGen}}}}
			&\includegraphics[width=3.15cm]{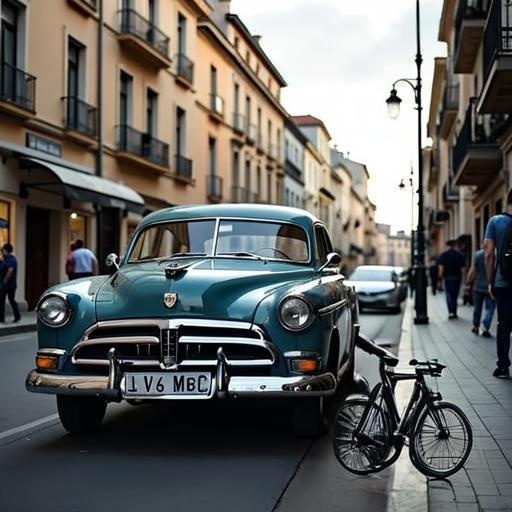}
			&\includegraphics[width=3.15cm]{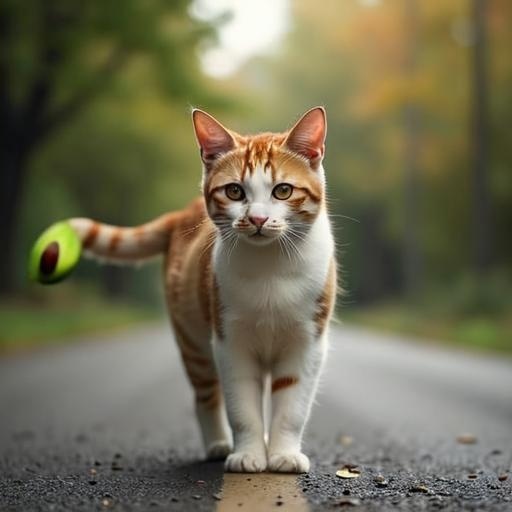}
			&\includegraphics[width=3.15cm]{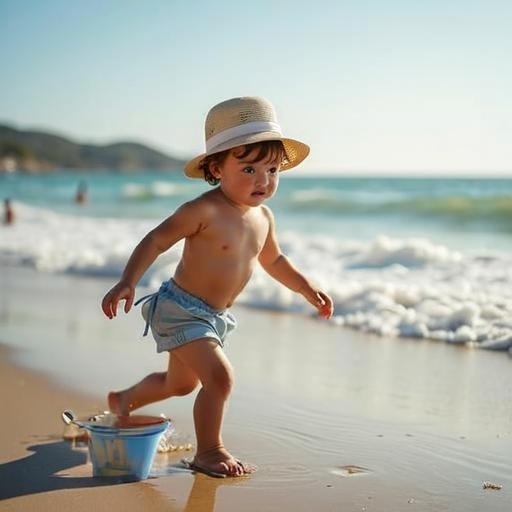}
			&\includegraphics[width=3.15cm]{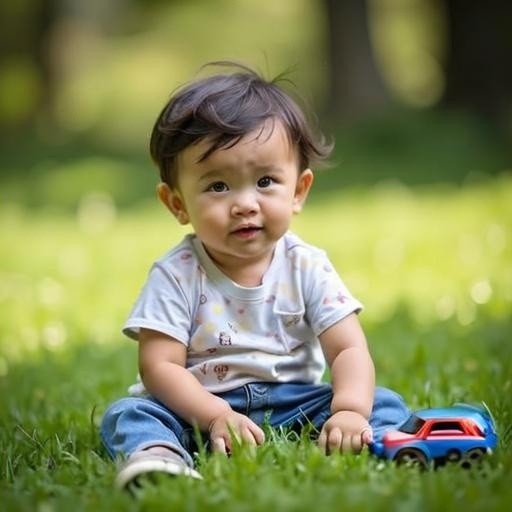}
			&\includegraphics[width=3.15cm]{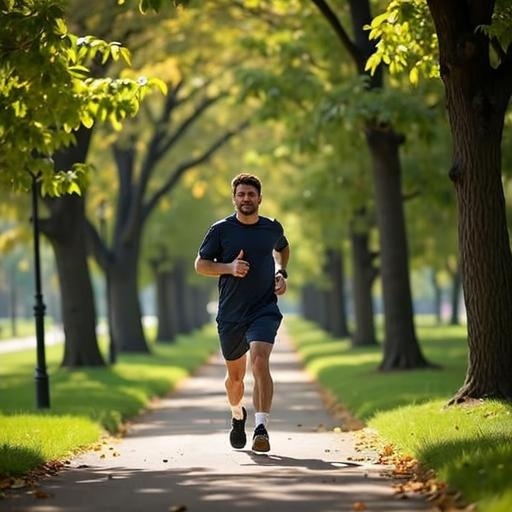}
			\\

			\raisebox{0.15cm}{\rotatebox[origin=c]{90}{\normalsize{{Ours}}}}
			&\includegraphics[width=3.15cm]{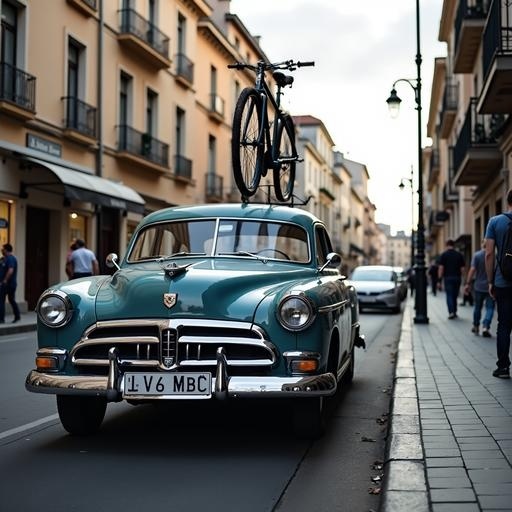}
			&\includegraphics[width=3.15cm]{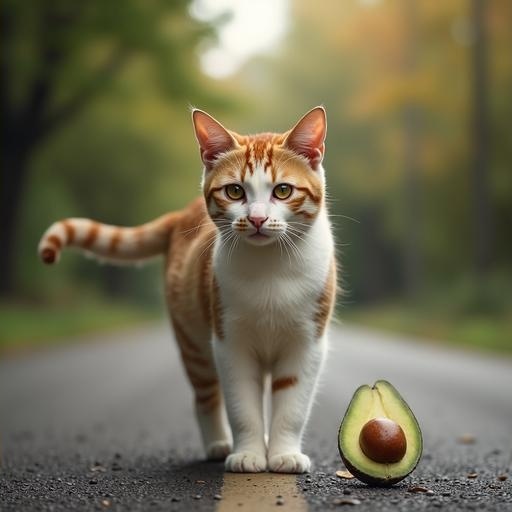}
			&\includegraphics[width=3.15cm]{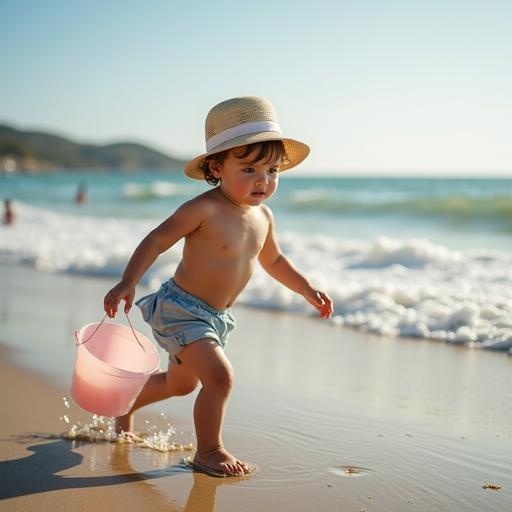}
			&\includegraphics[width=3.15cm]{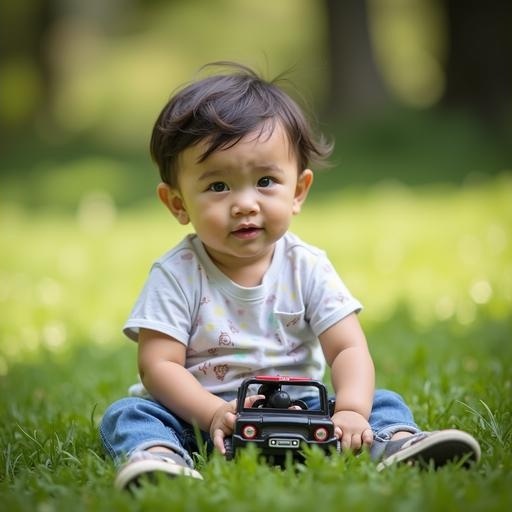}
			&\includegraphics[width=3.15cm]{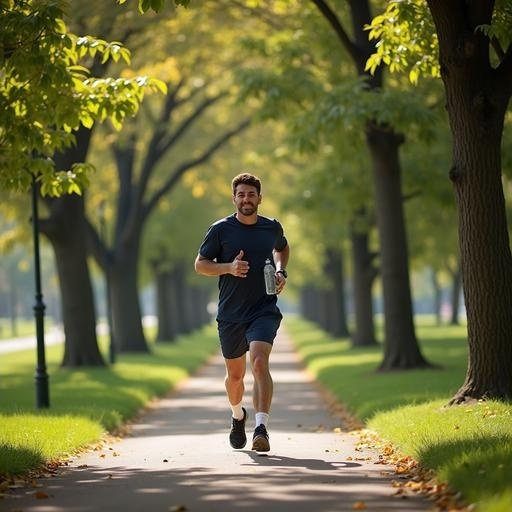}
			\\
			 & \small{\textit{``\textcolor{myblue}{Add a Bicycle}"}} & \small{\textit{``\textcolor{myblue}{Add an Avocado}"}}& \small{\textit{``\textcolor{myblue}{Add a Bucket}"}}& \small{\textit{``\textcolor{myblue}{Add a Toy Car}"}}& \small{\textit{``\textcolor{myblue}{Add a Water Bottle}"}}
			\\
			
		\end{tabular}
	\end{center}
	\caption{Qualitative comparison on the object addition task with training-free methods StableFlow~\cite{avrahami2024stable} and TamingRF~\cite{wang2024taming}, as well as general image editing models MagicBrush~\cite{zhang2023magicbrush} and OmniGen~\cite{xiao2024omnigen}. Our method not only achieves high-quality editing results but also demonstrates the best ability to preserve irrelevant regions.}
	\label{fig:quail_comparsion_add_object}
\end{figure*}

\begin{figure*}[t]
	\begin{center}
		\setlength{\tabcolsep}{0.5pt}
		\begin{tabular}{m{0.3cm}<{\centering}m{3.3cm}<{\centering}m{3.3cm}<{\centering}m{3.3cm}<{\centering}m{3.3cm}<{\centering}m{3.3cm}<{\centering}}
   
			\raisebox{0.3cm}{\rotatebox[origin=c]{90}{\normalsize{{Source}}}}
			&\includegraphics[width=3.15cm]{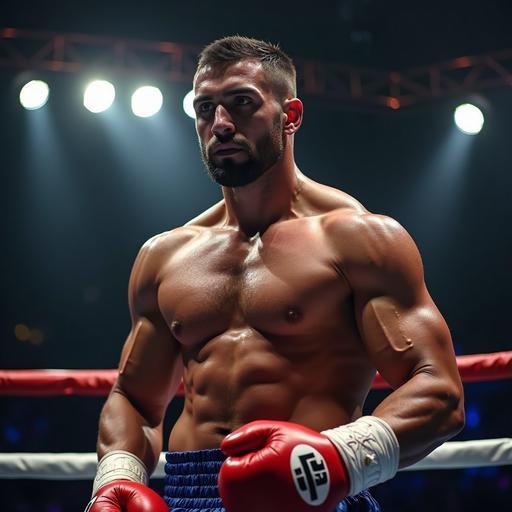}
			&\includegraphics[width=3.15cm]{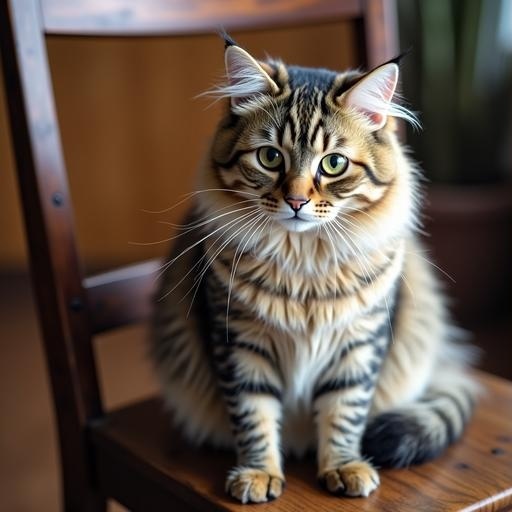}
			&\includegraphics[width=3.15cm]{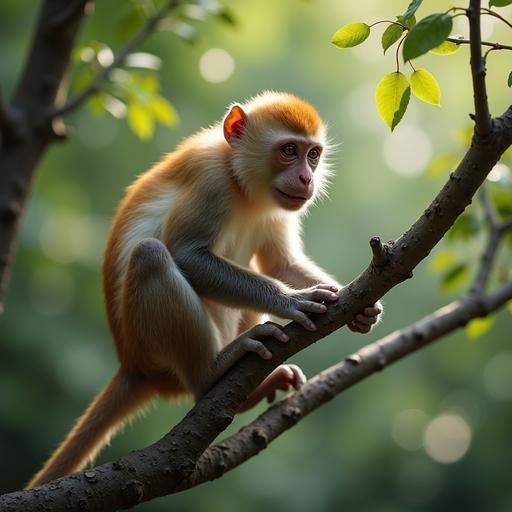}
			&\includegraphics[width=3.15cm]{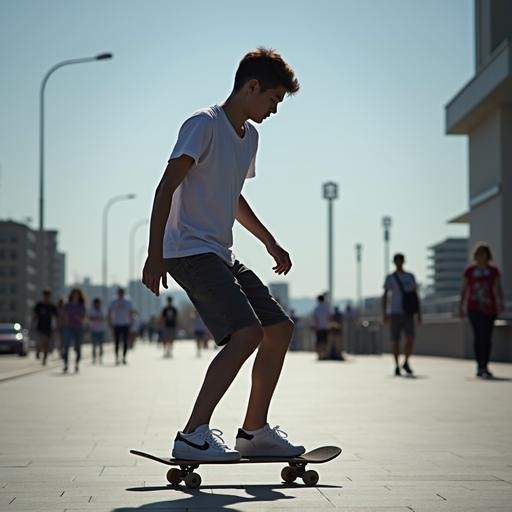}
			&\includegraphics[width=3.15cm]{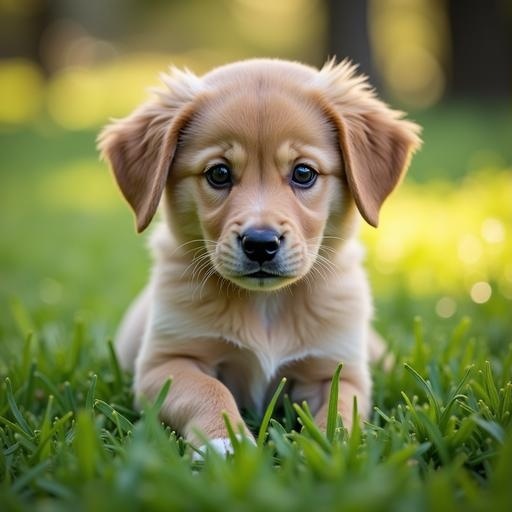}
			\\

			\raisebox{0.4cm}{\rotatebox[origin=c]{90}{\normalsize{{StableFlow}}}}
			&\includegraphics[width=3.15cm]{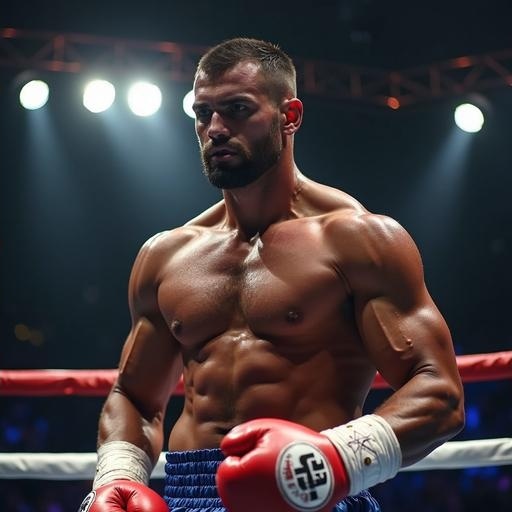}
			&\includegraphics[width=3.15cm]{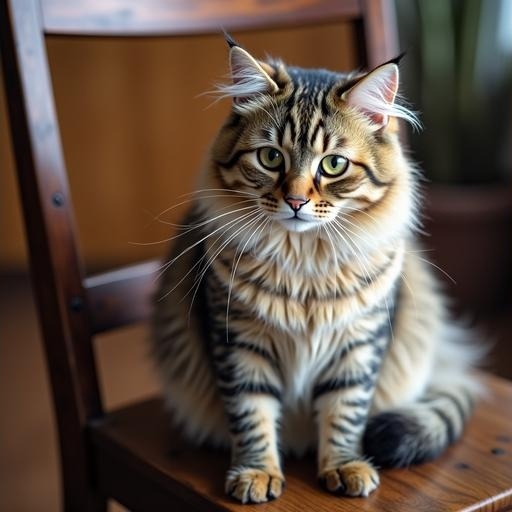}
			&\includegraphics[width=3.15cm]{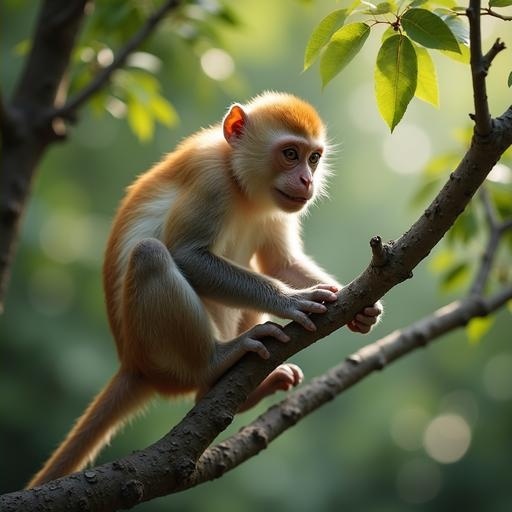}
			&\includegraphics[width=3.15cm]{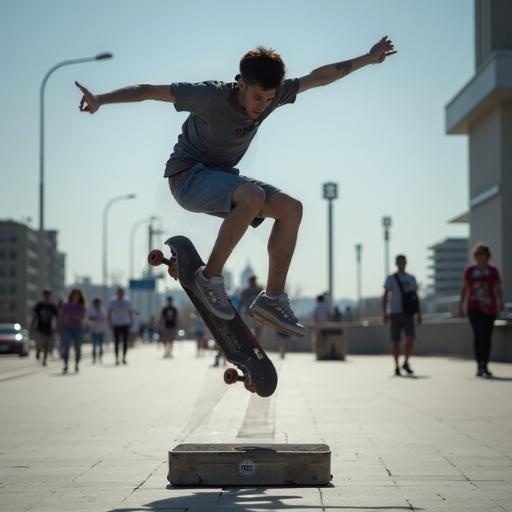}
			&\includegraphics[width=3.15cm]{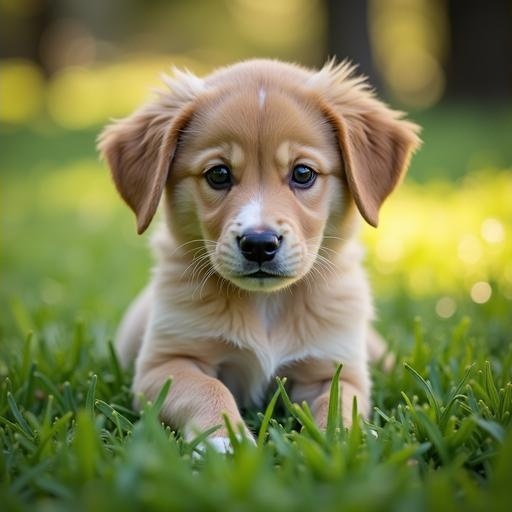}
			\\

			\raisebox{0.5cm}{\rotatebox[origin=c]{90}{\normalsize{{TamingRF}}}}
			&\includegraphics[width=3.15cm]{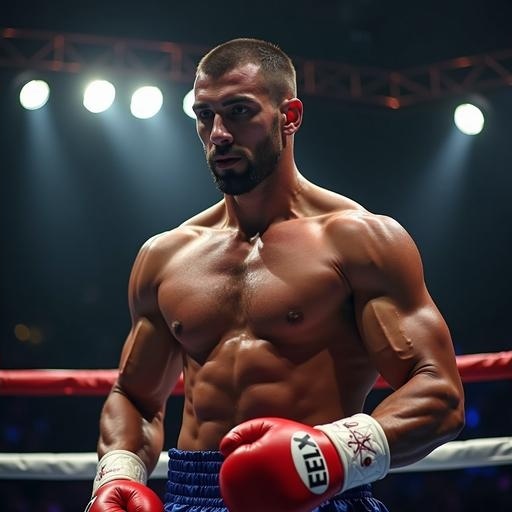}
			&\includegraphics[width=3.15cm]{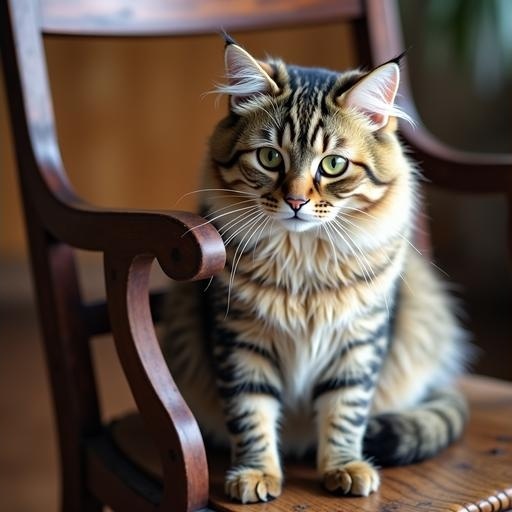}
			&\includegraphics[width=3.15cm]{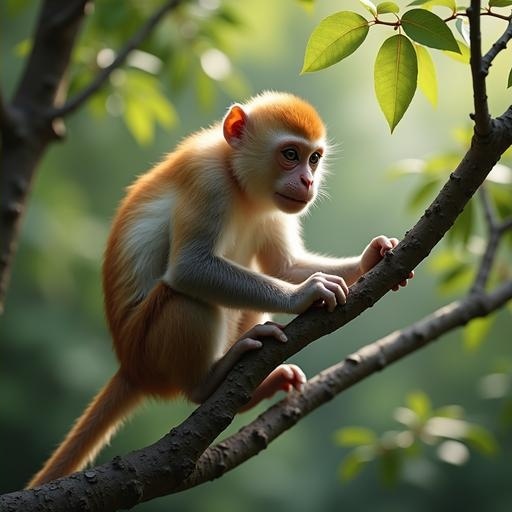}
			&\includegraphics[width=3.15cm]{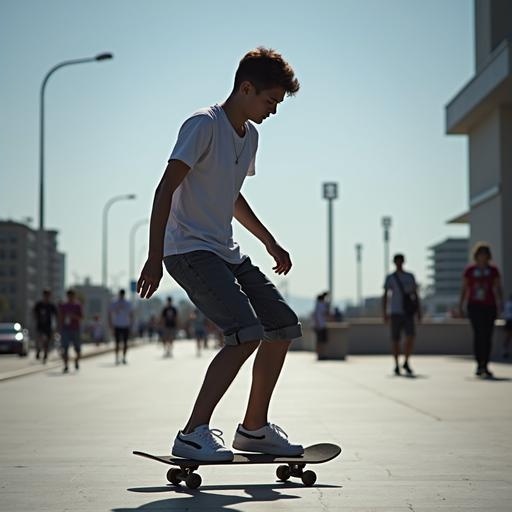}
			&\includegraphics[width=3.15cm]{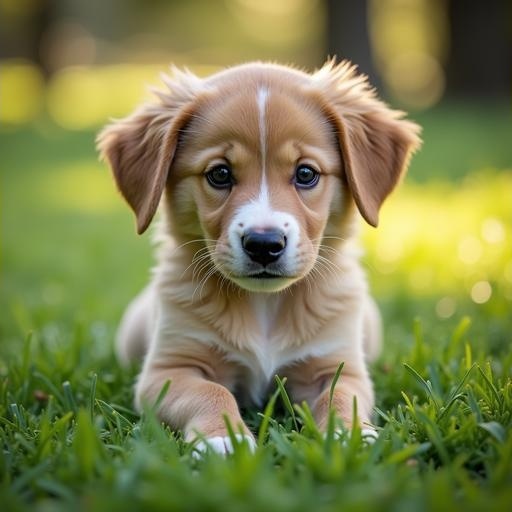}
			\\

			\raisebox{0.7cm}{\rotatebox[origin=c]{90}{\normalsize{{MagicBrush}}}}
			&\includegraphics[width=3.15cm]{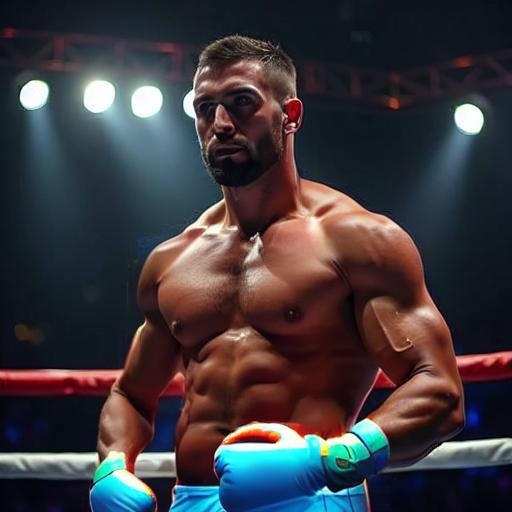}
			&\includegraphics[width=3.15cm]{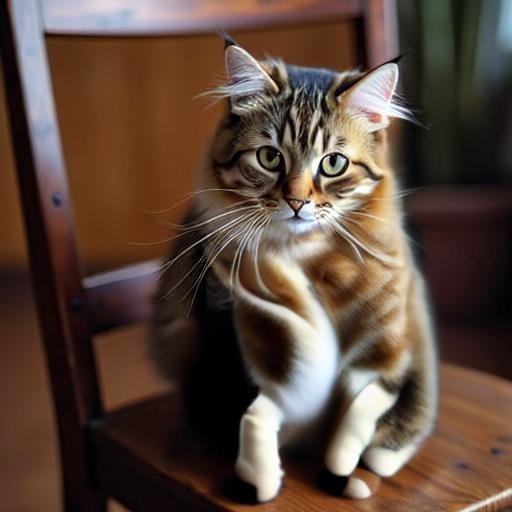}
			&\includegraphics[width=3.15cm]{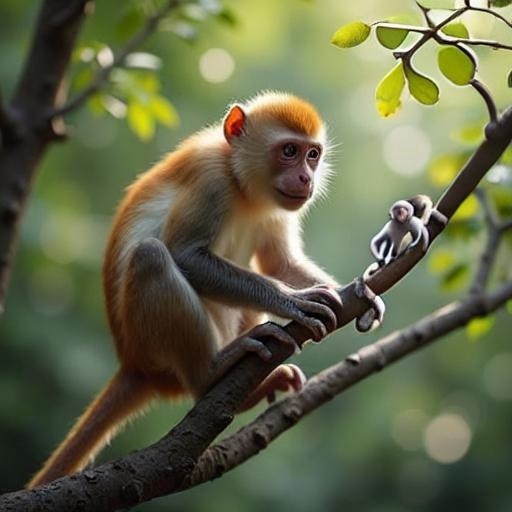}
			&\includegraphics[width=3.15cm]{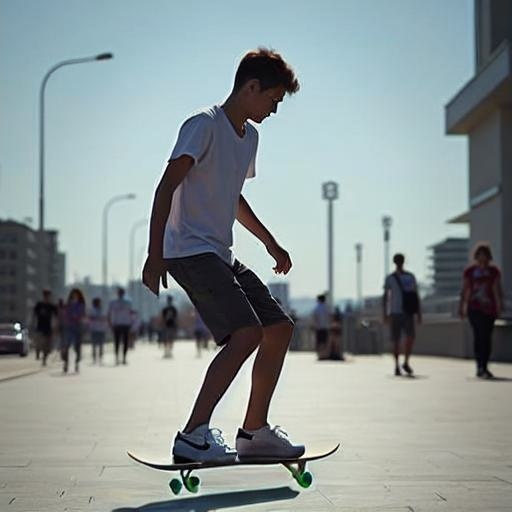}
			&\includegraphics[width=3.15cm]{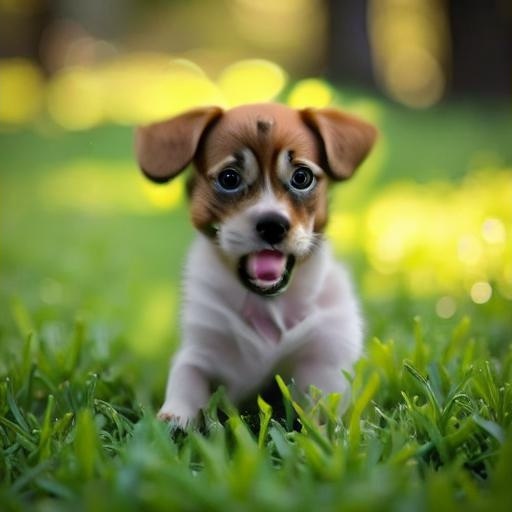}
			\\

			\raisebox{0.4cm}{\rotatebox[origin=c]{90}{\normalsize{{OmniGen}}}}
			&\includegraphics[width=3.15cm]{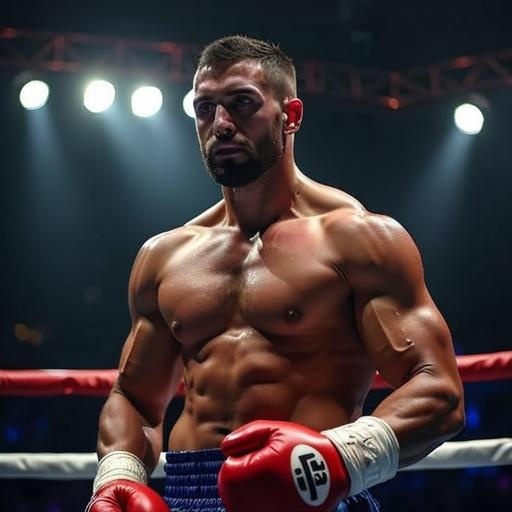}
			&\includegraphics[width=3.15cm]{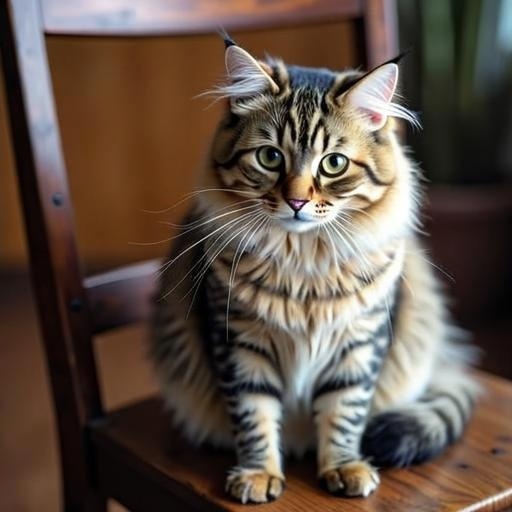}
			&\includegraphics[width=3.15cm]{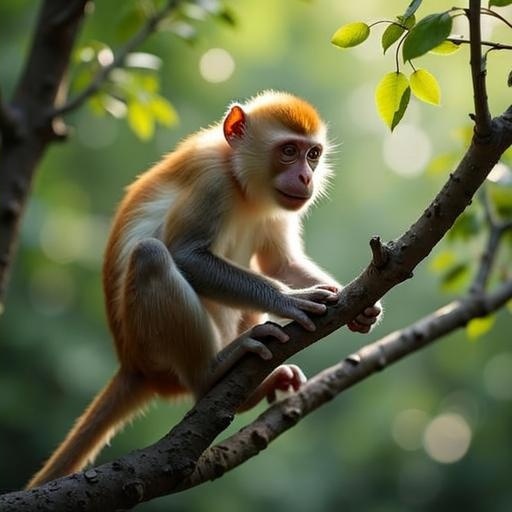}
			&\includegraphics[width=3.15cm]{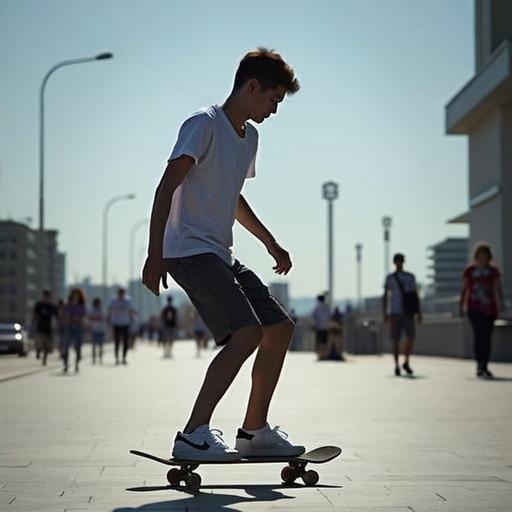}
			&\includegraphics[width=3.15cm]{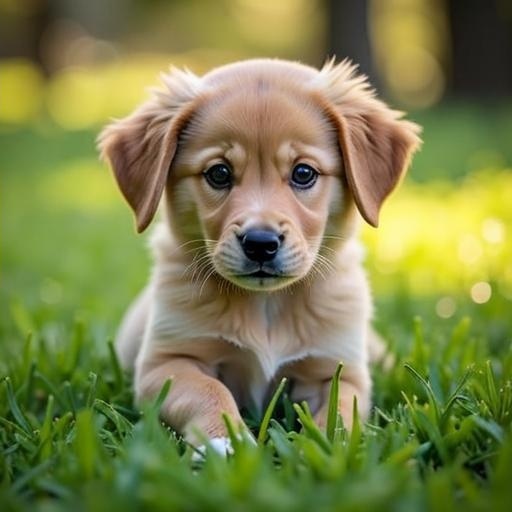}
			\\

			\raisebox{0.15cm}{\rotatebox[origin=c]{90}{\normalsize{{Ours}}}}
			&\includegraphics[width=3.15cm]{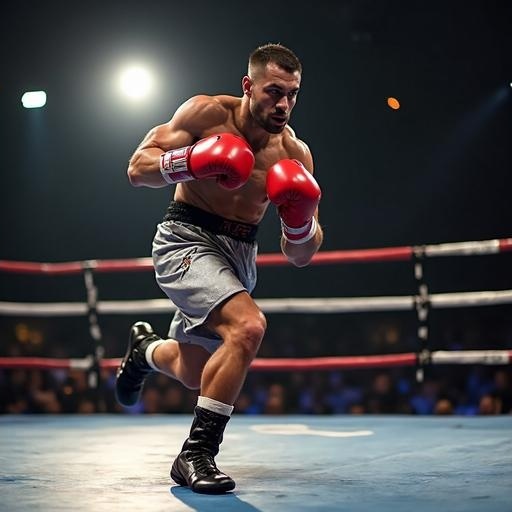}
			&\includegraphics[width=3.15cm]{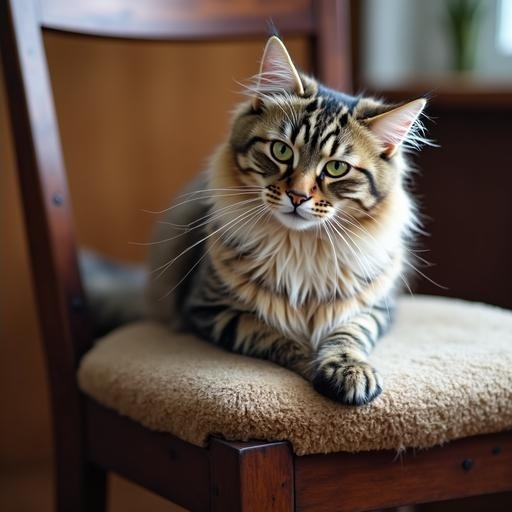}
			&\includegraphics[width=3.15cm]{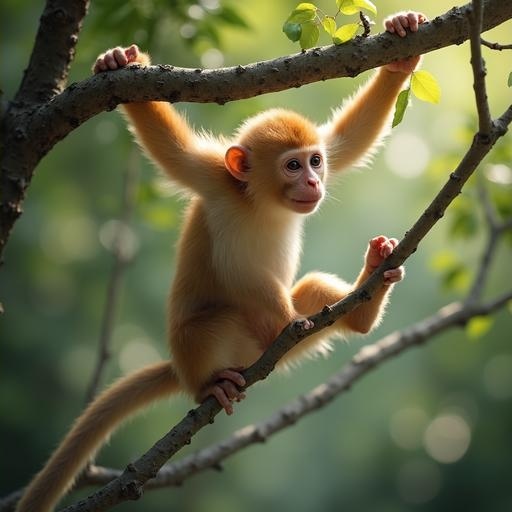}
			&\includegraphics[width=3.15cm]{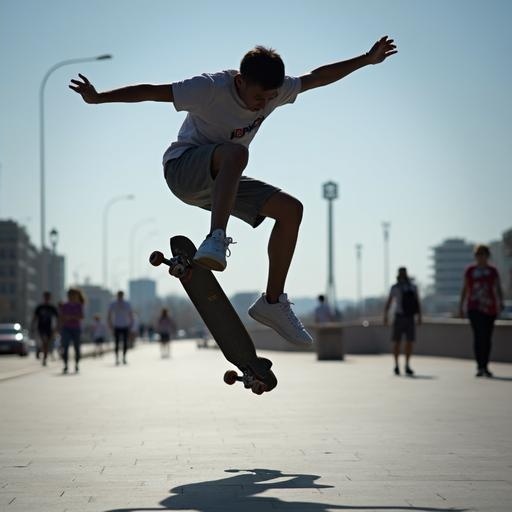}
			&\includegraphics[width=3.15cm]{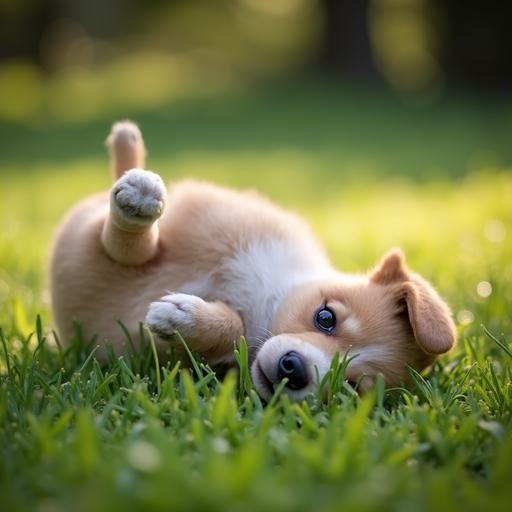}
			\\
			 & \small{\textit{``\textcolor{myblue}{Dodging}"}} & \small{\textit{``\textcolor{myblue}{Stretching}"}}& \small{\textit{``\textcolor{myblue}{Swinging}"}}& \small{\textit{``\textcolor{myblue}{Flipping}"}}& \small{\textit{``\textcolor{myblue}{Rolling}"}}
			\\
			
		\end{tabular}
	\end{center}
	\caption{Qualitative comparison on the non-rigid editing task with training-free methods StableFlow~\cite{avrahami2024stable} and TamingRF~\cite{wang2024taming}, as well as general image editing models MagicBrush~\cite{zhang2023magicbrush} and OmniGen~\cite{xiao2024omnigen}. Our method effectively balances object deformation and appearance transfer.}
	\label{fig:quail_comparsion_non_rigid}
\end{figure*}

\begin{figure*}[t]
	\begin{center}
		\setlength{\tabcolsep}{0.5pt}
		\begin{tabular}{m{0.3cm}<{\centering}m{3.3cm}<{\centering}m{3.3cm}<{\centering}m{3.3cm}<{\centering}m{3.3cm}<{\centering}m{3.3cm}<{\centering}}
   
			\raisebox{0.3cm}{\rotatebox[origin=c]{90}{\normalsize{{Source}}}}
			&\includegraphics[width=3.15cm]{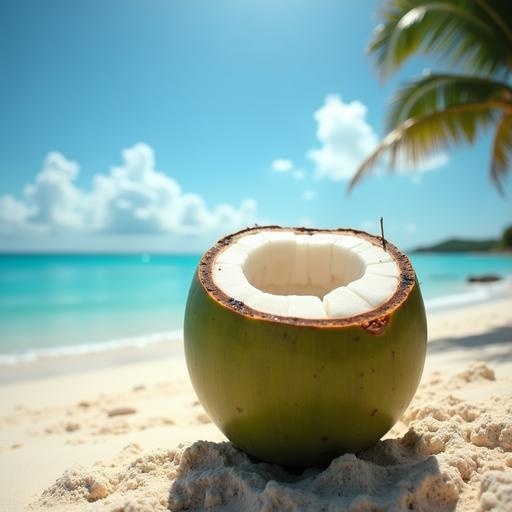}
			&\includegraphics[width=3.15cm]{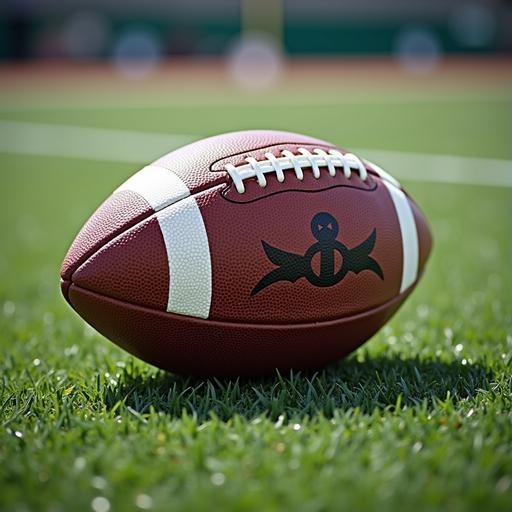}
			&\includegraphics[width=3.15cm]{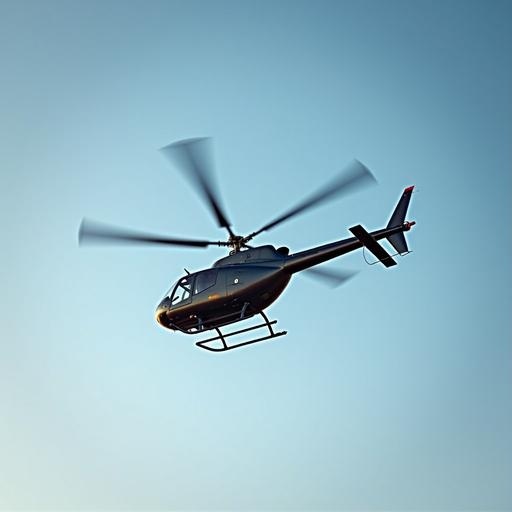}
			&\includegraphics[width=3.15cm]{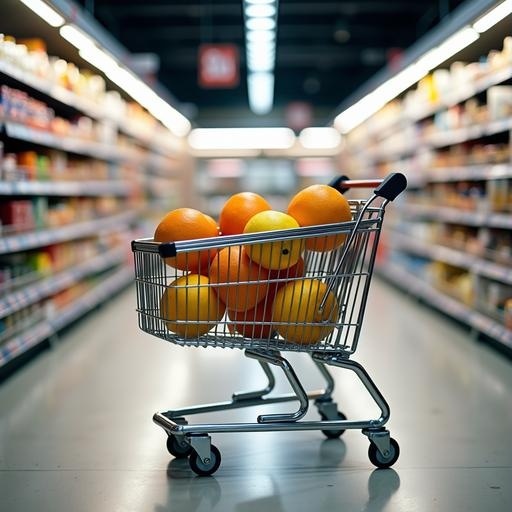}
			&\includegraphics[width=3.15cm]{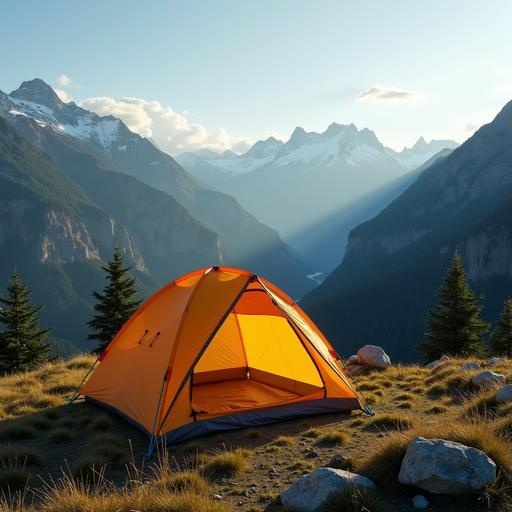}
			\\

			\raisebox{0.4cm}{\rotatebox[origin=c]{90}{\normalsize{{StableFlow}}}}
			&\includegraphics[width=3.15cm]{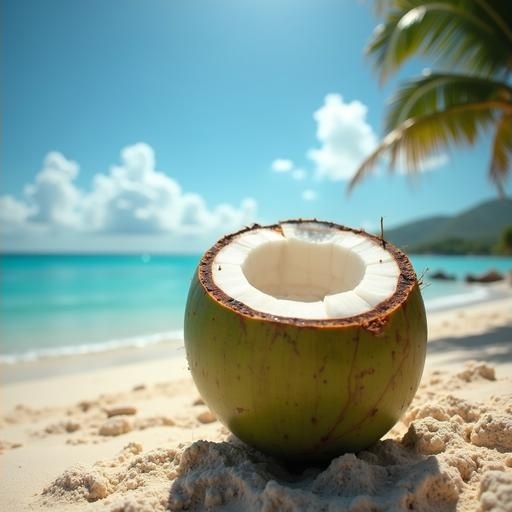}
			&\includegraphics[width=3.15cm]{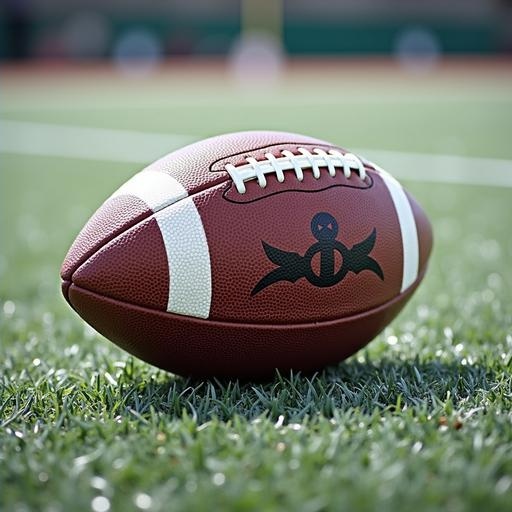}
			&\includegraphics[width=3.15cm]{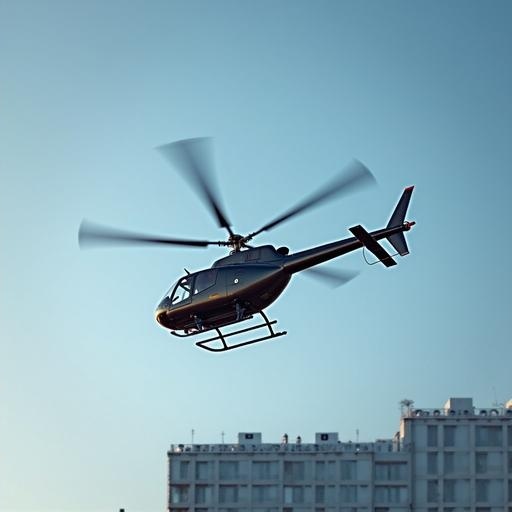}
			&\includegraphics[width=3.15cm]{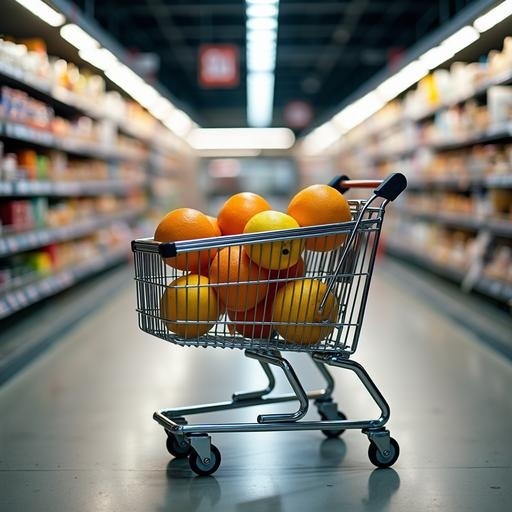}
			&\includegraphics[width=3.15cm]{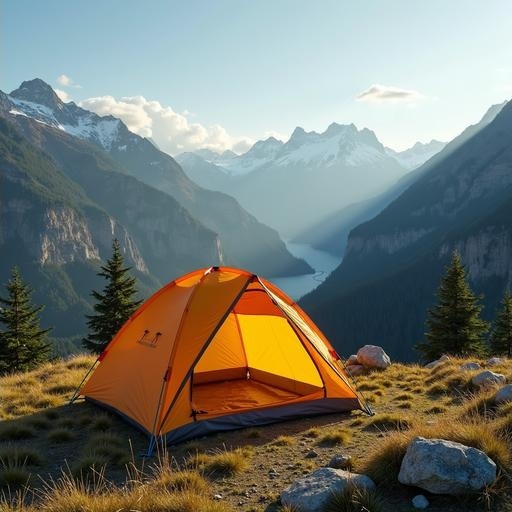}
			\\

			\raisebox{0.5cm}{\rotatebox[origin=c]{90}{\normalsize{{TamingRF}}}}
			&\includegraphics[width=3.15cm]{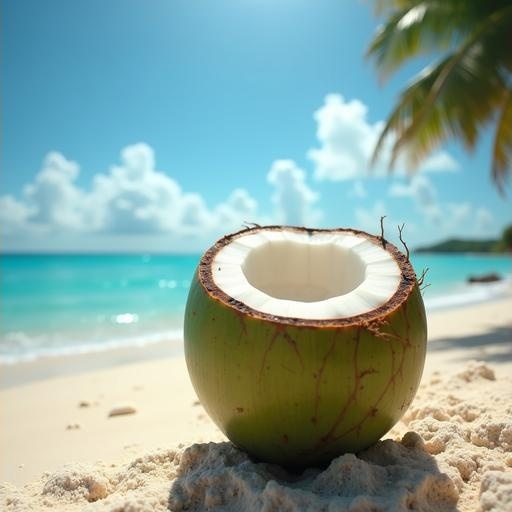}
			&\includegraphics[width=3.15cm]{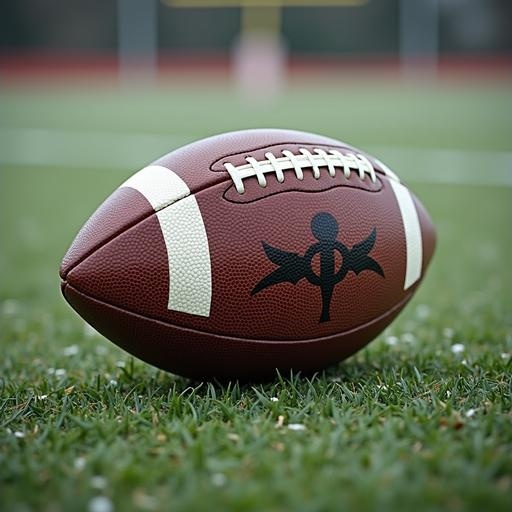}
			&\includegraphics[width=3.15cm]{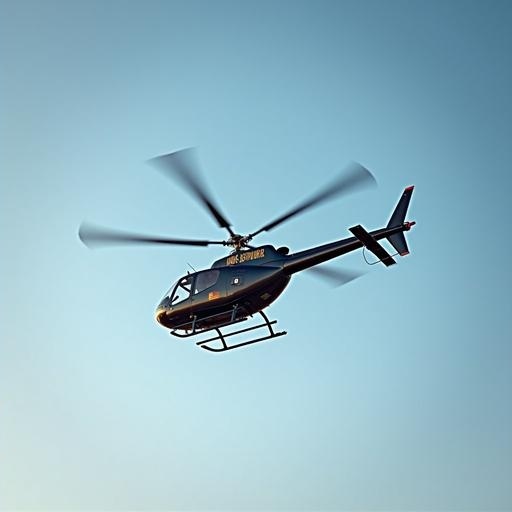}
			&\includegraphics[width=3.15cm]{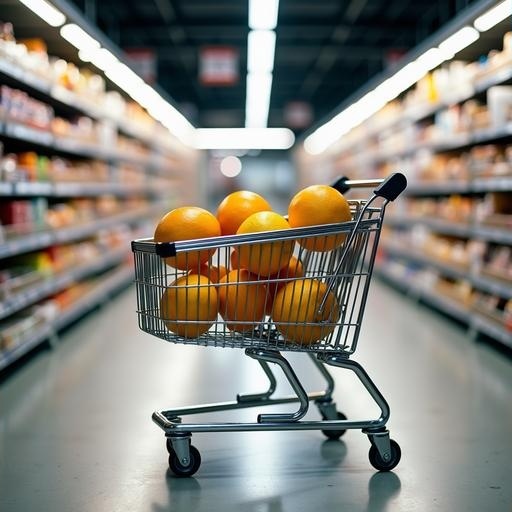}
			&\includegraphics[width=3.15cm]{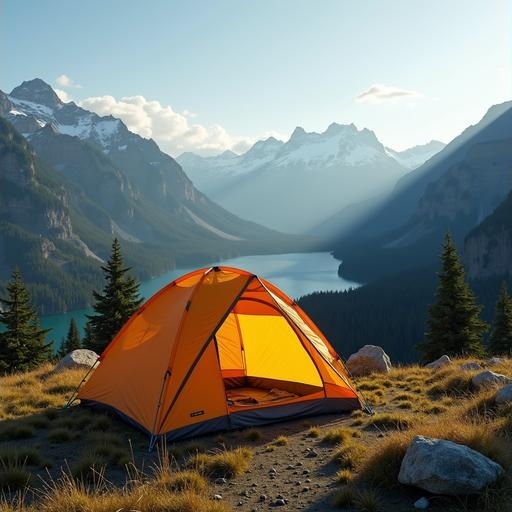}
			\\

			\raisebox{0.7cm}{\rotatebox[origin=c]{90}{\normalsize{{MagicBrush}}}}
			&\includegraphics[width=3.15cm]{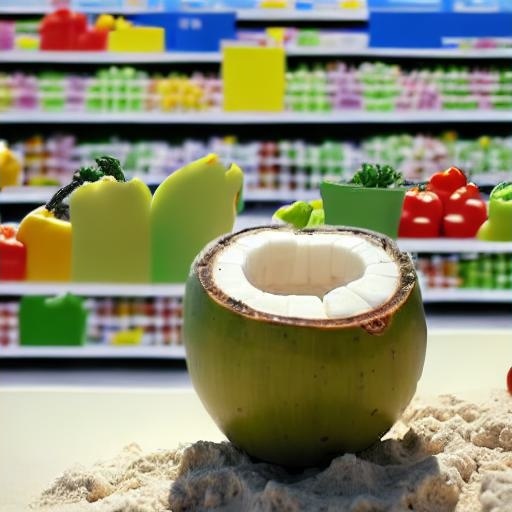}
			&\includegraphics[width=3.15cm]{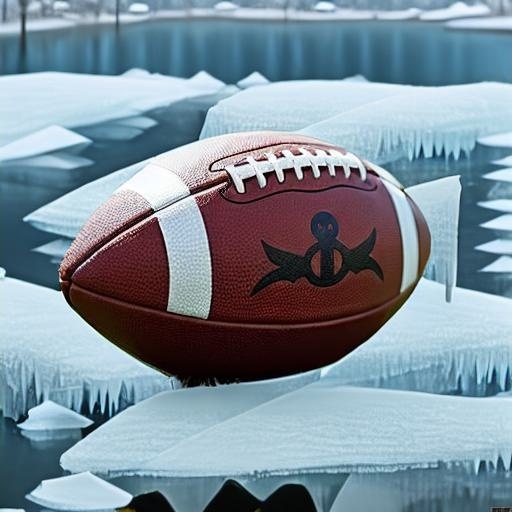}
			&\includegraphics[width=3.15cm]{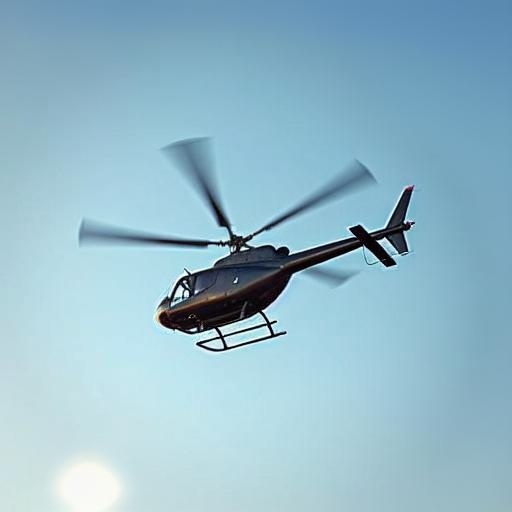}
			&\includegraphics[width=3.15cm]{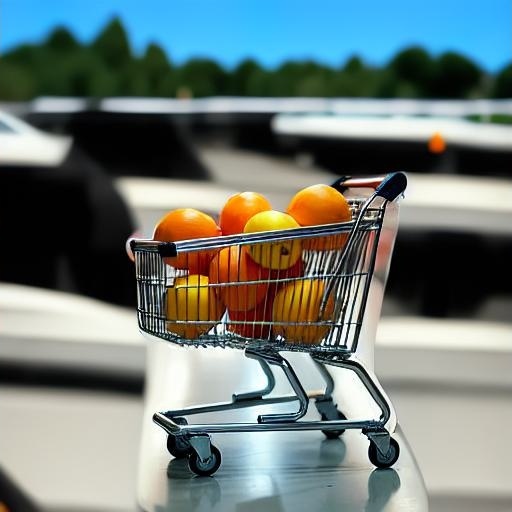}
			&\includegraphics[width=3.15cm]{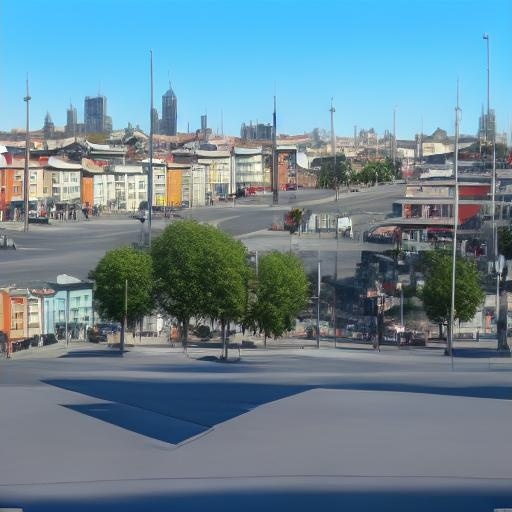}
			\\

			\raisebox{0.4cm}{\rotatebox[origin=c]{90}{\normalsize{{OmniGen}}}}
			&\includegraphics[width=3.15cm]{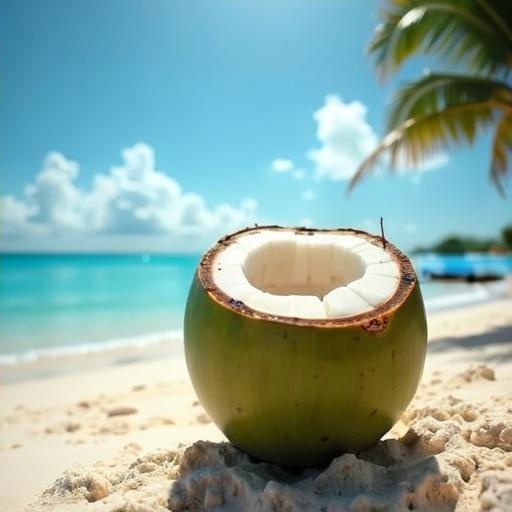}
			&\includegraphics[width=3.15cm]{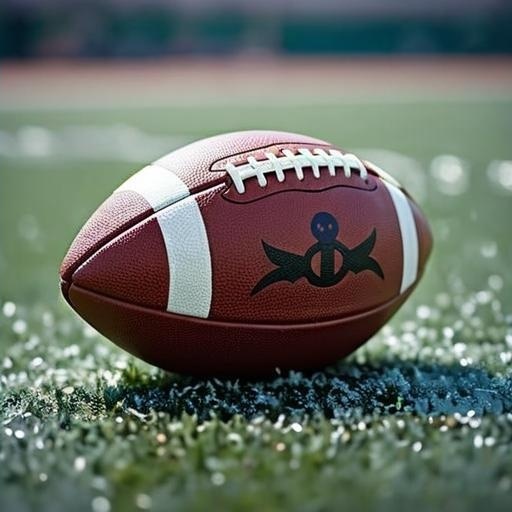}
			&\includegraphics[width=3.15cm]{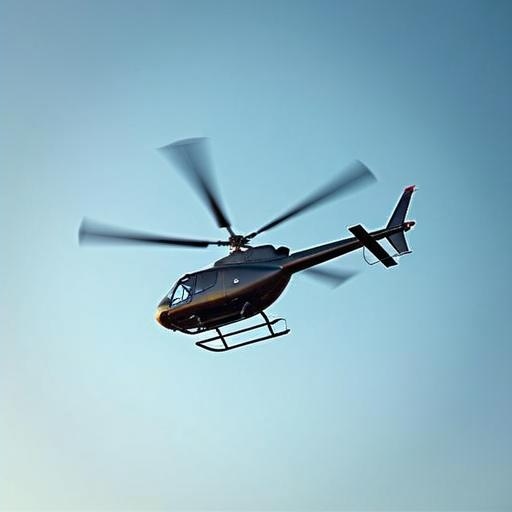}
			&\includegraphics[width=3.15cm]{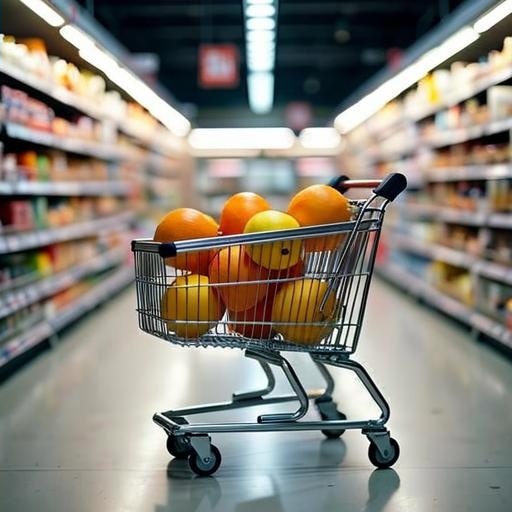}
			&\includegraphics[width=3.15cm]{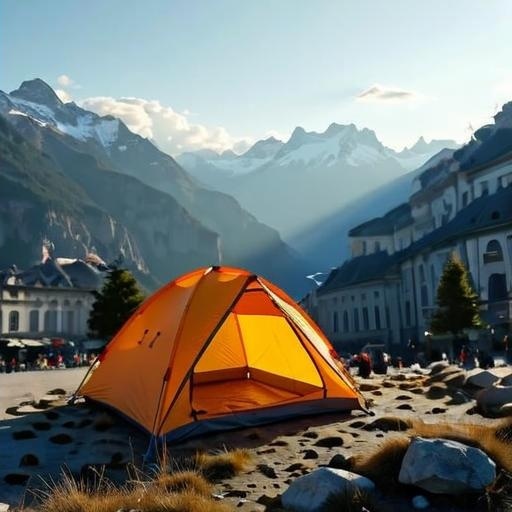}
			\\

			\raisebox{0.15cm}{\rotatebox[origin=c]{90}{\normalsize{{Ours}}}}
			&\includegraphics[width=3.15cm]{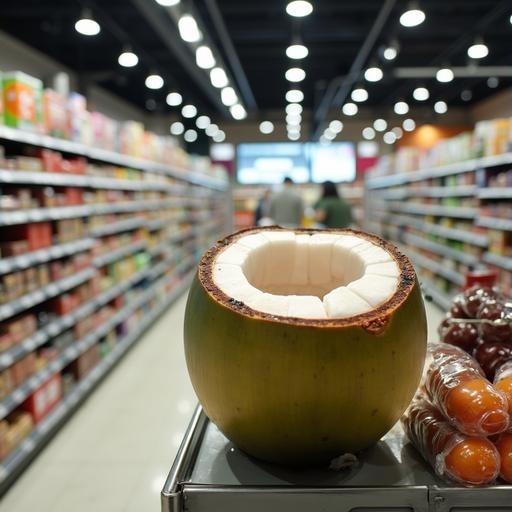}
			&\includegraphics[width=3.15cm]{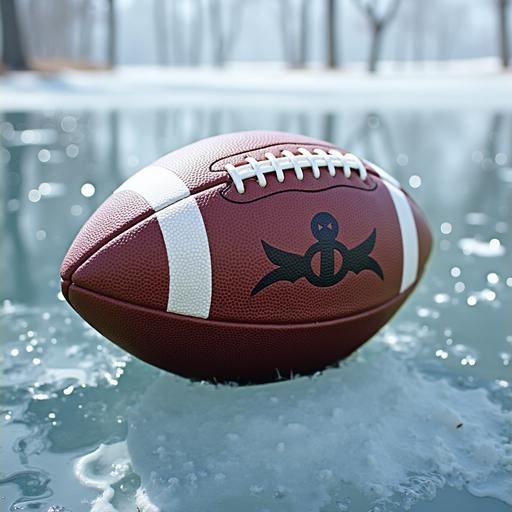}
			&\includegraphics[width=3.15cm]{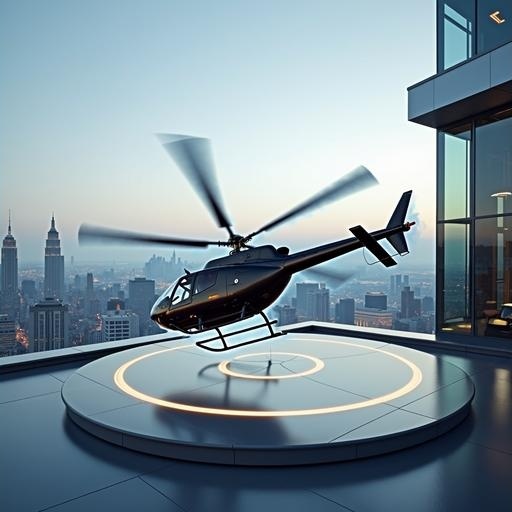}
			&\includegraphics[width=3.15cm]{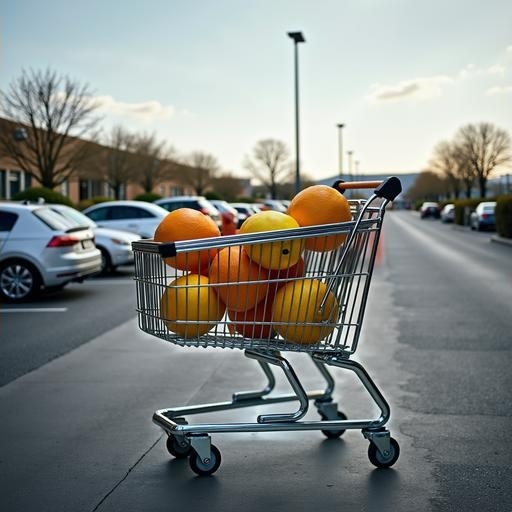}
			&\includegraphics[width=3.15cm]{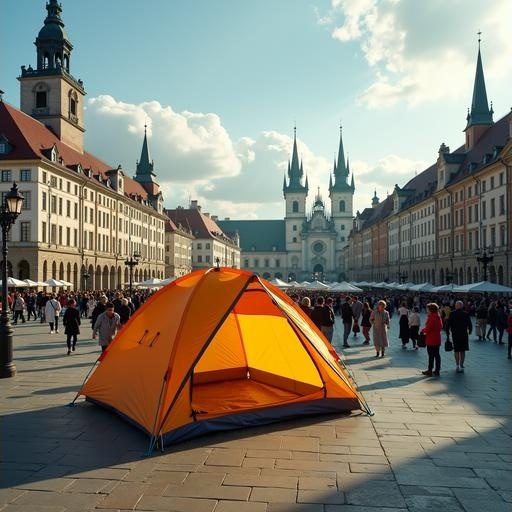}
			\\
			 & \small{\textit{``\textcolor{myblue}{Grocery Store}"}} & \small{\textit{``\textcolor{myblue}{Frozen Lake}"}}& \small{\textit{``\textcolor{myblue}{Rooftop Helipad}"}}& \small{\textit{``\textcolor{myblue}{Parking Lot}"}}& \small{\textit{``\textcolor{myblue}{City Square}"}}
			\\
			
		\end{tabular}
	\end{center}
	\caption{Qualitative comparison on the background replacement task with training-free methods StableFlow~\cite{avrahami2024stable} and TamingRF~\cite{wang2024taming}, as well as general image editing models MagicBrush~\cite{zhang2023magicbrush} and OmniGen~\cite{xiao2024omnigen}. Our approach delivers the most visually compelling background changes while preserving the foreground object intact.}
	\label{fig:quail_comparsion_bg_replace}
\end{figure*}

\begin{figure*}[t]
	\begin{center}
		\setlength{\tabcolsep}{0.5pt}
		\begin{tabular}{m{0.3cm}<{\centering}m{3.3cm}<{\centering}|m{3.3cm}<{\centering}m{3.3cm}<{\centering}m{3.3cm}<{\centering}m{3.3cm}<{\centering}}
			& \normalsize{Source Image} & \multicolumn{4}{ c }{{\normalsize{Edited Image}}}
			\\

			\multirow{3}{*}{\raisebox{-3.3cm}{\rotatebox[origin=c]{90}{\normalsize{{\textcolor{myblue}{Outpainting}}}}}}
			&\includegraphics[width=3.15cm]{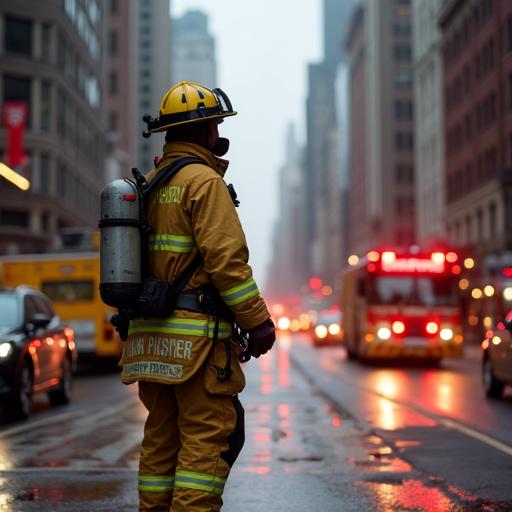}
			&\includegraphics[width=3.15cm]{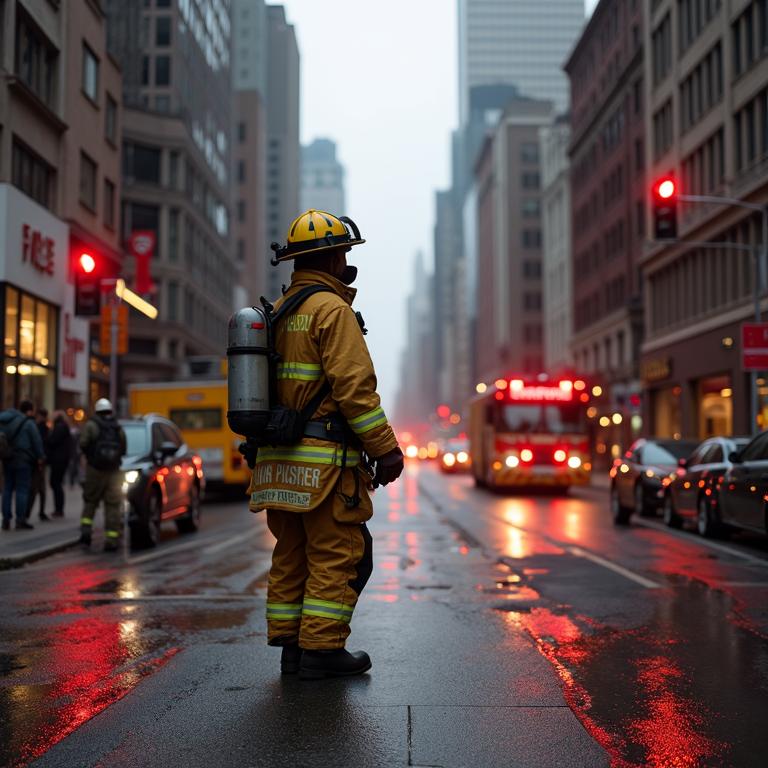}
			&\includegraphics[width=3.15cm]{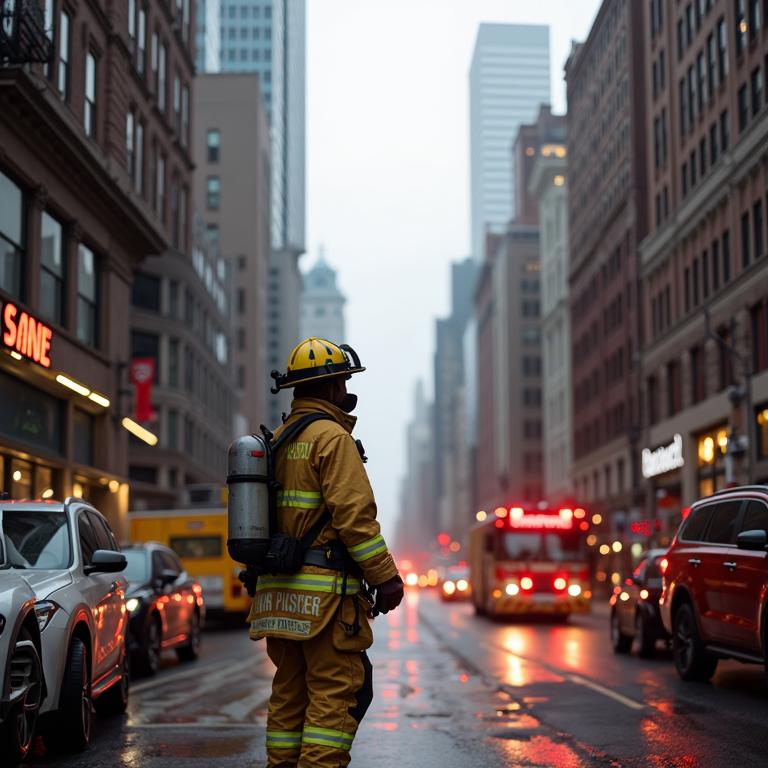}
			&\includegraphics[width=3.15cm]{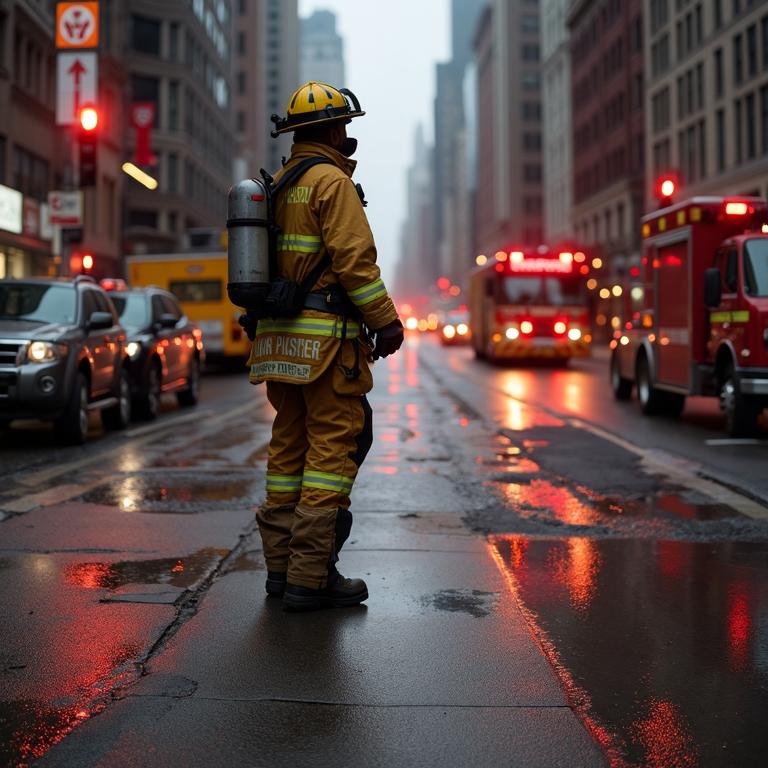}
			&\includegraphics[width=3.15cm]{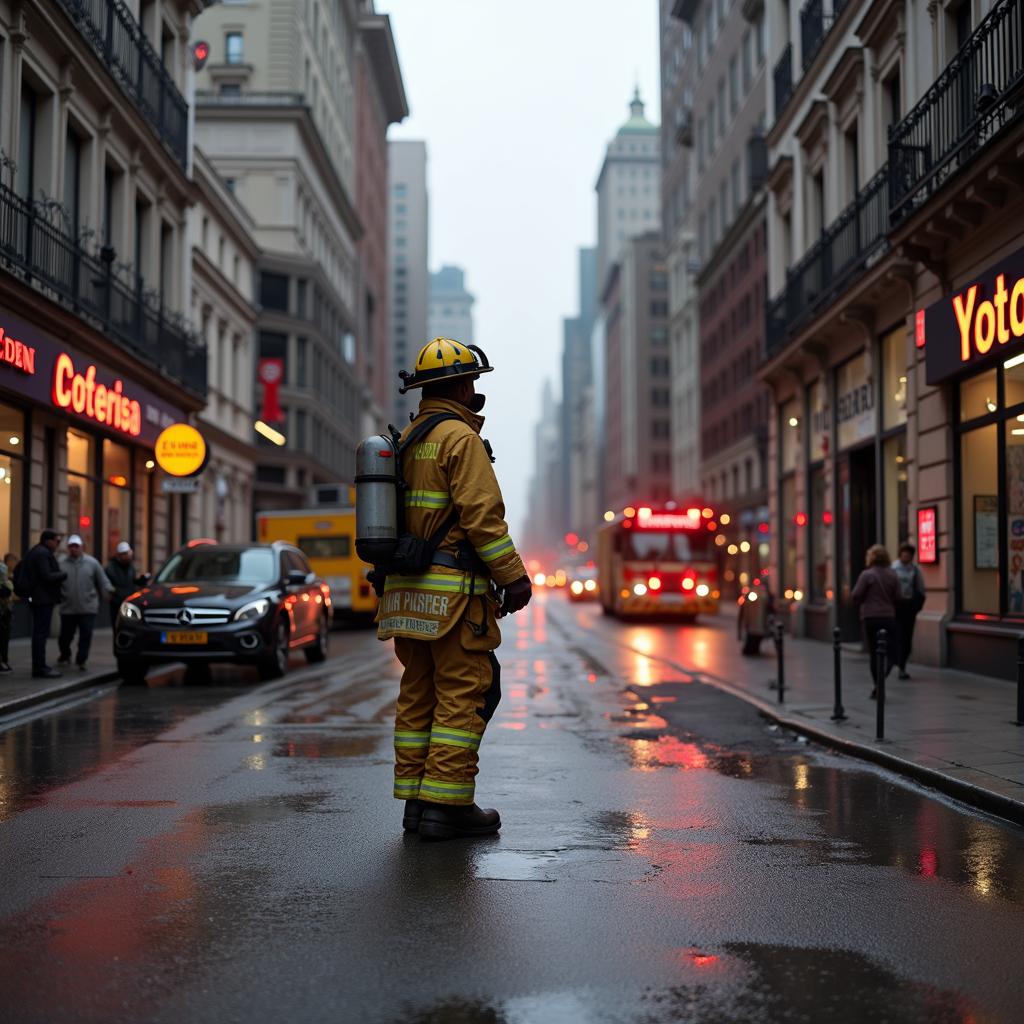}
			\\
			&\includegraphics[width=3.15cm]{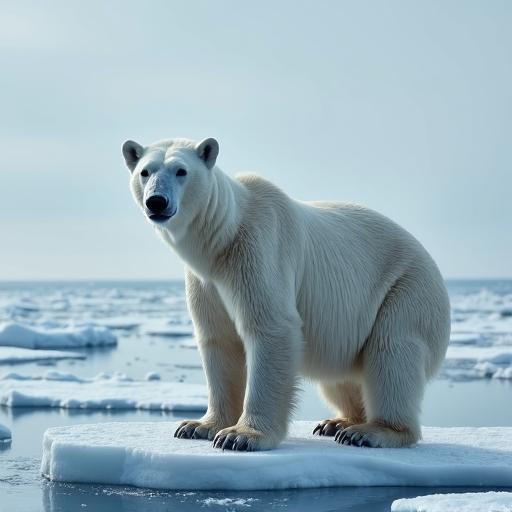}
			&\includegraphics[width=3.15cm]{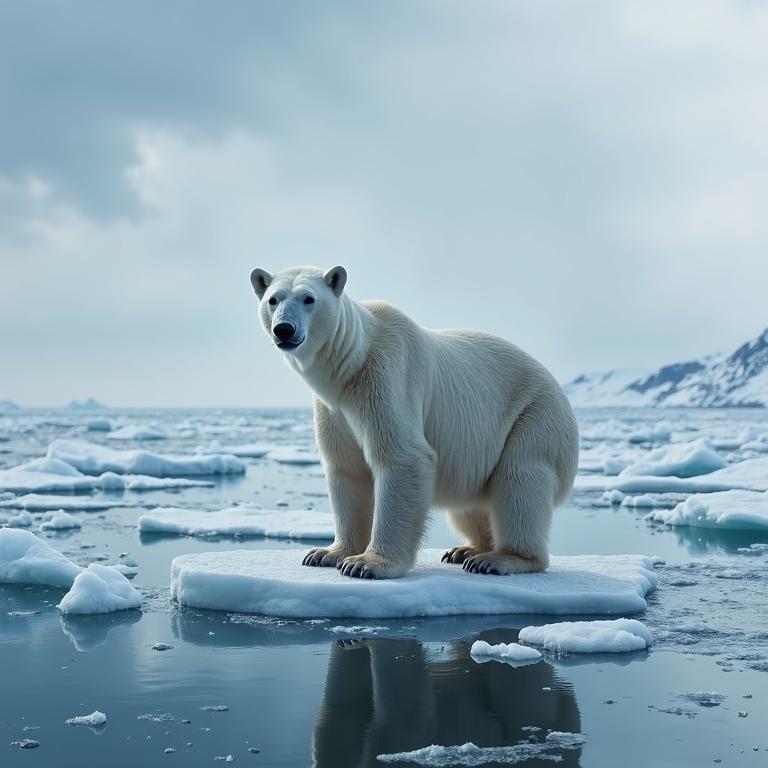}
			&\includegraphics[width=3.15cm]{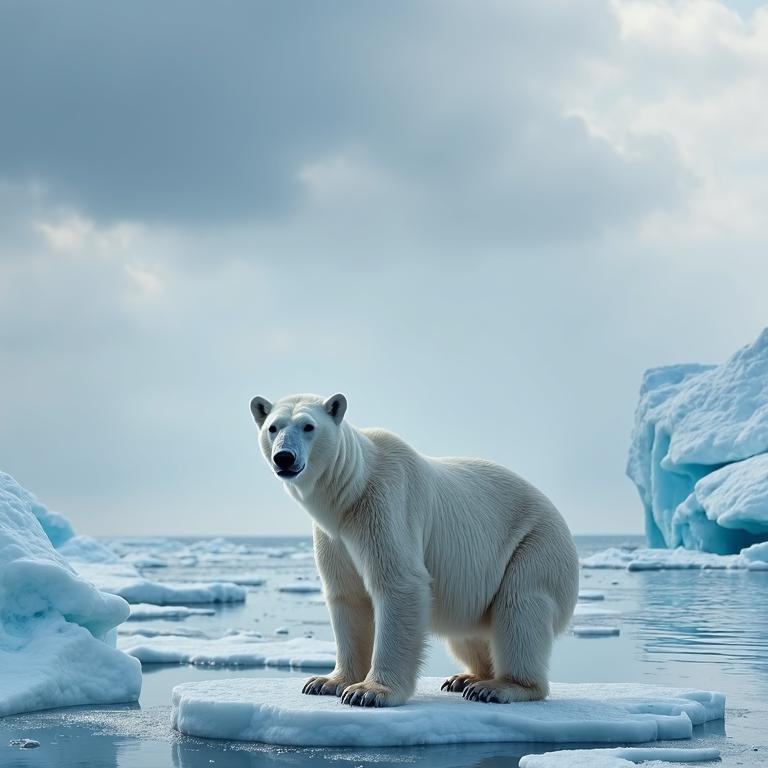}
			&\includegraphics[width=3.15cm]{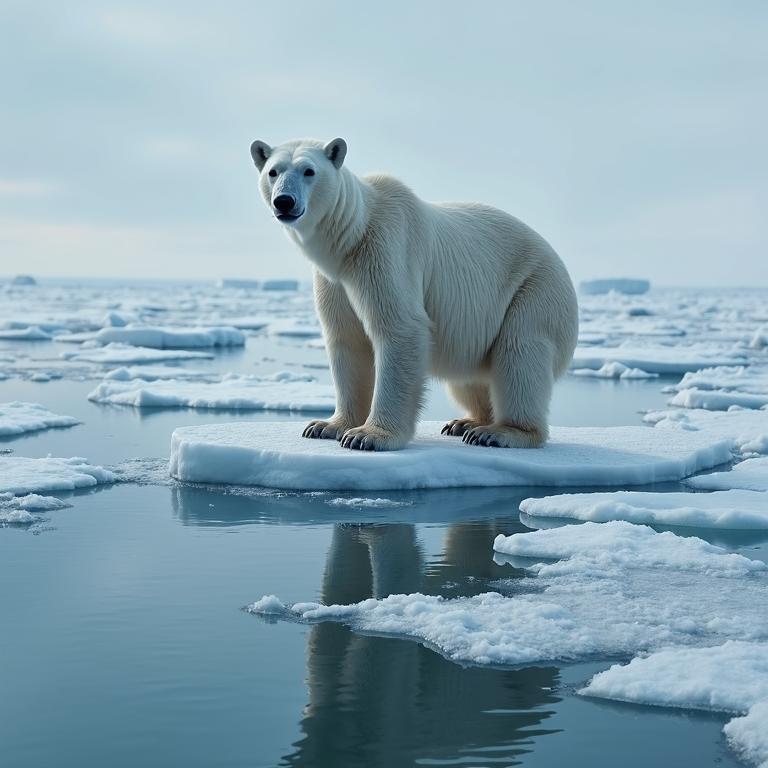}
			&\includegraphics[width=3.15cm]{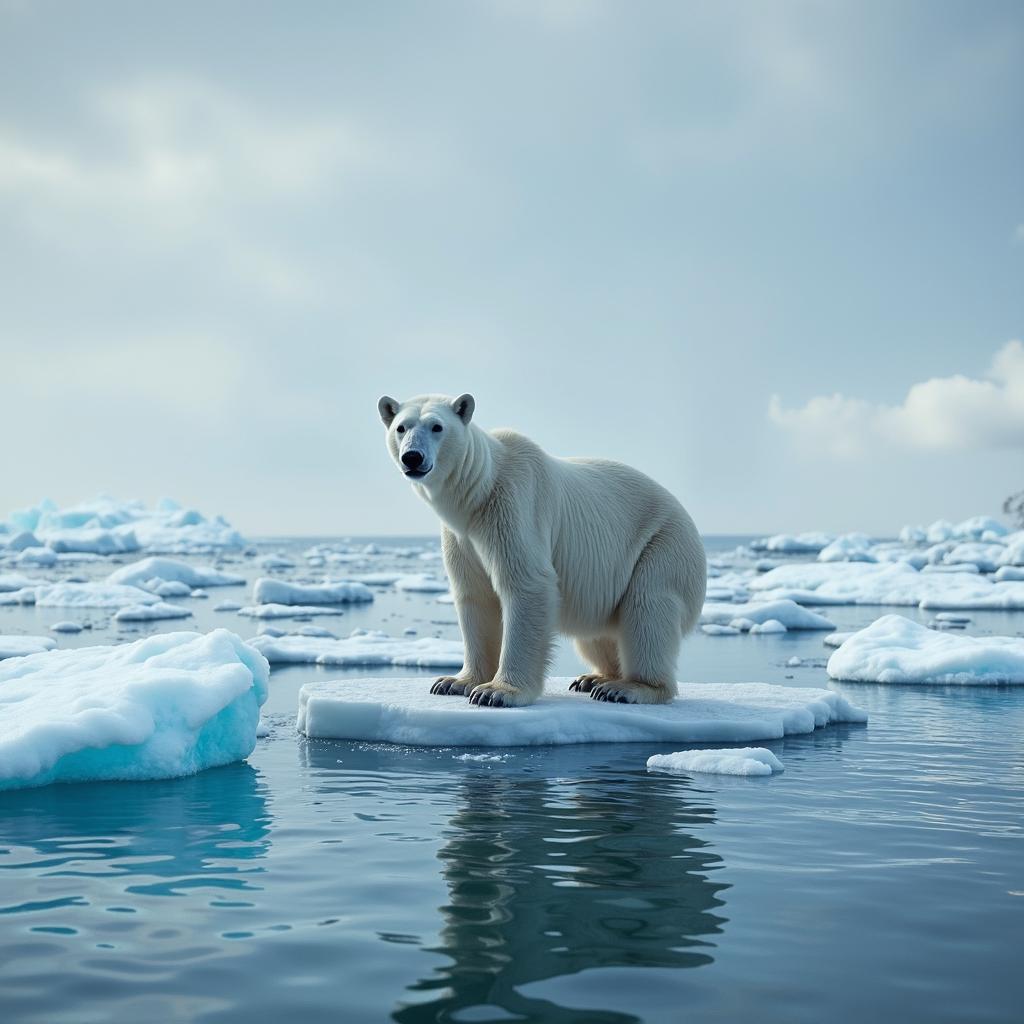}
			\\
			&\includegraphics[width=3.15cm]{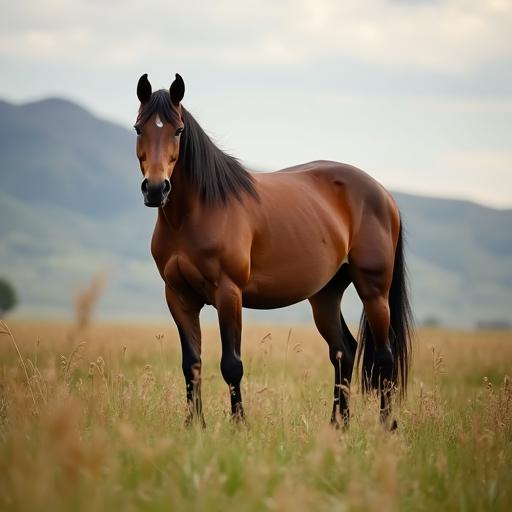}
			&\includegraphics[width=3.15cm]{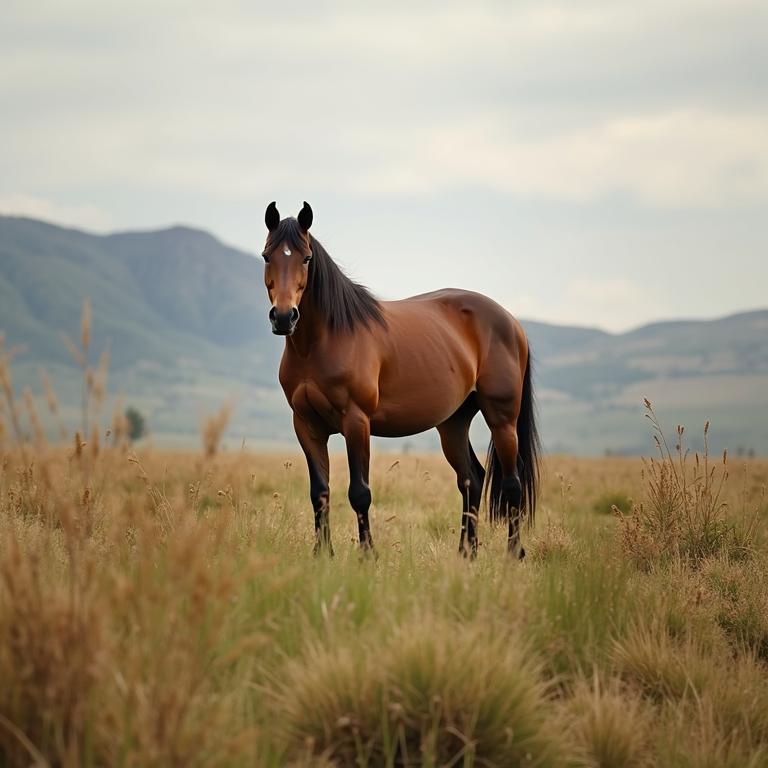}
			&\includegraphics[width=3.15cm]{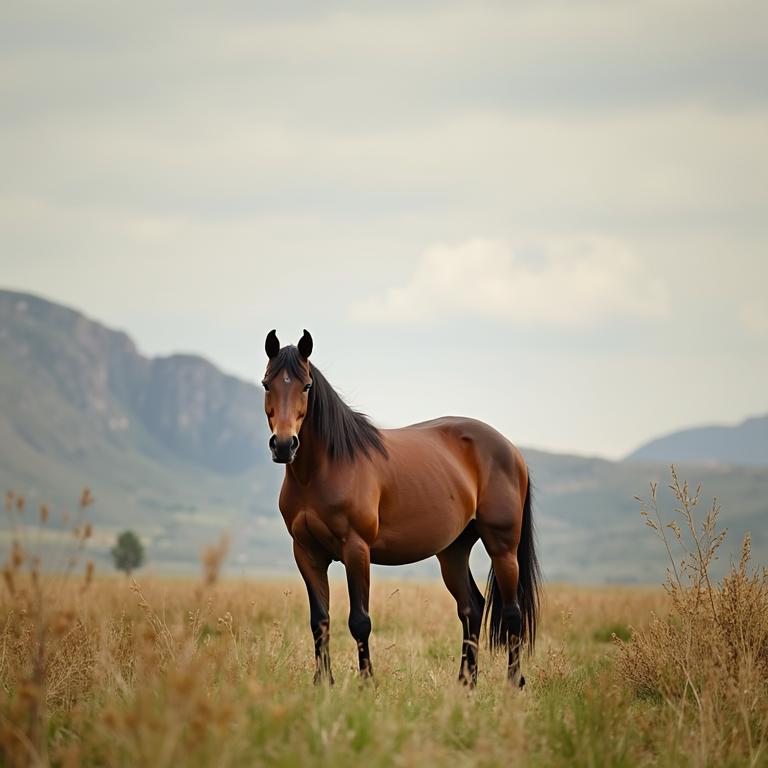}
			&\includegraphics[width=3.15cm]{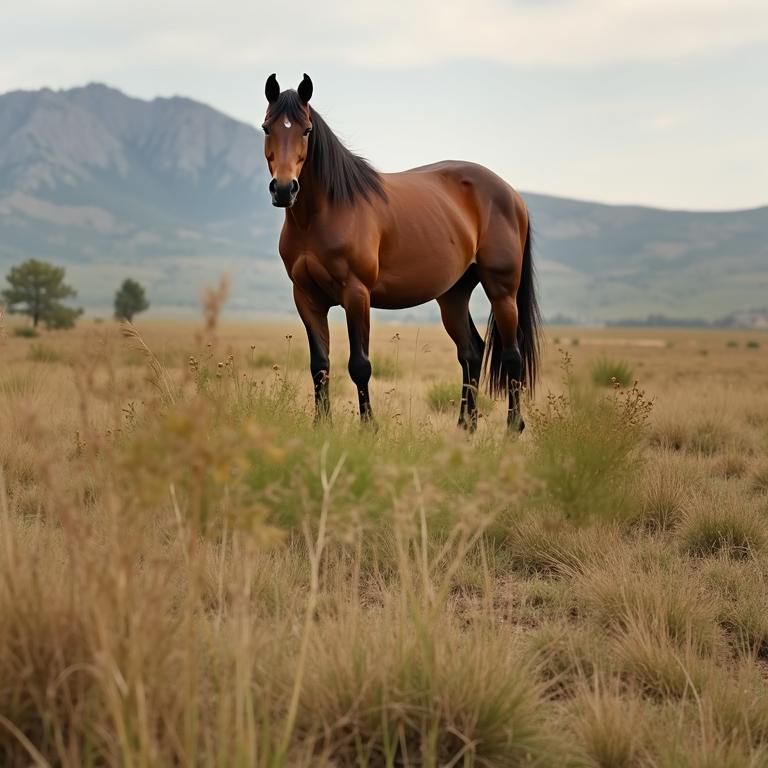}
			&\includegraphics[width=3.15cm]{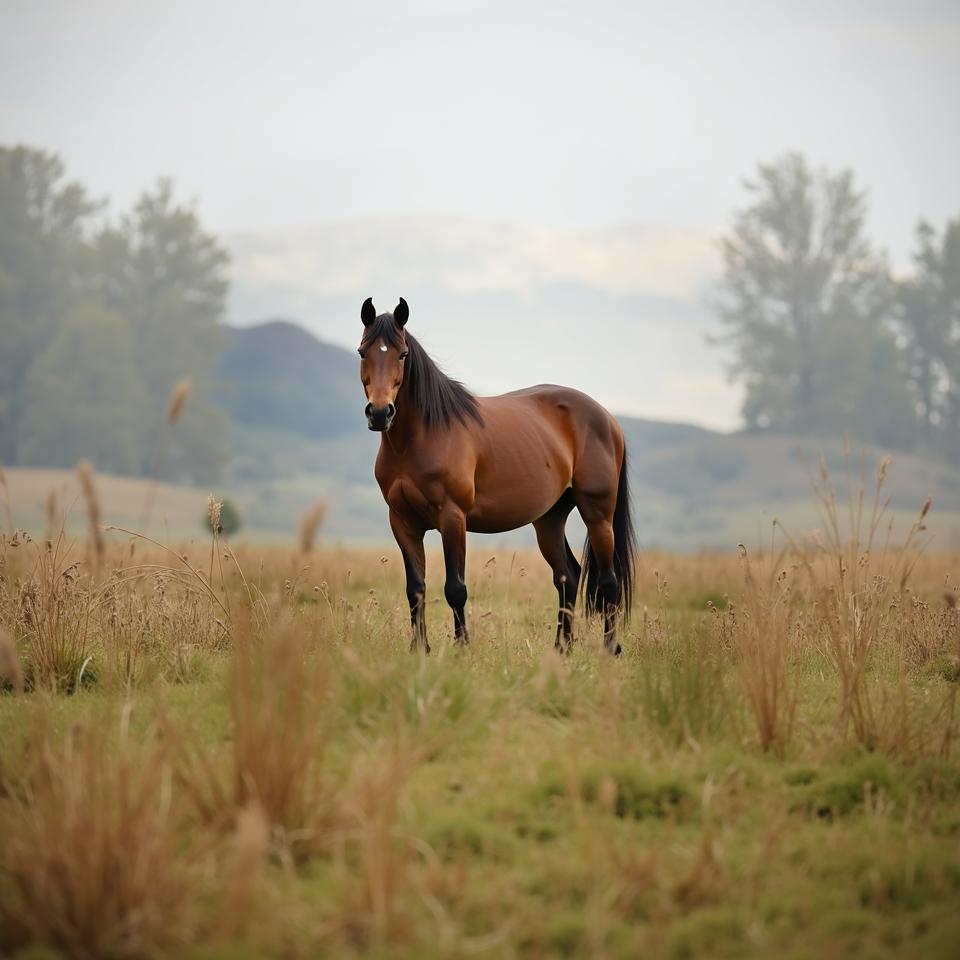}
			\\

                \cdashline{1-6}
                \noalign{\vskip 0.15cm}
            
			\multirow{3}{*}{\raisebox{-3.2cm}{\rotatebox[origin=c]{90}{\normalsize{{\textcolor{myblue}{Object Movement}}}}}}
			&\includegraphics[width=3.15cm]{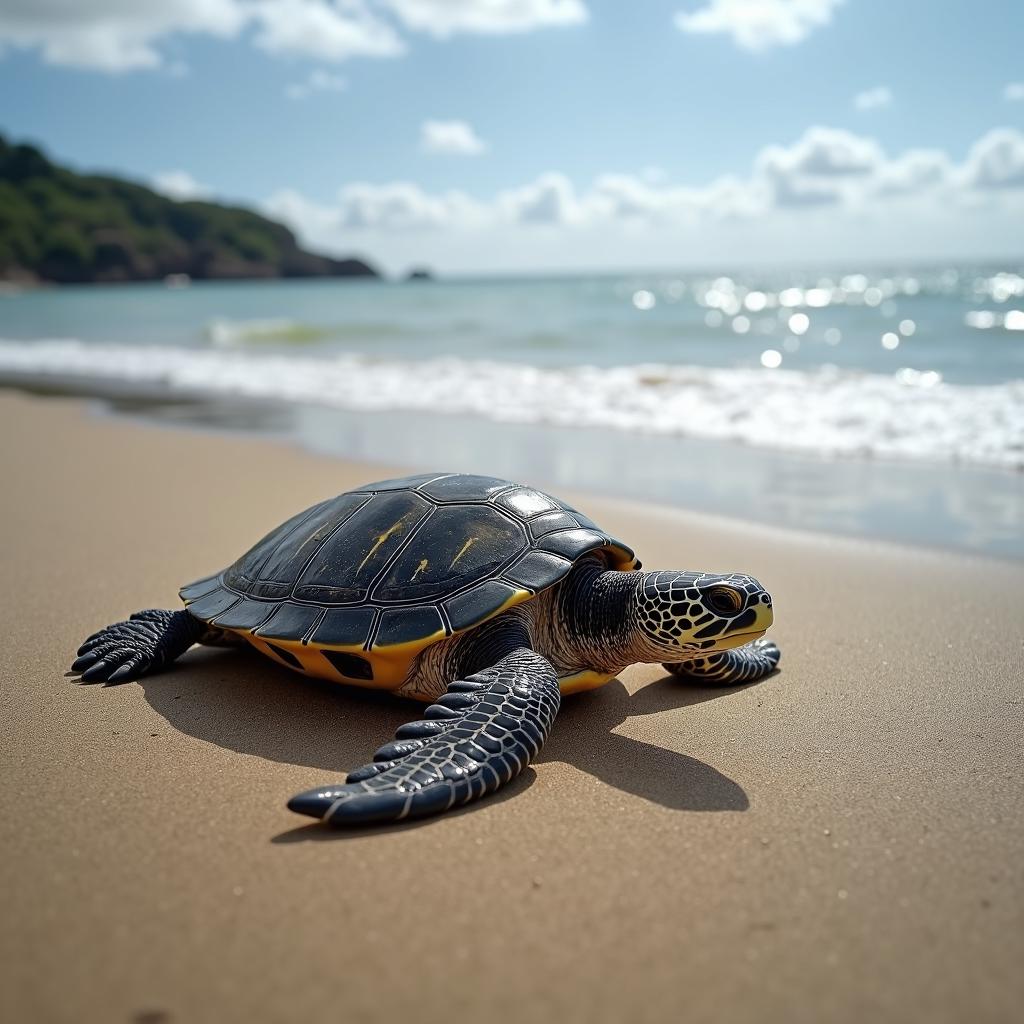}
			&\includegraphics[width=3.15cm]{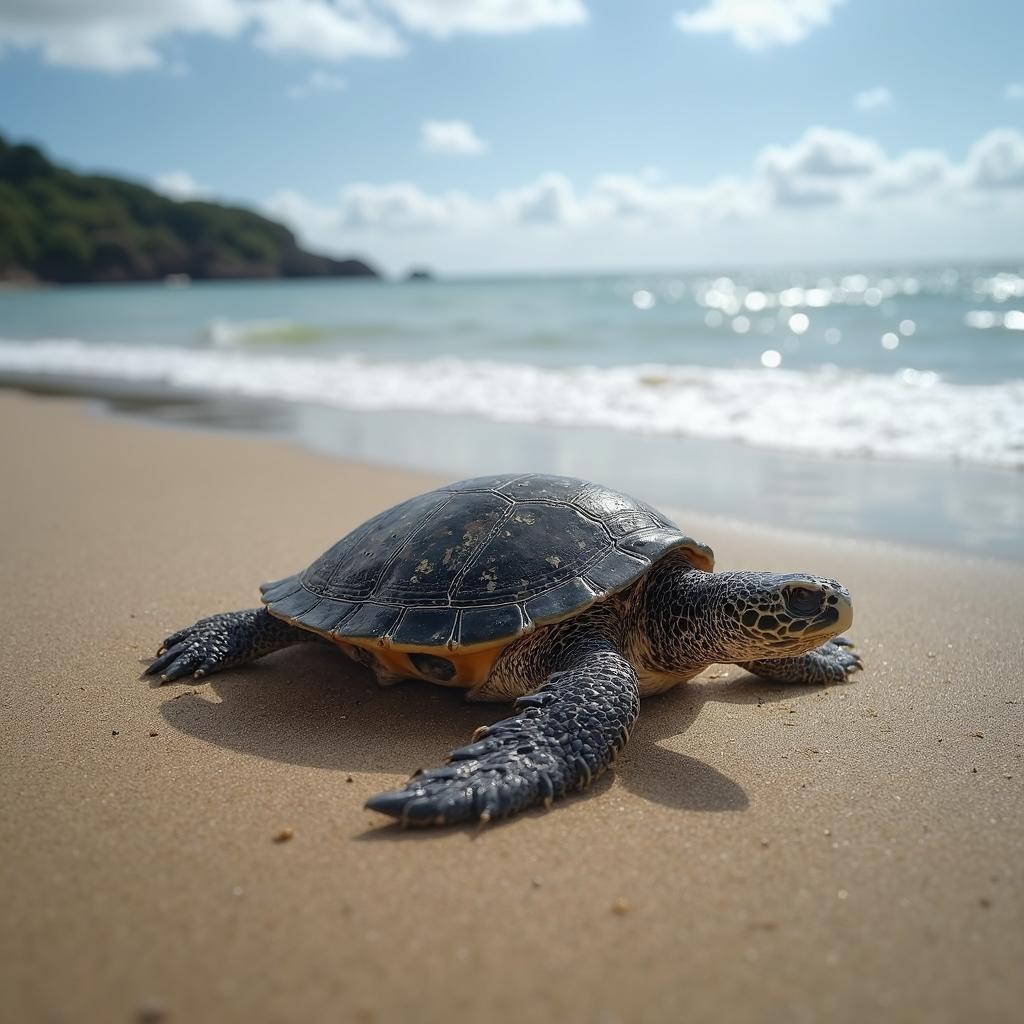}
			&\includegraphics[width=3.15cm]{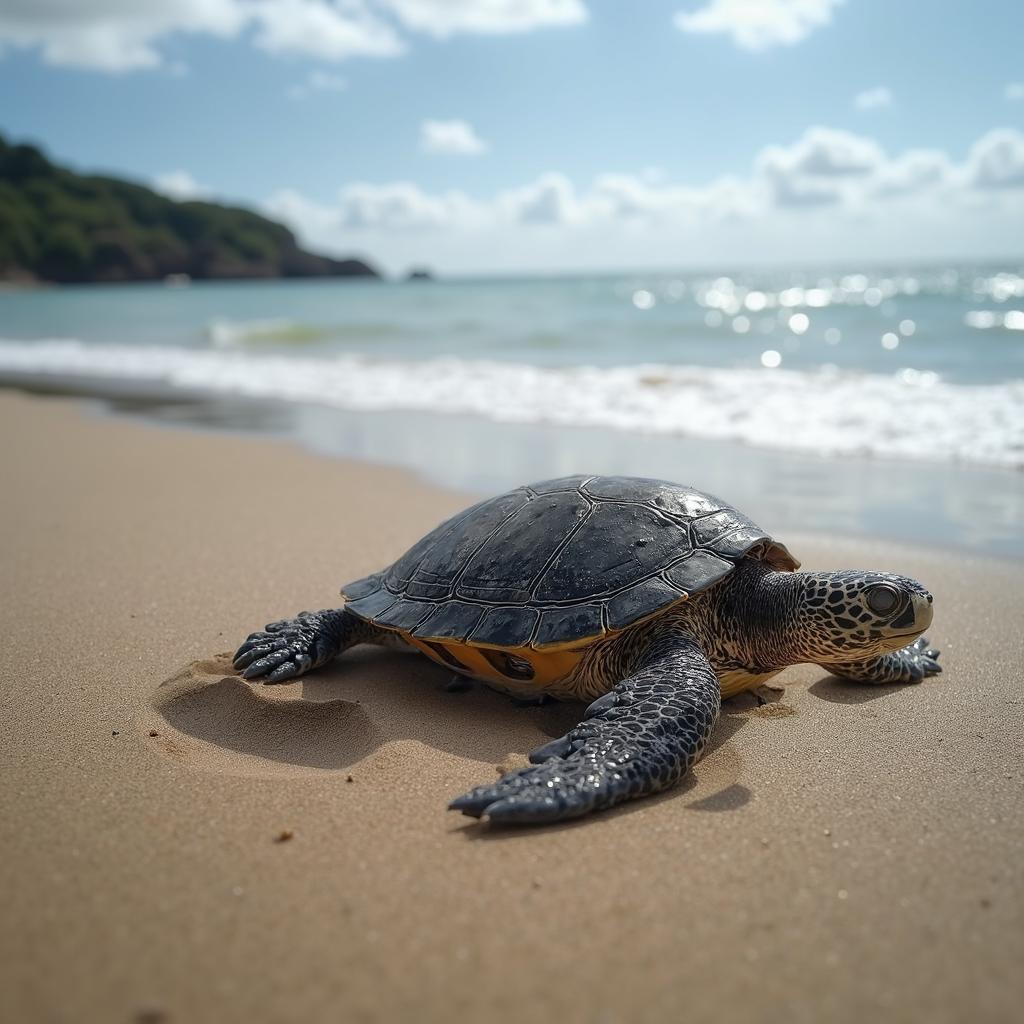}
			&\includegraphics[width=3.15cm]{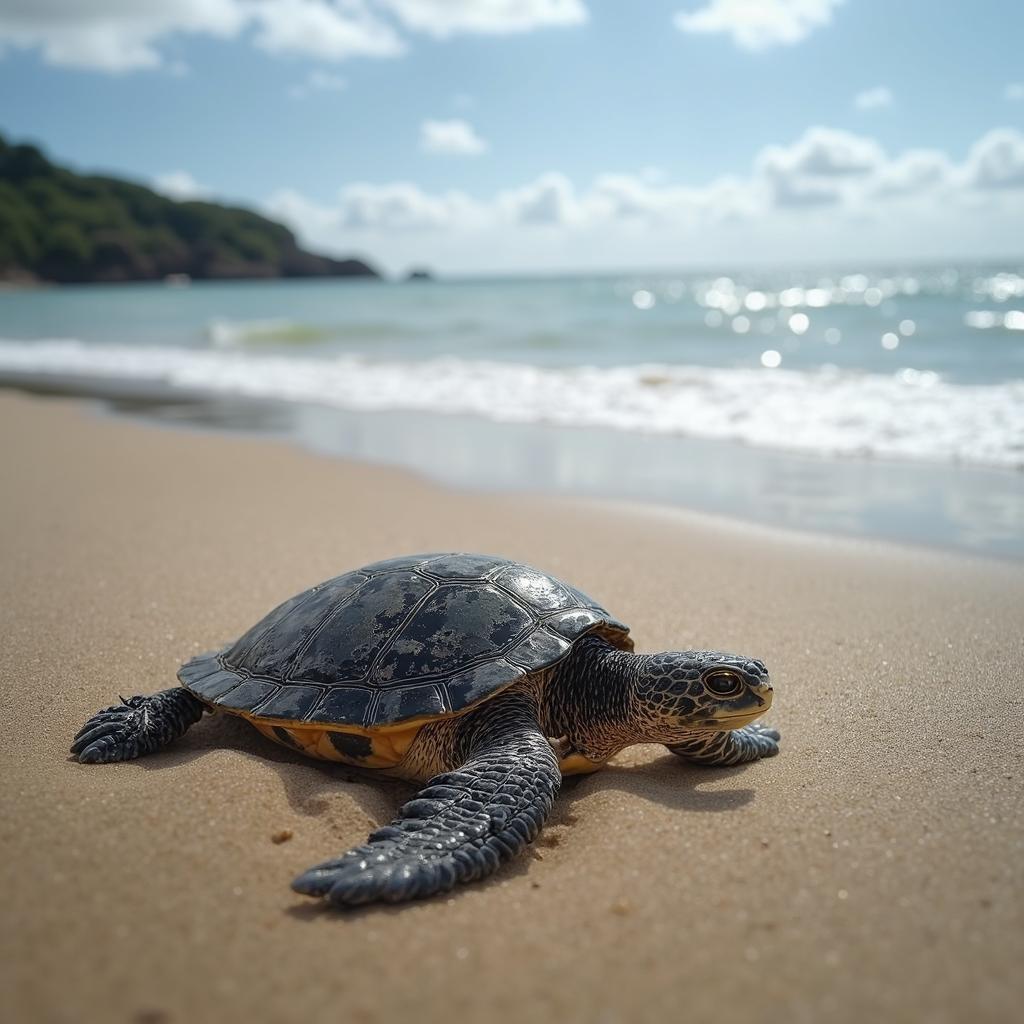}
			&\includegraphics[width=3.15cm]{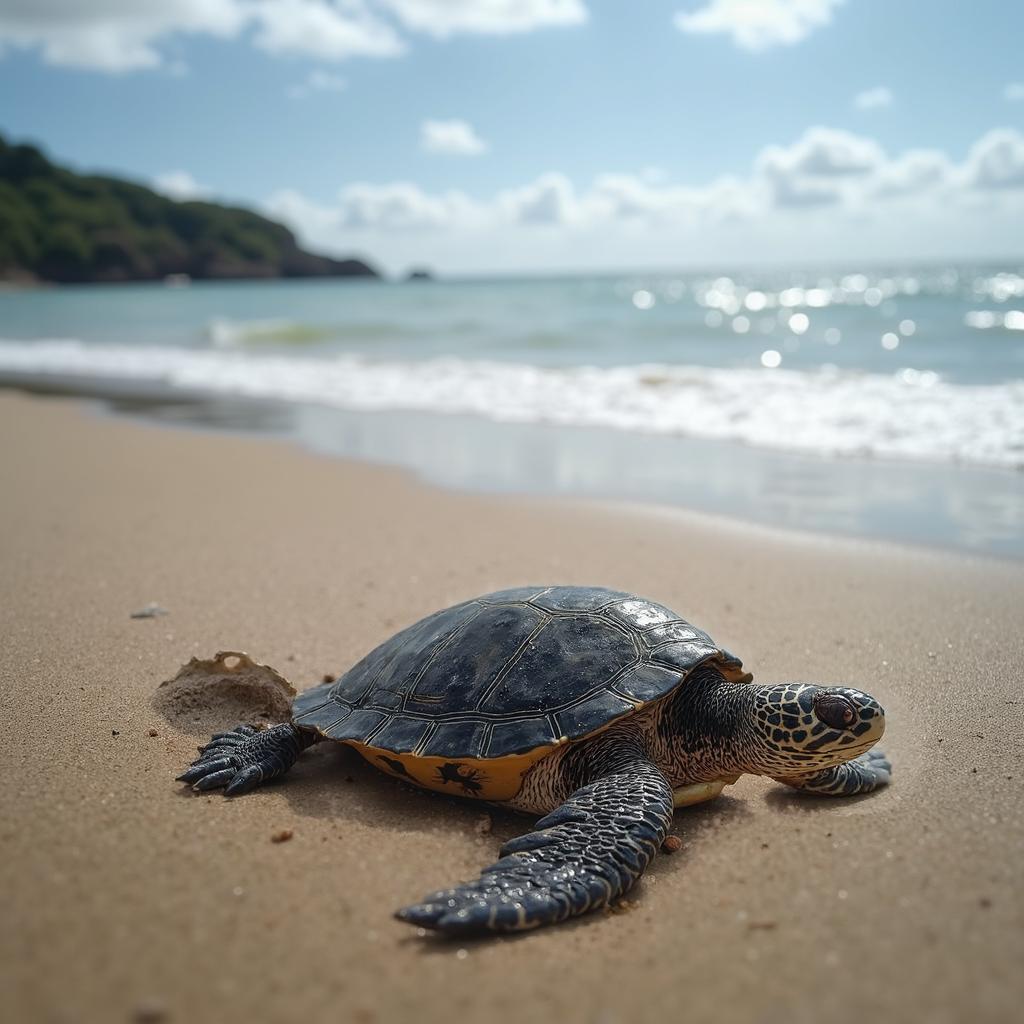}
			\\
			&\includegraphics[width=3.15cm]{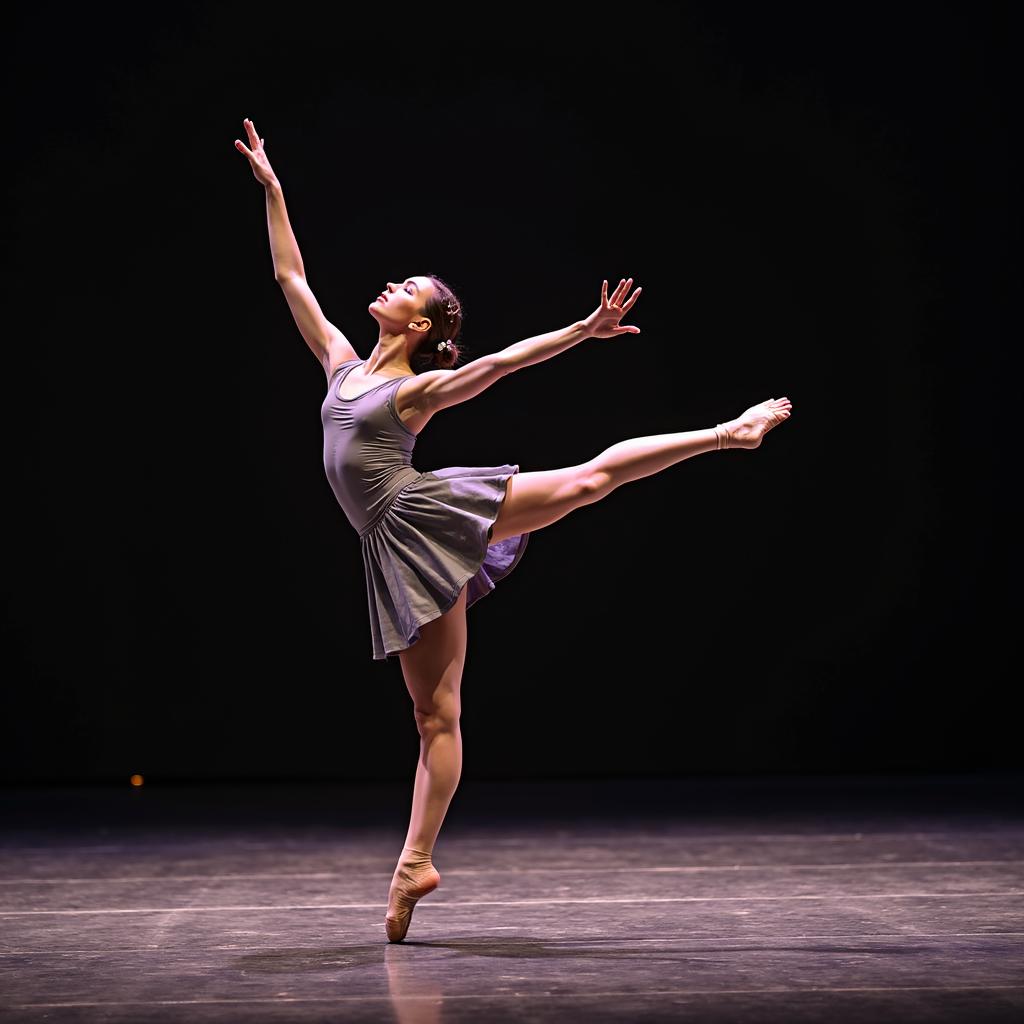}
			&\includegraphics[width=3.15cm]{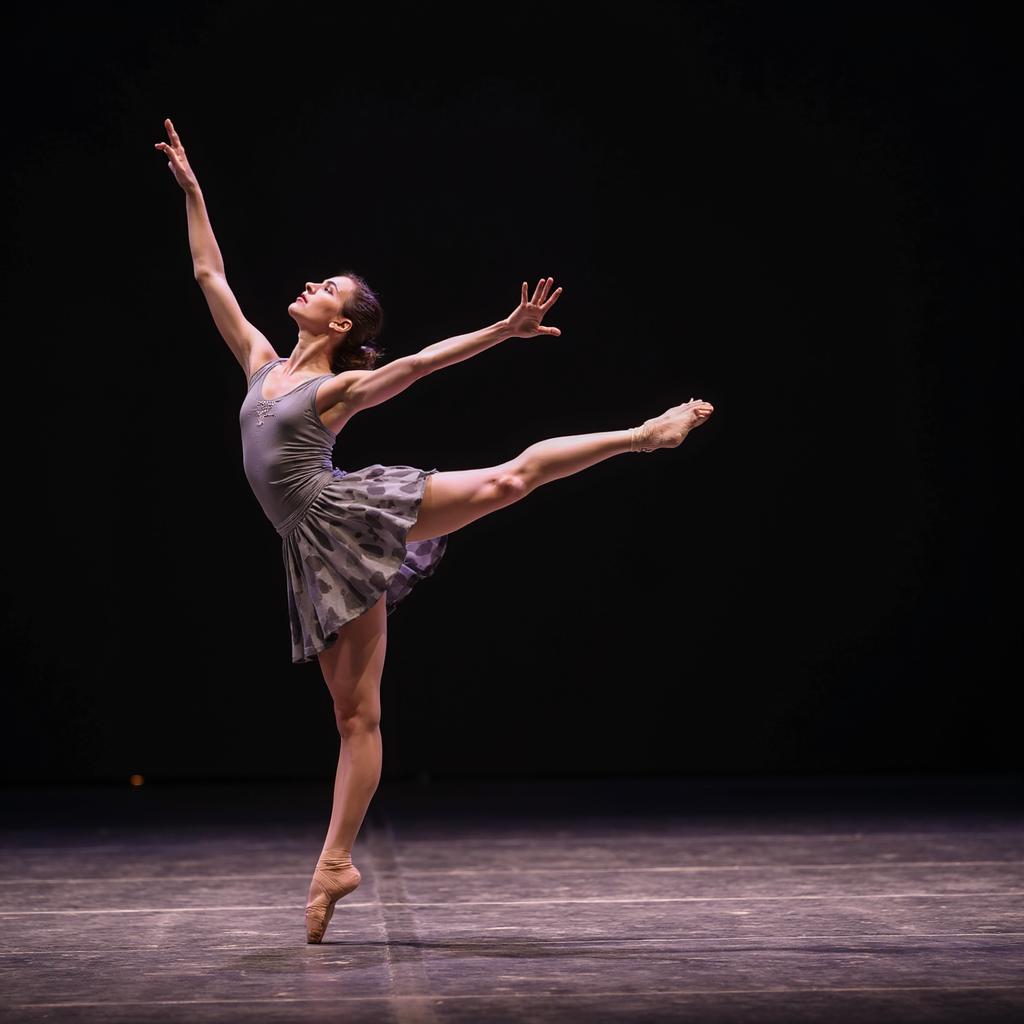}
			&\includegraphics[width=3.15cm]{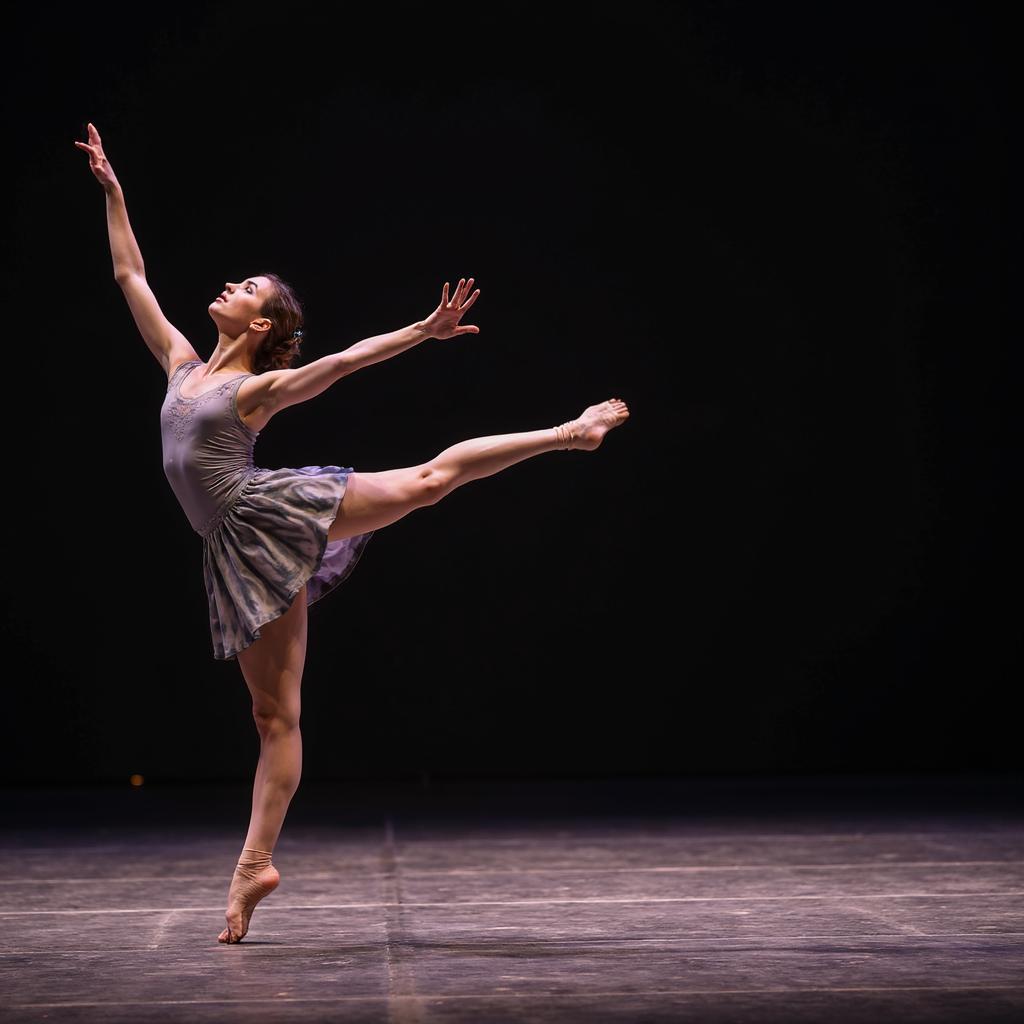}
			&\includegraphics[width=3.15cm]{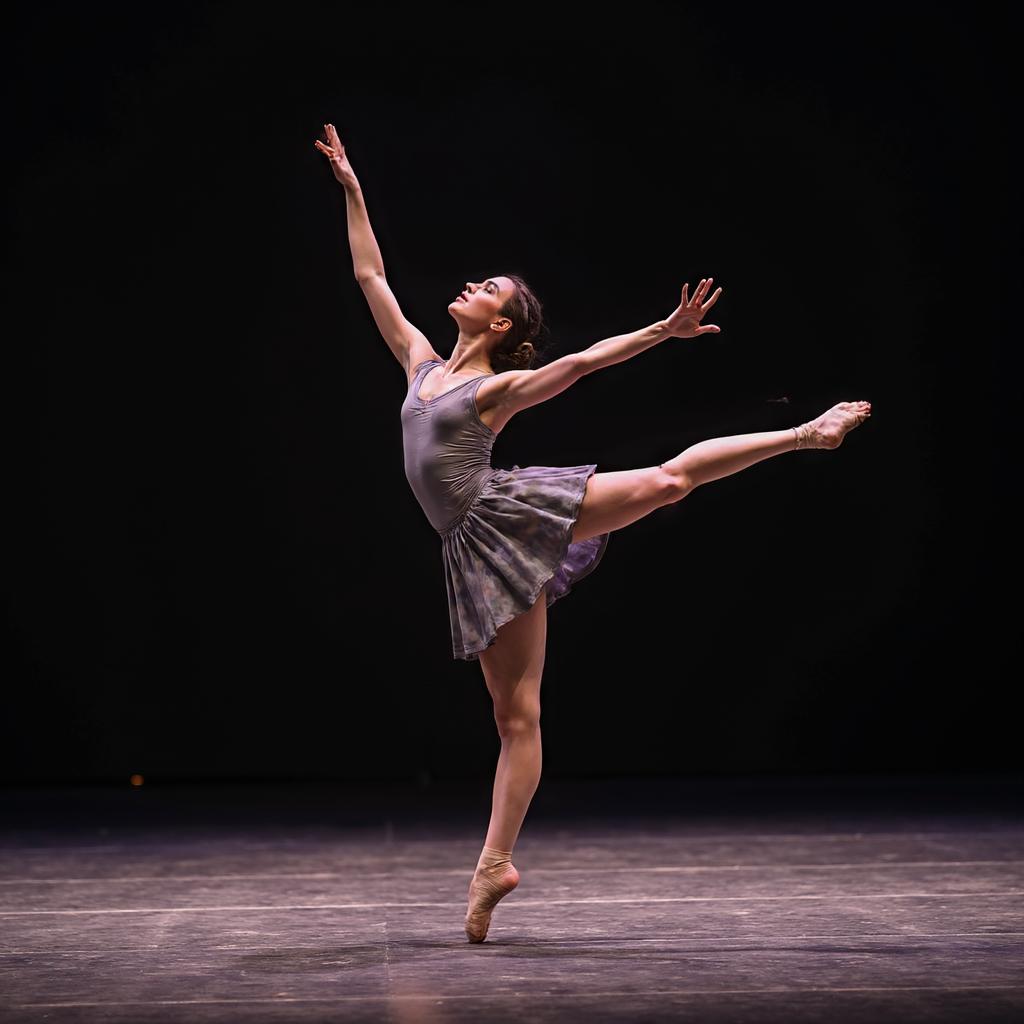}
			&\includegraphics[width=3.15cm]{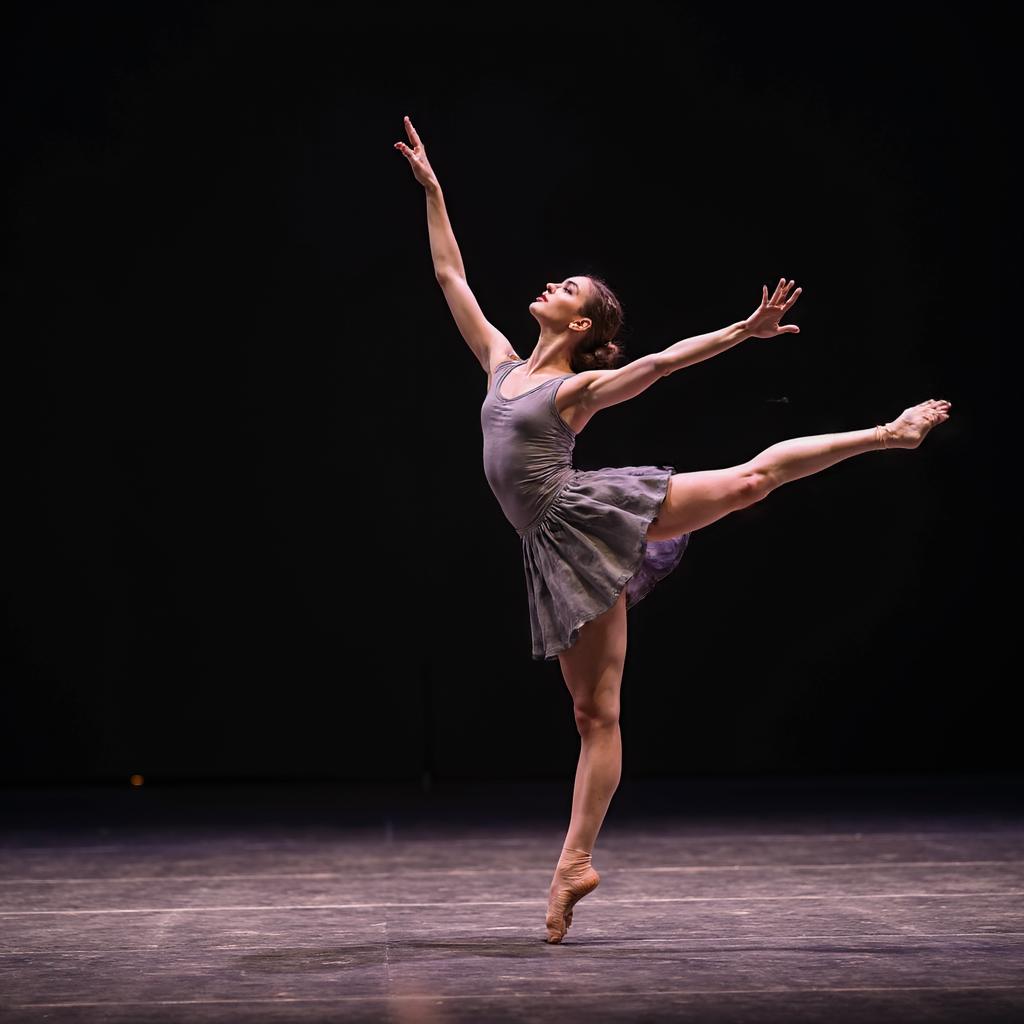}
			\\
			&\includegraphics[width=3.15cm]{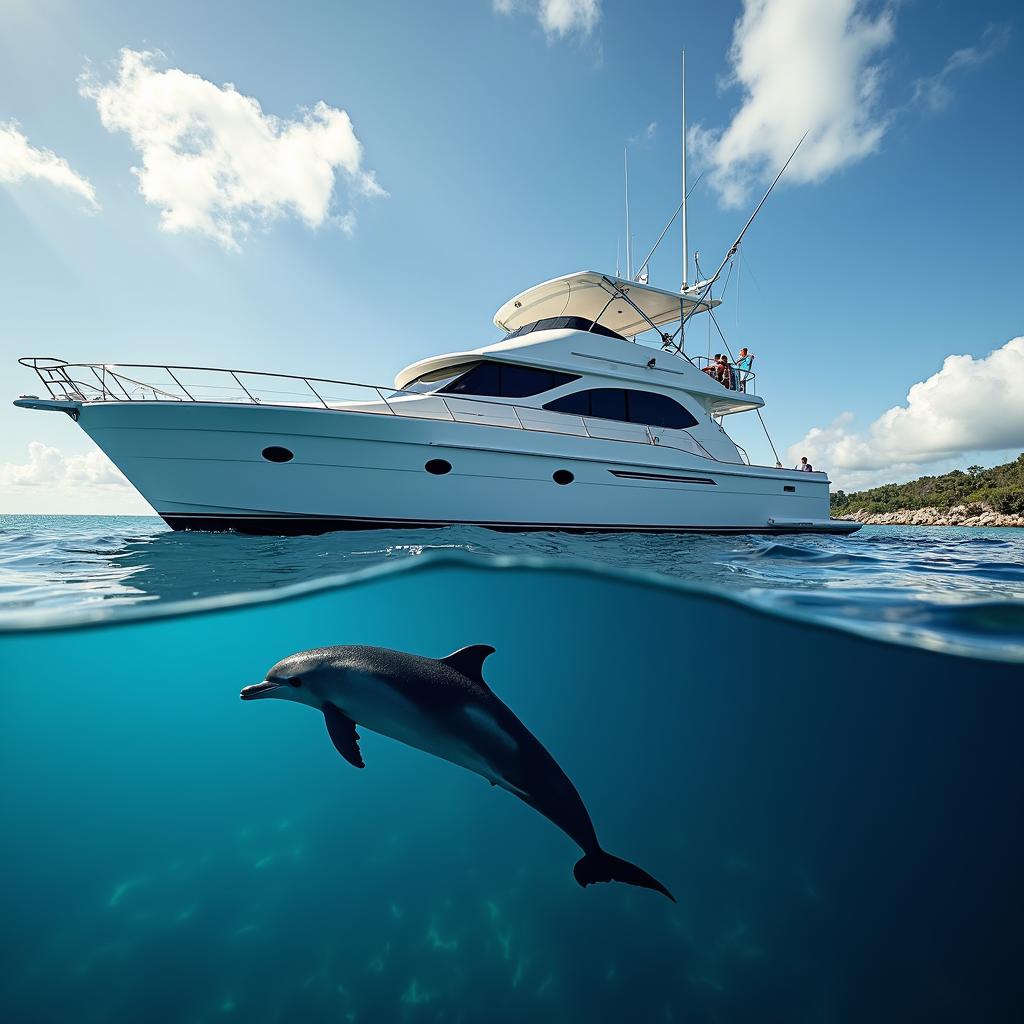}
			&\includegraphics[width=3.15cm]{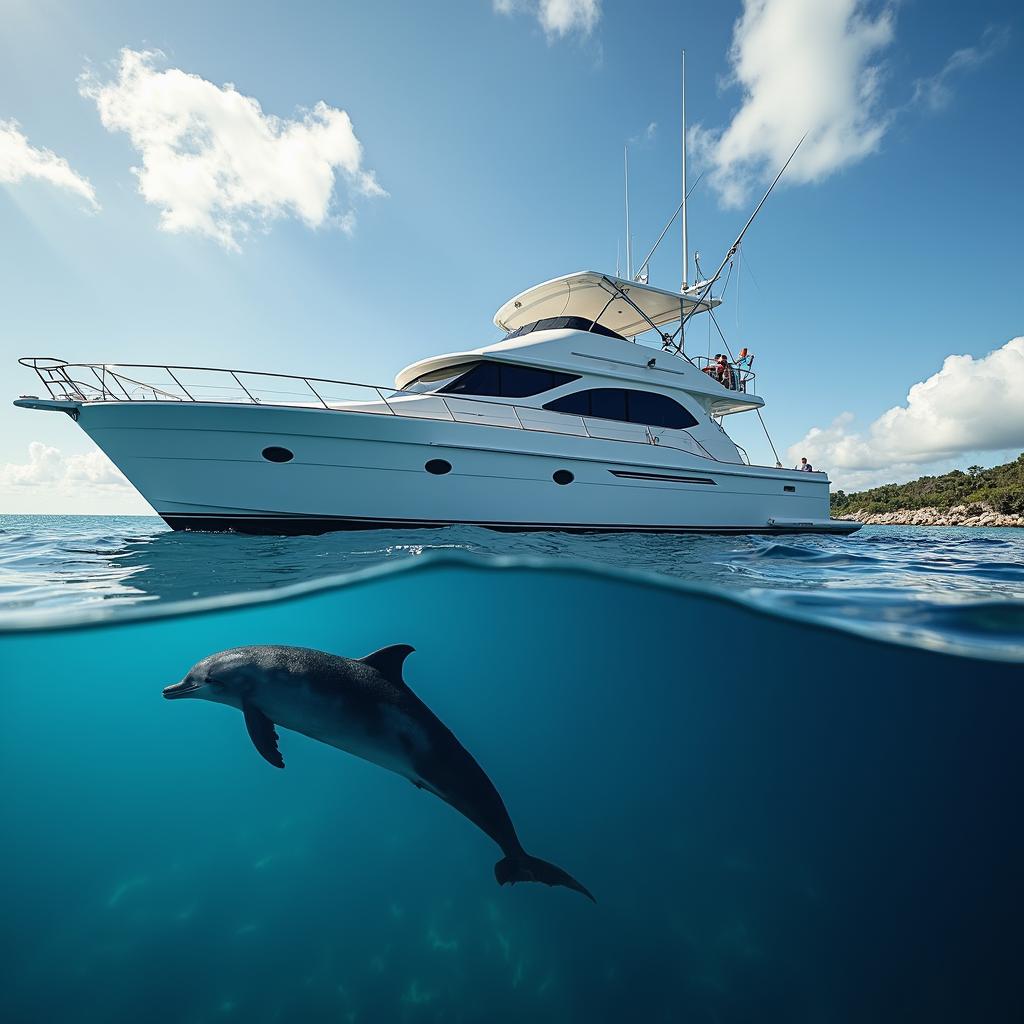}
			&\includegraphics[width=3.15cm]{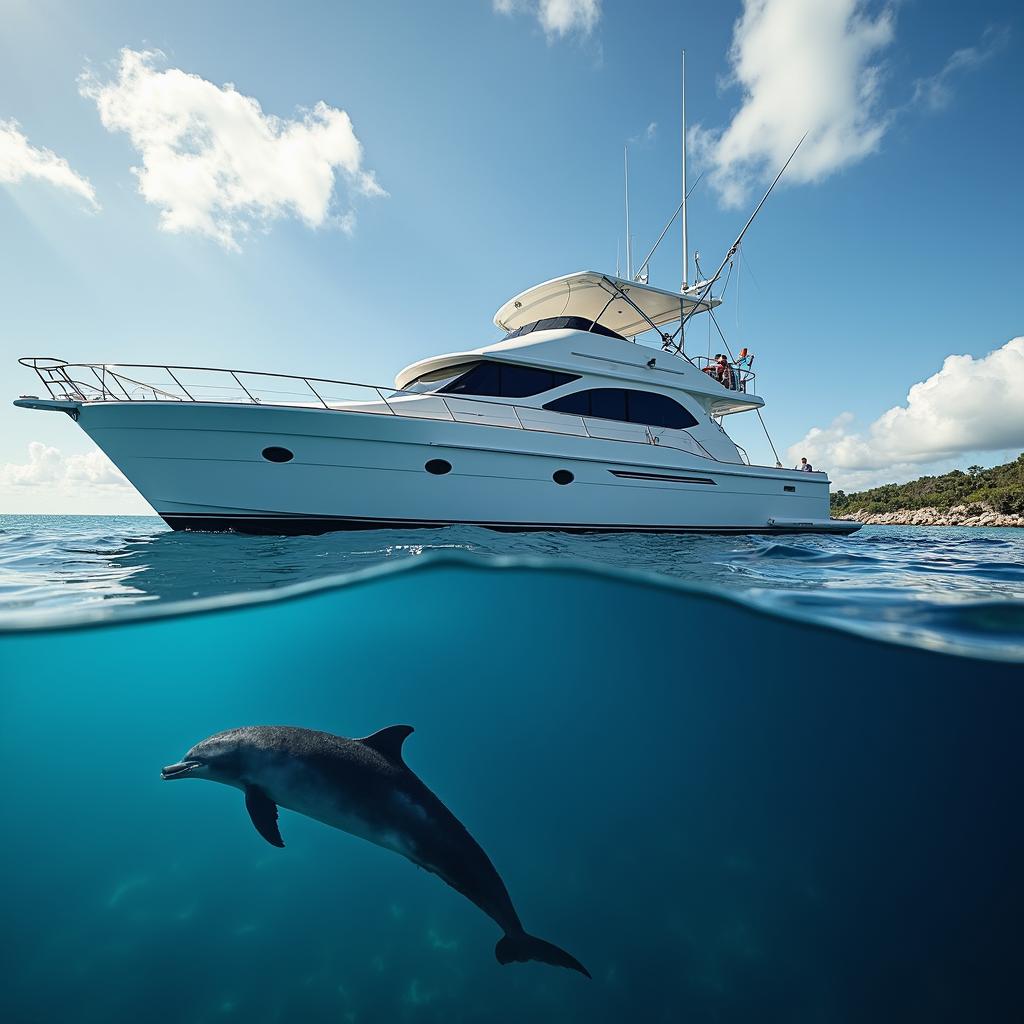}
			&\includegraphics[width=3.15cm]{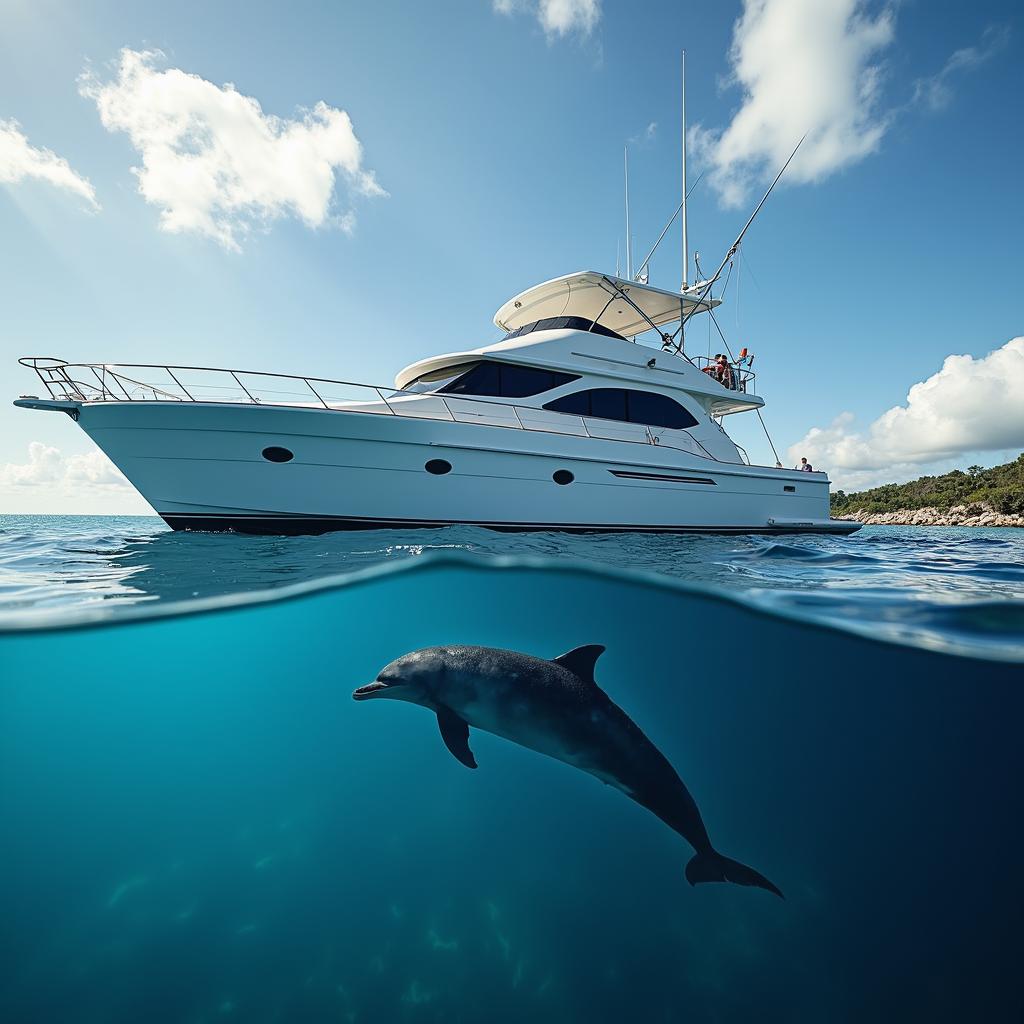}
			&\includegraphics[width=3.15cm]{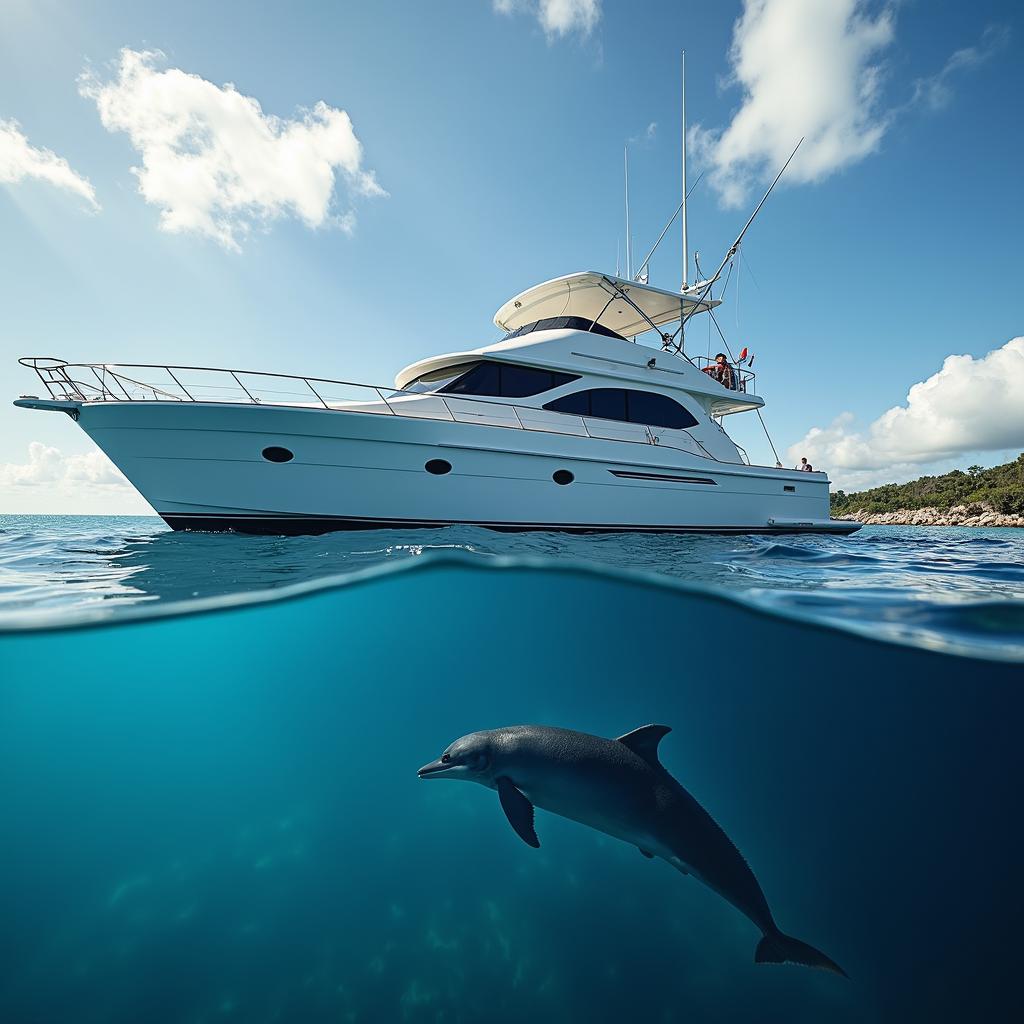}
			\\
            
		\end{tabular}
	\end{center}
	\caption{Visual results of our method on region-preserved editing tasks such as object movement and outpainting.} 
	\label{fig:move_outpainting_supp}
\end{figure*}

In Figures \ref{fig:quail_comparsion_add_object}, \ref{fig:quail_comparsion_non_rigid}, \ref{fig:quail_comparsion_bg_replace}, and \ref{fig:move_outpainting_supp} we give more visual comparison results with other methods and our results on the outpainting and object moving tasks.

\end{document}